\documentclass[10pt,twocolumn,letterpaper]{article}

\usepackage[pagenumbers]{cvpr} %

\usepackage{graphicx}
\usepackage{amsmath}
\usepackage{amssymb}
\usepackage{booktabs}

\usepackage{acronym}
\usepackage{caption}
\usepackage{subcaption}
\usepackage{cuted}
\usepackage{adjustbox}

\usepackage{siunitx}
\usepackage{booktabs,makecell}
\usepackage{etoolbox}

\usepackage[ruled,vlined]{algorithm2e}

\usepackage{setspace}

\usepackage[title]{appendix}

\usepackage{enumitem}
\setlist{topsep=0pt, leftmargin=*,noitemsep,topsep=0pt,parsep=0pt,partopsep=0pt}

\usepackage{hyperref}
\hypersetup{pagebackref,breaklinks,colorlinks}

\usepackage[capitalize]{cleveref}
\crefname{section}{Sec.}{Secs.}
\Crefname{section}{Section}{Sections}
\Crefname{table}{Table}{Tables}
\crefname{table}{Tab.}{Tabs.}

\def\cvprPaperID{5907} %
\def\confName{CVPR}
\def\confYear{2022}

\acrodef{nec}[NEC]{normal epipolar constraint}
\acrodef{pnec}[PNEC]{probabilistic normal epipolar constraint}
\acrodef{klt}[KLT]{Kanade-Lucas-Tomasi}
\acrodef{grq}[GRQ]{generalized Rayleigh quotient}
\acrodef{scf}[SCF]{self-consistent-field}
\acrodef{irls}[IRLS]{iteratively reweighted least squares}

\newcommand{\termformat}[1]{\emph{#1}}
\newcommand{\epipolarnormalplane}{\termformat{epipolar normal plane}}

\usepackage{xcolor}
\definecolor{LRD}{RGB}{27,158,119}
\definecolor{LRU}{RGB}{217,95,2}
\newcommand{\cu}[1]{\textcolor{LRU}{#1}}
\newcommand{\cd}[1]{\textcolor{LRD}{#1}}

\newcommand{\R}{\ensuremath{\mathbb{R}}}
\newcommand{\f}[1]{\ensuremath{\boldsymbol{#1}}}
\newcommand{\ft}{\f{t}}
\newcommand{\fR}{\f{R}}
\DeclareMathOperator{\Opt}{Opt}

\newcommand{\ff}{\f{f}}
\newcommand{\hff}{\hat{\ff}}
\newcommand{\fp}{\f{p}}
\newcommand{\fT}{\f{T}}
\newcommand{\fx}{\f{x}}
\DeclareMathOperator{\FibonacciLattice}{FibonacciLattice}

\newcommand{\pseudoparagraph}[1]{\textbf{#1}}

\newcommand\blfootnote[1]{%
  \begingroup
  \renewcommand\thefootnote{}\footnote{#1}%
  \addtocounter{footnote}{-1}%
  \endgroup
}

\begin{document}

\title{The Probabilistic Normal Epipolar Constraint for Frame-To-Frame Rotation Optimization under Uncertain Feature Positions}

\author{Dominik Muhle\textsuperscript{\textasteriskcentered \textdagger}, Lukas Koestler\textsuperscript{\textasteriskcentered \textdagger}, Nikolaus Demmel\textsuperscript{\textdagger}, Florian Bernard\textsuperscript{\textdaggerdbl} and Daniel Cremers\textsuperscript{\textdagger}\\
    \textsuperscript{\textdagger}Technical University of Munich \qquad
    \textsuperscript{\textdaggerdbl}University of Bonn}
\maketitle

\maketitle
\begin{strip}
    \vspace{-0.9cm}
    \centering
    \captionsetup{type=figure}
    \includegraphics[trim={0cm 24cm 0cm 0cm},clip,width=\linewidth]{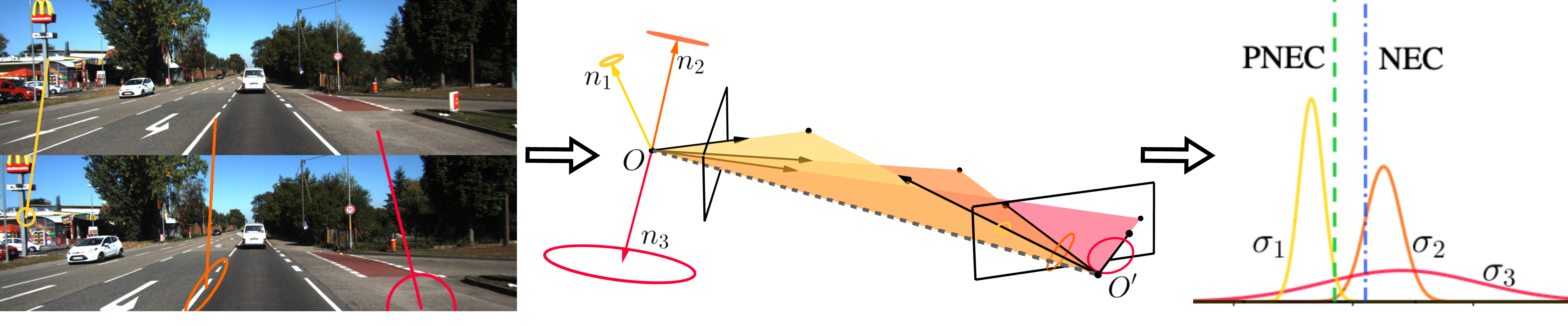}
    \vspace{-0.6cm}
    \captionof{figure}{Illustration of the proposed method. The feature correspondences (\textit{left: lines}) exhibit different position error distributions (\textit{left: ellipses}). In contrast to the \ac{nec}~\cite{OrigNEC_Kneip2012}, our \acf{pnec} correctly accounts for the uncertainty and the geometry of the problem (\textit{middle}) through
        a weighted (\ac{nec}) vs. unweighted (\ac{pnec}) averaging scheme
        (\textit{right}).
        The \ac{nec} weighs all residuals equally (\textit{right: dash-dotted blue line}), so that valuable information is lost. In contrast, our \ac{pnec}  takes into account a proper weighting of individual residuals (\textit{right: dashed green line}).
        Relative pose estimation with the \ac{pnec} instead of the \ac{nec} reduces the rotational-only versions of the $\text{RPE}_1$ and the $\text{RPE}_n$ error by up to $42\%$ and $55\%$ on the KITTI dataset \cite{Geiger2012CVPR}, respectively.
    }
    \label{fig:teaser}
\end{strip}

\begin{abstract}
    The estimation of the relative pose of two camera views is a fundamental problem in computer vision. Kneip et al. proposed to solve this problem by introducing the \acf{nec}.
    However, their approach does not take into account uncertainties, so that the accuracy of the estimated relative pose is highly dependent on accurate feature positions in the target frame.
    In this work, we introduce the \acf{pnec} that overcomes this limitation by accounting for anisotropic and inhomogeneous uncertainties in the feature positions.
    To this end, we propose a novel objective function, along with an efficient optimization scheme that effectively minimizes our objective while maintaining real-time performance. In experiments on synthetic data, we demonstrate that the novel \ac{pnec} yields more accurate rotation estimates than the original \ac{nec} and several popular relative rotation estimation algorithms. Furthermore, we integrate the proposed method into a state-of-the-art monocular rotation-only odometry system and achieve consistently improved results for the real-world KITTI dataset.
\end{abstract}
\vspace{-1cm}

\blfootnote{\textsuperscript{\textasteriskcentered} equal contribution. Project page: \href{https://go.vision.in.tum.de/pnec}{https://go.vision.in.tum.de/pnec}}

\section{Introduction}
Extracting the 3D geometry of a scene from images is a long-standing problem in computer vision and has numerous applications, including augmented and virtual reality, autonomous driving, or robots that can help with everyday life. One key component of many such approaches is the estimation of the relative pose between two viewpoints of a scene.
For example, relative pose estimation is the foundation of geometric vision algorithms like structure from motion (SfM) or visual odometry (VO).
Global SfM pipelines rely on accurate pairwise relative poses for use as fixed measurements in global motion averaging \cite{hartley2013rotation,moulon2013global}.
In VO, relative pose estimation is used to construct a trajectory from a stream of images. Like for all odometry systems, small errors in the relative pose estimation lead to a drift in VO.

The most widely used concept for relative pose estimation is the essential matrix~\cite{longuet1987readings} in the calibrated case, or the fundamental matrix~\cite{Hartley2004} in the general case. Respective approaches rely on correspondences between feature points, and are generally known to provide fast and accurate results~\cite{hartley1997defense}. However, approaches based on the essential matrix suffer from fundamental problems, with the most  prominent being {\em solution multiplicity}~\cite{Faugeras1989mul_solutions, Hartley2004} and
    {\em planar degeneracy} \cite{EigenNEC_Kneip2013}. To address such issues, often it is necessary to consider more involved solution strategies, which also lead to even more accurate relative poses as shown by Kneip~\etal~\cite{EigenNEC_Kneip2013} and in this work.

To this end, Kneip \etal~\cite{OrigNEC_Kneip2012} proposed a constraint that avoids these problems. Their {\em epipolar plane normal coplanarity constraint} (in later works just {\em normal epipolar constraint}, \ac{nec}) allows the estimation of the rotation independent of the translation. A later work by Kneip and Lynen~\cite{EigenNEC_Kneip2013} provides a fast and reliable eigenvalue-based solver for the \ac{nec}, which allows for real-time relative pose estimation. This approach has been incorporated into rotation-only VO systems that estimate the rotation independent of the translation and has led to promising results \cite{MRO_Chng2020, ROBA_Lee2020}.

Yet, like many VO systems, neither of them considers the quality of the correspondences. After removing outliers from the feature matches, every match contributes equally to the final result. However, two-dimensional feature correspondences exhibit different error distributions depending on the content of the image and the specific method used to extract the correspondences, which can be seen in \autoref{fig:kitti_covs}. A correspondence lying on an edge is accurately localized perpendicular to the edge and possesses higher position uncertainty parallel to it. This fine-grained information about the quality of the matches is completely ignored. It has been shown that considering the uncertainty is beneficial for fundamental matrix estimation \cite{pro_cov_Brooks2001}.
While Kanazawa \etal~\cite{con_cov_Kanazawa2001} argue that the uncertainty needs to be sufficiently inhomogeneous to see the aforementioned benefit, our experiments show that the \ac{pnec} improves over the \ac{nec} even for homogeneous uncertainty due to the geometry of the problem.

The main objective of our work is to improve the accuracy of rotation estimation techniques. We achieve this based on the following technical contributions:
\begin{itemize}
    \item We introduce the novel  \acf{pnec}, see~\autoref{fig:teaser}, which for the first time makes it possible to incorporate uncertainty information into the \acf{nec}.
    \item We propose an efficient two-stage optimization strategy for the \ac{pnec} that achieves real-time performance. %
    \item We analyse singularities in the \ac{pnec} energy function and address them with a simple regularization scheme.
    \item Experimentally, we compare our \ac{pnec} to several popular relative pose estimation algorithms, namely 8pt~\cite{hartley1997defense}, 7pt~\cite{Hartley2004}, Stewenius 5pt~\cite{stew5pt}, Nist\'er 5pt~\cite{EM_Nister2003}, and \ac{nec}~\cite{EigenNEC_Kneip2013}, and demonstrate that our \ac{pnec} delivers more accurate rotation estimates. Moreover, we integrate our \ac{pnec} into a visual odometry system and achieve state-of-the-art results on real-world data.
    \item We publish the code for all experiments to facilitate future research.
\end{itemize}

\section{Related Work}
The focus of this paper is the integration of feature position uncertainties into frame-to-frame rotation estimation and the application to visual odometry. Hence, we restrict our discussion of related work to \emph{relative pose estimation}, \emph{uncertainty for feature correspondences}, and \emph{visual odometry}. For a broader overview we refer the reader to the excellent books by Szeliski~\cite{szeliski2010computer} and by Hartley and Zisserman~\cite{Hartley2004} and to more topic-specific overview papers~\cite{triggs1999bundle,cadena2016past}.

\pseudoparagraph{Relative Pose Estimation.}
Estimating the relative pose between two viewpoints is a long-standing problem in computer vision with the first known solution proposed in 1913 by Kruppa~\cite{kruppa1913ermittlung}. Most methods either rely on previously computed feature correspondences (\emph{feature-based}) or directly consider the intensity differences between the two images (\emph{direct}). While direct methods have recently shown promising results~\cite{engel14eccv,DSO_Engel2016}, they are currently limited to images that exhibit photo consistency and hence cannot be used for general problems, e.g. structure from motion. Feature-based methods are considerably more robust to viewpoint and appearance changes. Therefore, we use feature correspondences within this paper.

Given feature correspondences, many methods~\cite{longuet1987readings,EM_Nister2003,stewenius2006recent,li2006five,kukelova2008polynomial} estimate the essential matrix in the case of a calibrated camera, or the fundamental matrix in the general case. Nist\'er~\cite{EM_Nister2003} proposes a minimal solution using polynomials and root bracketing, while the solver proposed by Longuet-Higgins~\cite{longuet1987readings} is linear and requires careful normalization for good performance~\cite{hartley1997defense}. Alternatively, the relative pose can be estimated directly using quaternions~\cite{fathian2018quest}.

The essential matrix constraint deteriorates in zero-translation situations without noise, due to it being a zero matrix. Most essential-matrix-based algorithms estimate the correct motion only implicitly \cite{EigenNEC_Kneip2013}.
To address this problem, recent works have proposed algorithms that can estimate the rotation independent of the translation~\cite{lim2010estimating,OrigNEC_Kneip2012}. Our work is based on the \acf{nec} proposed by Kneip \etal~\cite{OrigNEC_Kneip2012} and the direct optimization scheme proposed in a follow-up paper~\cite{EigenNEC_Kneip2013}. Briales \etal~\cite{Briales2018globalNEC} show how to obtain the global minimum for the \ac{nec}, however, their Shor relaxation is not applicable to our non-polynomial energy function.

\pseudoparagraph{Uncertainty for Feature Correspondences.}
\ac{klt}~tracks~\cite{KLT_Lukas1981,KLT_Tomasi1991} are widely used, and the position uncertainty has been extensively investigated \cite{forstner1987fast,Steele2005Foerster,sheorey2014uncertainty,zhang2017uncertainty}. Based on the unscented transform \cite{uhlmann1995dynamic}, the position uncertainty has also been integrated directly into the \ac{klt} tracking \cite{dorini2011unscented}. Zeisl \etal~\cite{SIFTCOV_Zeisl2009} have shown a method to obtain anisotropic and inhomogeneous covariances for SIFT~\cite{lowe2004distinctive} and SURF~\cite{bay2006surf} features.

\begin{figure}[t]
    \begin{center}
        \adjustbox{trim={.25\width} {.25\height} {0.25\width} {.20\height},clip}%
        {\includegraphics[width=0.94\textwidth]{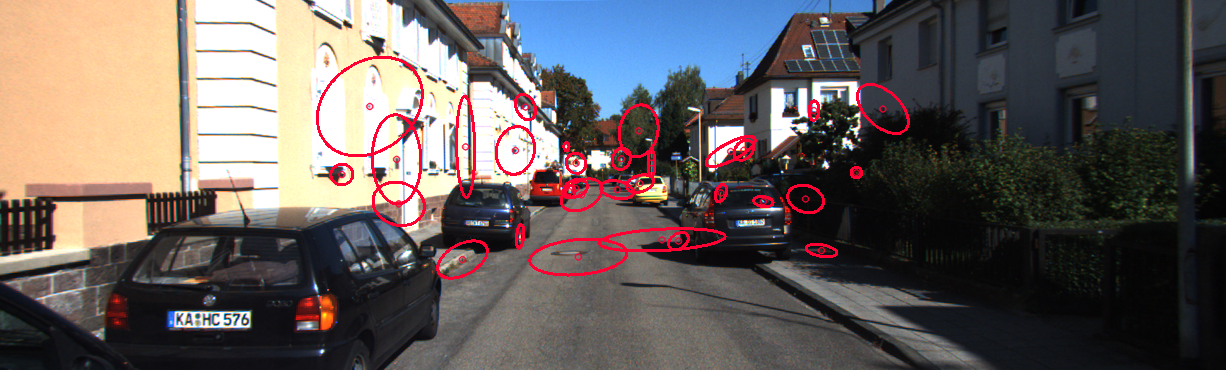}}
    \end{center}
    \caption{Covariance ellipses for position uncertainties in KITTI seq.~07. The tracks are generated with KLT tracking. Our \ac{pnec} correctly considers such anisotropic inhomogeneous error distributions. For visualization purposes only a sub-image with sub-sampled and enlarged covariance ellipses is shown.}
    \label{fig:kitti_covs}
    \vspace{-0.5cm}
\end{figure}

The integration of the position uncertainty into the alignment problem has been studied from a statistical perspective \cite{kanatani2004geometric,kanatani2008statistical}, in the photogrammetry community \cite{meidow2009reasoning}, as well as in the computer vision community \cite{pro_cov_Brooks2001,con_cov_Kanazawa2001}. Brooks \etal~\cite{pro_cov_Brooks2001} show that covariance information can be used beneficially if the estimated covariance is sufficiently accurate. Kanazawa \etal~\cite{con_cov_Kanazawa2001} question the practical use of covariance information if the covariance matrices are too similar and nearly isotropic, however, \autoref{fig:kitti_covs} shows clearly that the covariance matrices for real-world data are highly inhomogeneous and anisotropic.

\pseudoparagraph{Visual Odometry Systems.}
Most VO approaches utilize 3D-to-2D correspondences together with a sliding window formulation. Still, relative pose estimation is commonly used during initialization \cite{ORB_SLAM_Mur-Artal2015} and has been shown to provide excellent results on its own \cite{cvivsic2018soft, Cvisic2021}.
Based on different solutions to the correspondence problem many visual odometry systems have been proposed.
They include PTAM~\cite{Klein2007PTAM} and ORB-SLAM~\cite{ORB_SLAM_Mur-Artal2015, ORB_SLAM2_Mur-Artal2017} with indirect features, approaches with KLT-tracks~\cite{Basalt_Usenko2020}, as well as direct methods like LSD-SLAM~\cite{engel14eccv} and DSO~\cite{DSO_Engel2016}. Common among these approaches is a deteriorating performance for pure rotation without inertial data.

Multiple rotation-only approaches that are robust against pure rotation have been proposed recently \cite{choncha2021vo_init, MRO_Chng2020, ROBA_Lee2020}. Choncha \etal~\cite{choncha2021vo_init} use the \ac{nec} to initialize a scale-consistent map even for purely rotational motion. Chng \etal~\cite{MRO_Chng2020} and Lee and Civera \cite{ROBA_Lee2020} use rotation averaging and rotation-only bundle adjustment on the basis of the \ac{nec} to further improve their results, respectively. While these approaches show promising results on existing datasets, neither of them utilizes uncertainty information of feature correspondences in their system. We base our VO evaluation on MRO~\cite{MRO_Chng2020}.

\section{Probabilistic Normal Epipolar Constraint}
The \acf{nec}~\cite{OrigNEC_Kneip2012} enforces the coplanarity of \emph{epipolar plane normal vectors} constructed from feature correspondences. However, feature correspondences exhibit different error distributions that are not accounted for in the \ac{nec}. For example, an edge-like feature is well-localized perpendicular to the edge but not parallel to it, which is also known as the aperture problem. \autoref{fig:kitti_covs} clearly shows the anisotropicity and inhomogeneity of the position distributions. Moreover, it is well-known that correspondences in all areas of the image are required to sufficiently constrain the 3D geometry \cite{DSO_Engel2016,ORB_SLAM_Mur-Artal2015} and thus we cannot simply discard feature points.
To address this problem, we propose the \acf{pnec}, which is able to take into account the uncertainty of feature positions by associating an anisotropic covariance matrix to each feature point.

\pseudoparagraph{Notation.}
Vectors are denoted by bold lowercase letters (\eg $\boldsymbol{f}$) and matrices by bold uppercase letters (\eg  $\boldsymbol{\Sigma}$). The hat operator applied to a vector $\boldsymbol{u} \in \mathbb{R}^3$  gives a skew-symmetric matrix $\hat{\boldsymbol{u}} \in \mathbb{R}^{3\times 3}$ that computes the cross product between two vectors,~i.e.~$\boldsymbol{u} \times \boldsymbol{v} = \hat{\boldsymbol{u}} \boldsymbol{v}$.
The superscript $^\top$ denotes the transpose. A rigid-body transformation is represented by a rotation matrix $\boldsymbol{R} \in SO(3)$ and a unit length translation $\boldsymbol{t} \in \mathbb{R}^3$ ($\|\boldsymbol{t}\| = 1$ is imposed since the two-view problem is scale-invariant).

\subsection{Background -- NEC}
\label{sec:method_nec}
\begin{figure}[t]
    \begin{center}
        \includegraphics[trim={13cm 12.5cm 17cm 13cm},clip,width=0.478\textwidth]{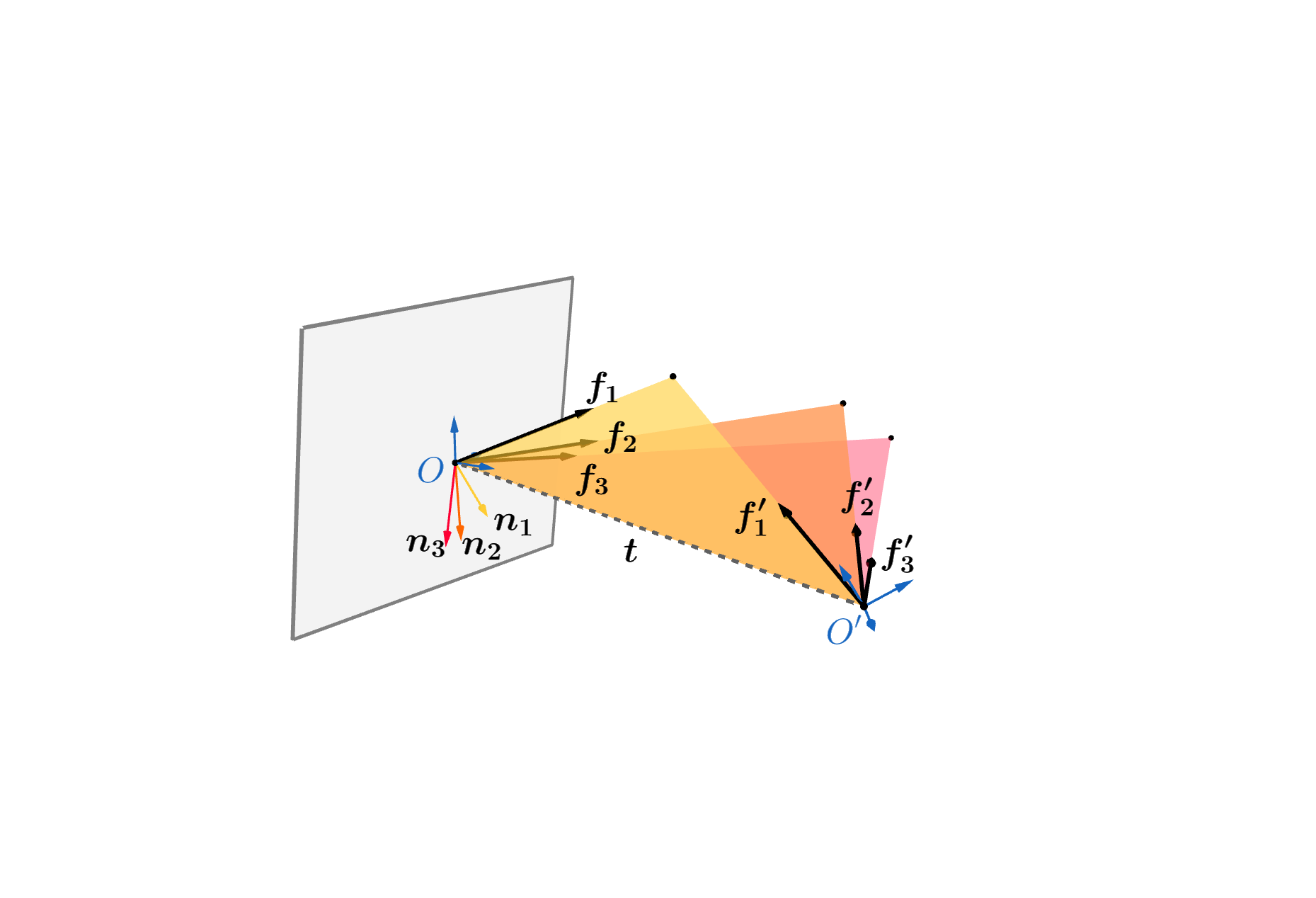}
    \end{center}
    \caption{Geometry of the \acf{nec}~\cite{OrigNEC_Kneip2012}. Feature correspondences are given by pairs of unit bearing vectors $\boldsymbol{f}_i$ and  $\boldsymbol{f}^\prime_i$ in the host frame ($O$) and target frame ($O^\prime$), respectively. Each pair of bearing vectors spans an epipolar plane (\textit{yellow, orange, red}), and has an associated normal vector $\boldsymbol{n}_i$ given in \autoref{eqn:normal_vector}. All epipolar planes intersect in the line defined by the translation $\boldsymbol{t}$ (\textit{dashed line}). The normal vectors span the \epipolarnormalplane{} (\textit{gray}) that is orthogonal to  $\boldsymbol{t}$. For visual clarity we show only three feature correspondences. %
    }
    \label{fig:sketch_nec}
    \vspace{-0.6cm}
\end{figure}
In the following we summarize the main idea of the \ac{nec} proposed in~\cite{OrigNEC_Kneip2012}.
Given are a host frame and a target frame that observe at least five feature correspondences that are defined by pairs of unit bearing vectors $\boldsymbol{f}_i$ and  $\boldsymbol{f}^\prime_i$ in the host and target frame, respectively (see \autoref{fig:sketch_nec}). A 3D point $\f{x}^\prime$ in the target frame is transformed into the host frame by applying the relative rotation $\fR$ and translation $\ft$ s.t. ${\f{x} = \fR \f{x}^\prime + \ft}$. In the ideal, error-free case, a single feature correspondence, together with the two viewpoints, creates an epipolar plane, represented by its normal vector
\begin{equation} \label{eqn:normal_vector}
    \boldsymbol{n}_i = \boldsymbol{f}_i \times \boldsymbol{R} \boldsymbol{f}^\prime_i.
\end{equation}
All normal vectors are orthogonal to the translation and they span the \epipolarnormalplane.

The rotation is estimated by enforcing the coplanarity of the normal vectors. The residual of the model is given by the normalized epipolar error
\begin{equation} \label{eqn:nec-residual}
    e_i = | \boldsymbol{t}^\top\boldsymbol{n}_i | \,,
\end{equation}
i.e.~the Euclidean distance of a normal vector to the \epipolarnormalplane.
An energy function
\begin{equation}
    E(\boldsymbol{R}, \boldsymbol{t}) = \sum_i e_i^2 = \sum_i | \boldsymbol{t}^\top (\boldsymbol{f}_i \times \boldsymbol{R} \boldsymbol{f}^\prime_i) |^2
    \label{eqn:nec-cost}
\end{equation}
is constructed from the residuals. For a more detailed derivation, we kindly refer the reader to the original paper~\cite{OrigNEC_Kneip2012} or the recent paper by Lee and Civera~\cite{lee2020geometric}, which offers numerous geometric interpretations of the \ac{nec}.

\subsection{Deriving the PNEC}
\label{sec:uncertainty}
\label{sec:method_pnec}

The \acf{pnec} extends the \ac{nec} by incorporating uncertainty. To be more specific, the \ac{pnec} allows the use of the anisotropic and inhomogeneous nature of the uncertainty of the feature position in the energy function. The feature position error is considered in the target frame as shown in \autoref{fig:teaser} and we assume that the position error follows a 2D Gaussian distribution in the image plane with a known covariance matrix $\boldsymbol{\Sigma}_{2\text{D},i}$ per feature. In \autoref{sec:klt} we show how the covariance matrix can be extracted for \ac{klt}~tracks from the KLT energy function using Laplace's approximation~\cite{DBLP:journals/jei/BishopN07}.

Given the 2D covariance matrix of the feature position in the target frame $\boldsymbol{\Sigma}_{2\text{D},i}$, we propagate it through the unprojection function using the unscented transform~\cite{uhlmann1995dynamic} in order to obtain the 3D covariance matrix $\boldsymbol{\Sigma}_i$ of the bearing vector $\boldsymbol{f}^\prime_i$.
Using the unscented transform ensures full-rank covariance matrices after the transform. We derive the details of the unscented transform in \autoref{sec:unsc_transform} and show qualitative examples.

Propagating this distribution to the normalized epipolar error gives the probabilistic distribution of the residual. Due to the linearity of the transformations, the distribution of the residual is a univariate Gaussian distribution  $\mathcal{N}(0, \sigma_i^2)$, with variance
\begin{equation} \label{eqn:pnec-weights}
    \sigma_i^2(\fR, \ft) = \boldsymbol{t}^\top \hat{\boldsymbol{f}_i} \boldsymbol{R} \boldsymbol{\Sigma}_i \boldsymbol{R}^\top \hat{\boldsymbol{f}_i}{}^\top \boldsymbol{t} \, .
\end{equation}

We integrate this variance into the cost-function of the \ac{nec} so that the Euclidean distance becomes the Mahalonobis distance~\cite{chandra1936generalised} and define the \textbf{PNEC Energy Function}
\begin{equation} \label{eqn:pnec-cost}
    E_P(\boldsymbol{R}, \boldsymbol{t}) = \sum_i \frac{e_i^2}{\sigma_i^2} = \sum_i \frac{| \boldsymbol{t}^\top (\boldsymbol{f}_i \times \boldsymbol{R} \boldsymbol{f}^\prime_i) |^2}{\boldsymbol{t}^\top \hat{\boldsymbol{f}_i} \boldsymbol{R} \boldsymbol{\Sigma}_i \boldsymbol{R}^\top \hat{\boldsymbol{f}_i}{}^\top \boldsymbol{t}} \,,
\end{equation}
which results in a weighted optimization problem. In \autoref{sec:geometry} we show a geometric interpretation of the above derivation.

\section{Optimization}
To optimize the \ac{pnec} energy function \autoref{eqn:pnec-cost}, we propose a two-stage optimization scheme consisting of an alternating iterative optimization and a joint refinement. We propose this two-stage approach since the eigenvalue-based optimization of the \ac{nec} \cite{EigenNEC_Kneip2013} cannot be naively applied to our derived \ac{pnec} energy function, which we show in \autoref{sec:opt_nec}. The whole PNEC optimization is given in \autoref{algo:optimization_scheme} and we detail the first stage in  \autoref{sec:opt_scf} \& \autoref{sec:opt_rot}, and the second stage in \autoref{sec:opt_joint}.

\subsection{Background -- Optimizing the NEC} \label{sec:opt_nec}
Following~\cite{EigenNEC_Kneip2013}, the \ac{nec} energy function \autoref{eqn:nec-cost} is re-written as ${E(\boldsymbol{R}, \boldsymbol{t}) = \boldsymbol{t}^\top \boldsymbol{M}(\boldsymbol{R}) \boldsymbol{t}}$ using the (symmetric and positive-semidefinite) Gramian matrix
\begin{equation}
    \boldsymbol{M}(\boldsymbol{R}) = \sum_i (\boldsymbol{f}_i \times \boldsymbol{R} \boldsymbol{f}^\prime_i) (\boldsymbol{f}_i \times \boldsymbol{R} \boldsymbol{f}^\prime_i)^\top \,.
\end{equation}
Because the energy is a quadratic form in the unit vector $\boldsymbol{t}$, the optimization over the translation $\boldsymbol{t}$ can be carried out analytically,~i.e.
\begin{equation} \label{eqn:nec-ev-trick}
    \min_{\substack{\boldsymbol{R} \in \text{SO}(3)\\\boldsymbol{t}:\; \|\boldsymbol{t}\| = 1}} \boldsymbol{t}^\top \boldsymbol{M}(\boldsymbol{R}) \boldsymbol{t}\; = \min_{\boldsymbol{R} \in \text{SO}(3)} \operatorname{\lambda}_{\text{min}}(\boldsymbol{M}(\boldsymbol{R})) \, .
\end{equation}
The eigenvector corresponding to the smallest eigenvalue $\lambda
    _{\text{min}}$ of $\boldsymbol{M}(\boldsymbol{R})$ minimizes the Rayleigh quotient $\boldsymbol{t}^\top \boldsymbol{M}(\boldsymbol{R}) \boldsymbol{t}$ over all unit-length vectors $\boldsymbol{t}$.
The constructed sub-problem is then optimized over the rotation $\boldsymbol{R}$ using the Levenberg-Marquardt~algorithm~\cite{levenberg1944method,marquardt1963algorithm}, whereas the translation $\boldsymbol{t}$ is obtained by solving an eigenvalue problem.

\begin{algorithm}[tb]
    \setstretch{1.25}
    \footnotesize
    \SetAlgoLined
    \nl Initialize weights $\tilde{\sigma}_{i,0} \gets 1 \; \forall i$\\
    \For{$s\gets1$ \KwTo $S$}{
    \nl Optimize over $\fR$ (cf.~\autoref{sec:opt_rot})\\
    $\fR_s \gets \Opt_{\fR} \; \operatorname{\lambda}_{\text{min}}(\boldsymbol{M}_P(\fR; \{\tilde{\sigma}_{i,s-1}\}_i))$\\[1.1ex]

    \nl Optimize over $\ft$ (cf.~\autoref{sec:opt_scf})\\
    $\ft_s \gets \Opt_{\ft} \; E_P(\fR_s, \ft)$\\[1.1ex]

    \nl Update the weights (cf.~\autoref{eqn:pnec-weights}) \\
    $\tilde{\sigma}_{i,s} \gets \sigma_i(\boldsymbol{R}_s, \boldsymbol{t}_s)  \; \forall i$
    }

    \nl Joint Refinement (cf.~\autoref{sec:opt_joint}) using $(\fR_S, \ft_S)$ as starting value\\
    $\fR^\ast, \ft^\ast \gets \Opt_{\fR, \ft} E_P(\fR, \ft)$
    \caption{\ac{pnec} Optimization Scheme}
    \label{algo:optimization_scheme}
\end{algorithm}

\subsection{Optimizing the PNEC - Translation}\label{sec:opt_scf}
The \ac{pnec} energy function \autoref{eqn:pnec-cost} is the sum of \acp{grq} in the translation $\boldsymbol{t}$, and thus the optimum is not simply given by an eigenvalue as for the \ac{nec}. Optimizing the sum of \acp{grq} over the unit sphere has recently been studied in the context of data science and wireless communications \cite{GRQZhang1, GRQZhang2, binbuhaer2019optimizing}, and it has been shown by Zhang~\etal~\cite{GRQZhang2} that the \ac{scf} algorithm~\cite{hartree1928wave} outperforms generic manifold optimization methods.

Since the sum of \acp{grq} can exhibit many local minima~\cite{binbuhaer2019optimizing}, and thus the \ac{scf}~iteration is not guaranteed to converge to a global optimum, we propose a simple, yet effective globalization strategy.
To this end, we make use of the intrinsic low dimension of the unit sphere in $\mathbb{R}^3$ by sampling evenly distributed initial points $\boldsymbol{t}_k$ efficiently using the Fibonacci lattice~\cite{gonzalez2010measurement}. We then pick the point with the lowest objective function and apply the \ac{scf} iteration for $N$ steps. Due to the inherent parallelism, the resulting optimization procedure can be implemented efficiently. We present the effectiveness of the globalization strategy, as well as technical details for the \ac{scf} iteration, in the supplementary material in \autoref{sec:scf} and \autoref{tab:hyperparameter-study}.

\subsection{Optimizing the PNEC - Rotation}\label{sec:opt_rot}
Kneip and Lynen~\cite{EigenNEC_Kneip2013} have shown how to optimize \autoref{eqn:nec-ev-trick} efficiently using the Levenberg-Marquardt~algorithm with the rotation parametrized based on the Cayley transformation \cite{cayley1846algebraic}.
To account for the weights in the \ac{pnec} energy function \autoref{eqn:pnec-cost}, we employ an optimization scheme similar to the popular \ac{irls} algorithm~\cite{lawson1961contribution}. Specifically, given a previous estimate of the rotation and translation $(\fR_p, \ft_p)$, we compute fixed weights $\tilde{\sigma}_i = \sigma_i(\fR_p, \ft_p)$ for all $i$, and define the weighted  matrix
\begin{equation} \label{eqn:MP}
    \boldsymbol{M}_P(\boldsymbol{R}; \{\tilde{\sigma}_i\}_i) = \sum_i \frac{(\boldsymbol{f}_i \times \boldsymbol{R} \boldsymbol{f}^\prime_i) (\boldsymbol{f}_i \times \boldsymbol{R} \boldsymbol{f}^\prime_i)^\top}{\tilde{\sigma}_i^2}
\end{equation}
that depends only on the rotation $\fR$. The rotation is obtained by finding $\boldsymbol{R}$ such that the smallest eigenvalue of $\boldsymbol{M}_P(\boldsymbol{R}; \{\tilde{\sigma}_i\}_i)$ is minimal. After doing so based on the optimizer of Kneip and Lynen~\cite{EigenNEC_Kneip2013}, the weights~$\{\tilde{\sigma}_i\}_i$ are updated with new $\boldsymbol{R}, \ft$.

\subsection{Optimizing the PNEC - Joint Refinement} \label{sec:opt_joint}
After the first stage we improve the result using joint refinement. Specifically, we use a least-squares optimization strategy, which is effective for finding a local optimum of the energy function given a good starting point \cite{DBLP:books/sp/NocedalW99}. For the \ac{pnec} we optimize over
\begin{equation}
    E_P(\boldsymbol{R}, \boldsymbol{t}) = \sum_i \left(\frac{\boldsymbol{t}^\top (\boldsymbol{f}_i \times \boldsymbol{R} \boldsymbol{f}^\prime_i)}{\sqrt{\boldsymbol{t}^\top \hat{\boldsymbol{f}_i} \boldsymbol{R} \boldsymbol{\Sigma}_i \boldsymbol{R}^\top \hat{\boldsymbol{f}_i}{}^\top \boldsymbol{t}}}\right)^{2} \,,
\end{equation}
the least-squares formulation of the constraint. The Levenberg-Marquardt~algorithm optimizes the objective function in the rotation $\boldsymbol{R}$ and translation $\boldsymbol{t}$ simultaneously and uses the solution of the first stage as the starting value.
Because $\boldsymbol{R}$ is a rotation matrix, we use manifold optimization \cite{DBLP:journals/inffus/HertzbergWFS13} to optimize over the special orthogonal group $\text{SO}(3)$. For the translation $\boldsymbol{t}$ we use spherical coordinates with the radius fixed to one in order to ensure that $\|\boldsymbol{t}\| = 1$ holds. We would like to highlight that this joint refinement is different from bundle adjustment. Most notably, it does not need to calculate the 3D position of the features.

\subsection{Singularities of the \ac{pnec}} \label{sec:singularities}
\begin{figure}[t]
    \begin{center}
        \includegraphics[width=0.475\textwidth]{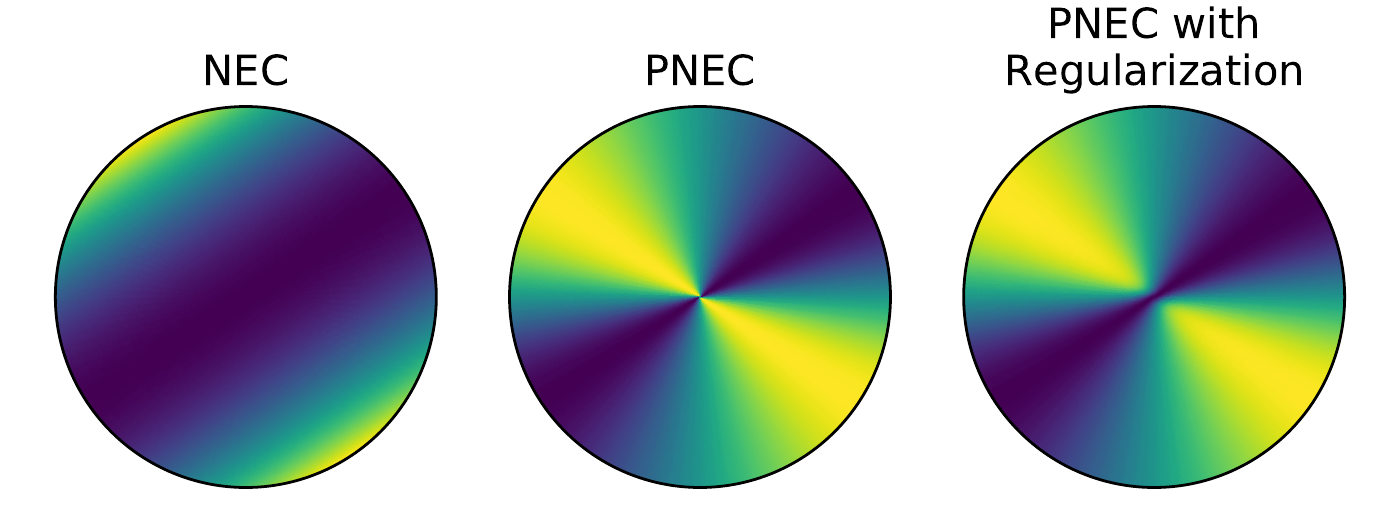}
    \end{center}
    \caption{Visualization of the \ac{nec}, the \ac{pnec}, and the \ac{pnec} with the regularization proposed in \autoref{sec:singularities}. The plot shows $\boldsymbol{t}$ in a neighborhood of $\boldsymbol{f}$ (in polar coordinates), where the center of the circle corresponds to $\boldsymbol{t} = \boldsymbol{f}$. For $\boldsymbol{t} = \boldsymbol{f}$, the \ac{pnec} shows a finite discontinuity, for which the limit depends on the direction. Our regularization eliminates this singularity while also maintaining the overall shape of the energy function.}
    \label{fig:nec_pnec_polar_plot}
\end{figure}
The \ac{pnec} energy function \autoref{eqn:pnec-cost} has a singularity if the translation $\boldsymbol{t}$ is parallel to a bearing vector $\boldsymbol{f}_i$ because the variance $\sigma_i^2$ vanishes due to $\hat{\boldsymbol{f}_i} \boldsymbol{t} = \boldsymbol{f}_i \times \ft = \boldsymbol{0}$. On the other hand, the numerator involves the same term and thus the energy function is bounded and possesses a finite discontinuity, as illustrated in~\autoref{fig:nec_pnec_polar_plot}. In \autoref{sec:limit} we present the derivation of the directional limit of the energy function.

While the discontinuity is finite and less problematic than an infinite discontinuity, it still poses challenges.
First, in contrast to the function values, the derivatives of the energy function are not bounded, which is problematic for the joint refinement.
Second, the matrix $\boldsymbol{M}_P$, unlike the energy function, includes $\hat{\boldsymbol{f}_i} \boldsymbol{t}$ only in its denominator not the numerator. Hence $\boldsymbol{M}_P$ tends to infinity for $\boldsymbol{t} \to \boldsymbol{f}_i$.
To address these issues, we consider a variance of the form $\sigma_{i}^{\prime}{}^2 = \sigma_{i}^2 + c$ with regularization constant $c > 0$. \autoref{fig:nec_pnec_polar_plot} shows the effect that the regularizer has on the energy function.

\begin{figure*}[t]
    \centering
    \begin{subfigure}[b]{0.32\textwidth}
        \centering
        {\includegraphics[width=\textwidth]{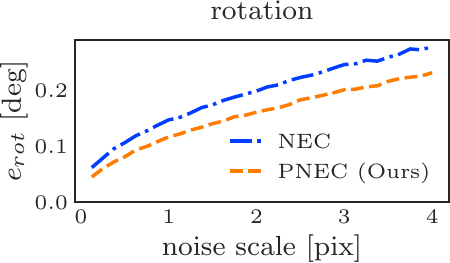}}
        \caption{Experiment w/ translation: $e_{\text{rot}}$}
        \label{fig:omni_rot_t}
    \end{subfigure}
    \hfill
    \begin{subfigure}[b]{0.31\textwidth}
        \centering
        {\includegraphics[width=\textwidth]{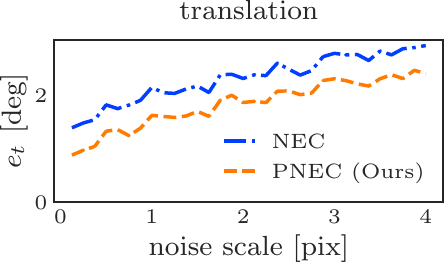}}
        \caption{Experiment w/ translation: $e_{\text{t}}$}
        \label{fig:omni_t}
    \end{subfigure}
    \hfill
    \begin{subfigure}[b]{0.32\textwidth}
        \centering
        {\includegraphics[width=\textwidth]{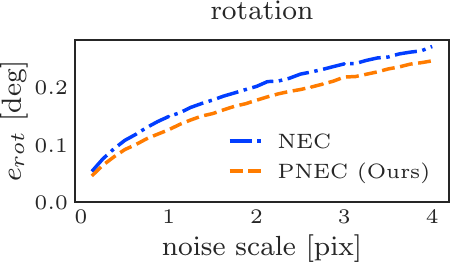}}
        \caption{Experiment w/o translation: $e_{\text{rot}}$}
        \label{fig:omni_no_t}
    \end{subfigure}
    \caption{Experiments for omnidirectional cameras. Results are averaged over 10\,000 random instantiations for anisotropic inhomogeneous noise over different noise intensities. Our \ac{pnec} consistently leads to smaller errors compared to the \ac{nec}~\cite{EigenNEC_Kneip2013} for all noise levels. This holds for rotation and translation estimates in the general case in \autoref{fig:omni_rot_t} and \autoref{fig:omni_t}, respectively, as well as the rotation in the zero-translation case in \autoref{fig:omni_no_t}.
    }
    \label{fig:omni}
\end{figure*}
\begin{figure*}[t]
    \centering
    \begin{subfigure}[b]{0.32\textwidth}
        \centering
        {\includegraphics[width=\textwidth]{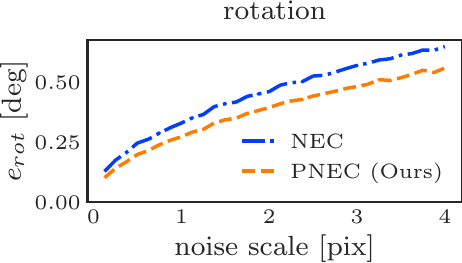}}
        \caption{Experiment w/ translation: $e_{\text{rot}}$}
        \label{fig:pin_rot_t}
    \end{subfigure}
    \hfill
    \begin{subfigure}[b]{0.31\textwidth}
        \centering
        {\includegraphics[width=\textwidth]{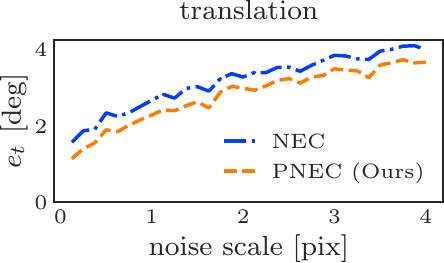}}
        \caption{Experiment w/ translation: $e_{\text{t}}$}
        \label{fig:pin_t}
    \end{subfigure}
    \hfill
    \begin{subfigure}[b]{0.32\textwidth}
        \centering
        {\includegraphics[width=\textwidth]{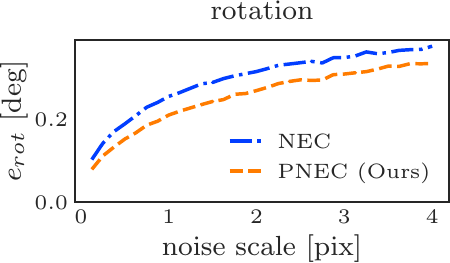}}
        \caption{Experiment w/o translation: $e_{\text{rot}}$}
        \label{fig:pin_no_t}
    \end{subfigure}
    \caption{Experiments for pinhole cameras. Results are averaged over 10\,000 random instantiations for anisotropic inhomogeneous noise over different noise intensities. As for an omnidirectional camera in \autoref{fig:omni}, our \ac{pnec} consistently leads to smaller errors compared to the \ac{nec}~\cite{EigenNEC_Kneip2013}. For a pinhole camera the average errors are higher for both methods in comparison to an omnidirectional camera. The experiments show that our \ac{pnec} is viable for the two most common camera types.}
    \label{fig:pin}
\end{figure*}
\section{Evaluation}
We evaluate the performance of the \ac{pnec} and compare it to the original \ac{nec} on simulated data as well as in a visual odometry setting on real world data. On the simulated data, the proposed \ac{pnec} achieves better results than the \ac{nec} and several other popular relative pose estimation algorithms. On KITTI we compare our approach to the MRO algorithm~\cite{MRO_Chng2020} that uses the \ac{nec} for rotation estimation. For evaluating the \ac{pnec} we replace the ORB features in MRO with \acf{klt}~tracks~\cite{KLT_Lukas1981,KLT_Tomasi1991}, which allow for uncertainty extraction as discussed in \autoref{sec:uncertainty}, and the \ac{nec} with the \ac{pnec}.

In \autoref{sec:simulated}, a more detailed analysis including translational errors shows that compared to the NEC the \ac{pnec} is not only significantly more accurate on average, but also more consistent. An ablation study on KITTI (cf.~\autoref{sec:kitti}) demonstrates that all stages of our optimization scheme are essential for best results. We furthermore detail all hyperparameter choice and detail the experimental protocol for reproducibility (cf~ \autoref{sec:Hyperparameters}).

\subsection{Frame-to-Frame Simulation}
\begin{table*}[t]
    \small
    \centering
    \sisetup{detect-weight,mode=text}
    \renewrobustcmd{\bfseries}{\fontseries{b}\selectfont}
    \renewrobustcmd{\boldmath}{}
    \newrobustcmd{\B}{\bfseries}
    \addtolength{\tabcolsep}{-3.5pt}
    \begin{tabular} {p{2.4cm} r r  r r  r r|r  r  r|r r  r r  r r|r  r  r  r}
        \toprule
                                                             & \multicolumn{9}{c}{\scshape Omnidirectional} & \multicolumn{9}{c}{\scshape Pinhole} &                                                                                                                                                                                                                                                                                                                                                                                                                           \\[1.1ex]
                                                             & \multicolumn{6}{c}{\scshape w/ t}            & \multicolumn{3}{c}{\scshape w/o t}   & \multicolumn{6}{c}{\scshape w/ t} & \multicolumn{3}{c}{\scshape w/o t} &                                                                                                                                                                                                                                                                                                                                                  \\
        Noise level [px]                                     &
        \multicolumn{2}{c}{\scshape 0.5}                     & \multicolumn{2}{c}{\scshape 1.0}             & \multicolumn{2}{c}{\scshape 1.5}     & {\scshape 0.5}                    & {\scshape 1.0}                     & {\scshape 1.5}     & \multicolumn{2}{c}{\scshape 0.5} & \multicolumn{2}{c}{\scshape 1.0} & \multicolumn{2}{c}{\scshape 1.5} &
        {\scshape 0.5}                                       & {\scshape 1.0}                               & {\scshape 1.5}                                                                                                                                                                                                                                                                                                                                                                                                                                                   \\
        Metric [degree]                                      & {$e_{\text{rot}}$}                           & {$e_{\text{t}}$}                     & {$e_{\text{rot}}$}                & {$e_{\text{t}}$}                   & {$e_{\text{rot}}$} & {$e_{\text{t}}$}                 & {$e_{\text{rot}}$}               & {$e_{\text{rot}}$}               & {$e_{\text{rot}}$} & {$e_{\text{rot}}$} & {$e_{\text{t}}$} & {$e_{\text{rot}}$} & {$e_{\text{t}}$} & {$e_{\text{rot}}$} & {$e_{\text{t}}$} & {$e_{\text{rot}}$} & {$e_{\text{rot}}$} & {$e_{\text{rot}}$} &        \\

        \midrule

        7pt \cite{Hartley2004}/8pt \cite{hartley1997defense} & 0.19                                         & \underline{1.76}                     & 0.26                              & 2.36                               & 0.33               & 2.64                             & \underline{0.10}                 & \underline{0.15}                 & 0.20               & 0.62               & 4.64             & 0.89               & 6.09             & 1.07               & 6.74             & \underline{0.17}   & \underline{0.23}   & \underline{0.28}   & +95\%  \\
        Stewenius 5pt \cite{stew5pt}                         & 0.23                                         & 2.34                                 & 0.30                              & 2.97                               & 0.37               & 3.38                             & 0.18                             & 0.30                             & 0.33               & 0.61               & 3.14             & 0.71               & 4.04             & 0.89               & 4.57             & 0.46               & 0.64               & 0.77               & +137\% \\
        Nist\'er 5pt \cite{EM_Nister2003}                    & 1.61                                         & 6.82                                 & 1.64                              & 7.55                               & 1.92               & 8.42                             & 0.29                             & 0.39                             & 0.42               & 3.26               & 8.72             & 3.46               & 9.80             & 3.76               & 10.39            & 0.51               & 0.67               & 0.83               & +643\% \\
        NEC  \cite{EigenNEC_Kneip2013}                       & \underline{0.11}                             & 1.90                                 & \underline{0.15}                  & \underline{2.10}                   & \underline{0.17}   & \underline{2.11}                 & 0.11                             & \underline{0.15}                 & \underline{0.18}   & \underline{0.25}   & \underline{2.41} & \underline{0.34}   & \underline{2.78} & \underline{0.41}   & \underline{2.91} & 0.19               & 0.25               & 0.29               & +24\%  \\
        PNEC  (Ours)                                         & \bfseries 0.08                               & \bfseries 1.29                       & \bfseries 0.12                    & \bfseries 1.60                     & \bfseries 0.14     & \bfseries 1.66                   & \bfseries 0.09                   & \bfseries 0.13                   & \bfseries 0.15     & \bfseries 0.20     & \bfseries 2.06   & \bfseries 0.28     & \bfseries 2.38   & \bfseries 0.34     & \bfseries 2.54   & \bfseries 0.15     & \bfseries 0.21     & \bfseries 0.25     &        \\

        \bottomrule
    \end{tabular}
    \caption{Rotation and translation error for different algorithms. Results for omnidirectional and pinhole cameras for experiments with and without translation over different noise levels for anisotropic and inhomogenous noise. Errors are averaged over 10\,000 random problems each with 10 points. For experiments without translation (W/O T) only $e_{\text{rot}}$ is reported, due to $e_{\text{t}}$ not being defined for zero translation. For all other algorithms apart from our \ac{pnec} we use the implementations from OpenGV \cite{Kneip2014opengv}. (7pt) falls back to (8pt) for the non-minimal number of 10 correspondences. Our \ac{pnec} consistently achieves the best results, outperforming the \ac{nec} and several popular relative pose estimation algorithms. The last column gives the average error increase compared to the \ac{pnec}.
    }
    \label{tab:further_results}
\end{table*}
With the simulated experiments we evaluate the performance of the \ac{pnec} in a frame-to-frame setting. The experiments consist of randomly generated problems of two frames with known correspondences.
We use
\begin{align}
    e_{\text{rot}} & := \angle (\boldsymbol{R}^\top \Tilde{\boldsymbol{R}}),~\text{and} \\
    e_{\text{t}}   & := \arccos{(\boldsymbol{t}^\top \Tilde{\boldsymbol{t}})}
\end{align}
as error metrics between the ground truth $\boldsymbol{R}, \boldsymbol{t}$ and the estimated values $\Tilde{\boldsymbol{R}}, \Tilde{\boldsymbol{t}}$, where $\angle(\cdot)$ returns the angle of the rotation matrix.

\pseudoparagraph{Omnidirectional Camera.}
In this experiment we follow the experimental outline proposed by Kneip and Lynen~\cite{EigenNEC_Kneip2013} closely. We differ from the original experiments in the following ways: we only add noise to the points in the second frame; to compensate for the lack of noise in the first frame, we scale the standard deviation by a factor of $2$; we recreate the experiment with different noise types based on the classification by Brooks \etal~\cite{pro_cov_Brooks2001} and generate individual covariance matrices for each point. A detailed description of how the matrices are generated can be found in \autoref{sec:simulated}. To show the effectiveness of the \ac{pnec} still holds even for pure rotation, we repeat the experiment with the translational difference fixed to zero.

\autoref{fig:omni} shows the results for anisotropic inhomogeneous noise for both experiments. The \ac{pnec} achieves consistently better results for the rotation over all noise levels in both experiments.

\pseudoparagraph{Pinhole Camera.}
Since most cameras are modeled as pinhole cameras we also repeat the previous experiments for pinhole cameras. The generation of the frames stays the same. Points are sampled in viewing direction of the coordinate system of the first frame. The points are projected into the world coordinate system and then into the two frames using a pinhole camera model. The noise offset is added in the image plane. As with the omnidirectional camera experiment, we repeat this experiment for pure rotation.

\autoref{fig:pin} shows the results for the pinhole camera experiment for anisotropic inhomogeneous noise. While the overall error for both methods is slightly higher for pinhole cameras than for omnidirectional cameras, the \ac{pnec} still outperforms the \ac{nec} consistently.

\autoref{tab:further_results} gives a quantitative comparison of our \ac{pnec}
to other relative pose estimation algorithms from the literature on the experiments presented in \autoref{fig:omni} and \autoref{fig:pin}. Our \ac{pnec} consistently achieves the best result for both camera models for experiments with and without translation.

Additional experiments on other noise types show that our \ac{pnec} outperforms the \ac{nec} even in cases of isotropic and homogeneous noise. Although the covariance matrices are identical, the variance for each residual is different due to the geometry of the problem.  Furthermore, our \ac{pnec} is robust against wrongly estimated noise parameters. We present the results to these experiments in the \autoref{sec:simulated} - \autoref{sec:offset}.

\subsection{Visual Odometry}
Besides the simulated experiments, we also validate the \ac{pnec} on real world data, namely the highly popular KITTI odometry dataset~\cite{Geiger2012CVPR}. We compare our results with the MRO algorithm by Chng \etal~\cite{MRO_Chng2020} that uses the optimization from \cite{EigenNEC_Kneip2013}. For MRO and our algorithm we disable rotation averaging and loop closure to focus on local rotation estimation. Our approach differs from MRO in two ways.

First, we use the KLT-based tracking implementation also used in~\cite{Basalt_Usenko2020} to extract feature keypoints instead of ORB features. Second, we replace the \ac{nec} with our \ac{pnec} for relative rotation estimation. %
To capture the effects of both changes, we compare the rotation estimation of MRO, as reported in \cite{MRO_Chng2020}, KLT-\ac{nec}, using KLT tracks and the \ac{nec}, and KLT-\ac{pnec}, the proposed \ac{pnec} with KLT tracks. Both KLT-\ac{nec} and KLT-\ac{pnec} use the same KLT tracks for the relative rotation estimation.

The proposed \ac{pnec} can account for uncertainties in the feature correspondence positions that approximately follow a Gaussian distribution. To overcome outlier correspondences from failed \ac{klt} tracks, we use the same RANSAC~\cite{fischler1981random} routine as the \ac{nec} for estimating the rotation in the first loop of \autoref{algo:optimization_scheme}.

\begin{figure}[t]
    \begin{center}
        \includegraphics[trim={0cm 0.0cm 0cm 0cm},clip,width=0.47\textwidth]{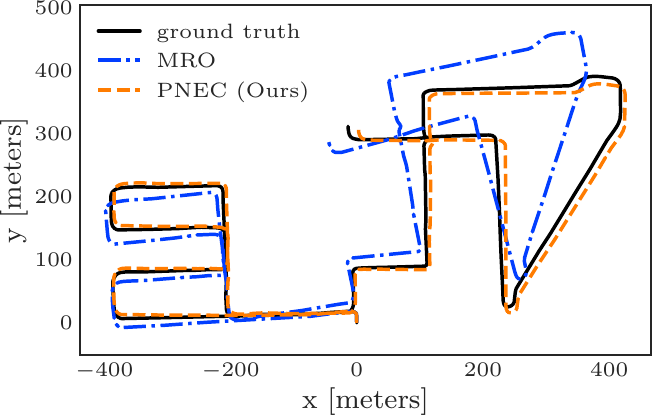}
        \vspace{-0.6cm}
    \end{center}
    \caption{Qualitative trajectory comparison for KITTI seq.~08. The trajectory was generated with the estimated rotations of MRO~\cite{MRO_Chng2020} and PNEC, respectively, and are combined with the ground truth translations for visualization purposes. Relative rotations computed with the proposed \ac{pnec} lead to a significantly reduced drift.}
    \vspace{-0.2cm}
    \label{fig:kitti_trajectory}
\end{figure}

\autoref{fig:kitti_trajectory} shows a trajectory generated from the rotation estimates of MRO and our approach. In \autoref{tab:kitti} we compare the mean performance over 5 runs of all approaches in the rotation-only version of the {\em Relative Pose Error} ($\text{RPE}$) for $n$ camera poses as defined in~\cite{MRO_Chng2020}. The $\text{RPE}$ evaluates the root mean square error ($\text{RMSE}$) of rotational residuals over frame pairs. The residual for a ``time-step" $\Delta$ is
\begin{equation}
    E_i := \angle ((\boldsymbol{R}^\top_{i} \boldsymbol{R}_{i + \Delta})^\top (\Tilde{\boldsymbol{R}}{}^\top_{i} \Tilde{\boldsymbol{R}}_{i + \Delta})).
\end{equation}
The $\text{RMSE}$ is calculated over $m:=n - \Delta$ residuals
\begin{equation}
    \text{RMSE}(\Delta) := \left(\frac{1}{m} \sum_{i=1}^{m} E_i^2 \right)^{\frac{1}{2}}.
\end{equation}
For our evaluation we use
\begin{align}
    \text{RPE}_1 & := \text{RMSE}(1),~\text{and}                             \\
    \text{RPE}_n & := \frac{1}{n} \sum_{\Delta = 1}^{n} \text{RMSE}(\Delta),
\end{align}
to capture local frame-to-frame rotation error and long term drift, respectively.

\begin{table}[t]
    \small
    \centering
    \sisetup{detect-weight,mode=text}
    \renewrobustcmd{\bfseries}{\fontseries{b}\selectfont}
    \renewrobustcmd{\boldmath}{}
    \newrobustcmd{\B}{\bfseries}
    \addtolength{\tabcolsep}{-3.5pt}

    \begin{tabular} {p{1.6cm} r r|r r|r r}
        \toprule
                                  & \multicolumn{2}{c}{\scshape MRO~\cite{MRO_Chng2020}} & \multicolumn{2}{c}{\scshape KLT-\ac{nec}} & \multicolumn{2}{c}{\scshape KLT-\ac{pnec}}                                                              \\
                                  & \multicolumn{2}{c}{}                                 & \multicolumn{2}{c}{}                      & \multicolumn{2}{c}{(Ours)}                                                                              \\
        Seq.                      &
        {\scshape $\text{RPE}_1$} & {\scshape $\text{RPE}_n$}                            &
        {\scshape $\text{RPE}_1$} & {\scshape $\text{RPE}_n$}                            &
        {\scshape $\text{RPE}_1$} & {\scshape $\text{RPE}_n$}                                                                                                                                                                                  \\
        \midrule
        00                        & 0.360                                                & 8.670                                     & \underline{0.125}                          & \underline{5.922} & \B0.119           & \B3.429            \\
        01*                       & \B0.290                                              & \B16.030                                  & \underline{0.695}                          & 27.406            & 0.782             & \underline{23.500} \\
        02                        & 0.290                                                & 16.030                                    & \B0.093                                    & \B6.693           & \underline{0.122} & \underline{9.687}  \\
        03                        & 0.280                                                & 5.470                                     & \underline{0.073}                          & \underline{2.728} & \B0.059           & \B1.411            \\
        04                        & \underline{0.040}                                    & 1.080                                     & 0.041                                      & \underline{0.619} & \B0.038           & \B0.463            \\
        05                        & 0.250                                                & 11.360                                    & \underline{0.079}                          & \underline{4.489} & \B0.070           & \B3.203            \\
        06                        & 0.180                                                & 4.720                                     & \underline{0.073}                          & \underline{3.162} & \B0.042           & \B2.322            \\
        07                        & 0.280                                                & 7.490                                     & \underline{0.105}                          & \underline{4.640} & \B0.074           & \B2.065            \\
        08                        & 0.270                                                & 9.210                                     & \underline{0.070}                          & \underline{5.523} & \B0.060           & \B3.347            \\
        09                        & 0.280                                                & 9.850                                     & \underline{0.088}                          & \underline{3.533} & \B0.080           & \B3.514            \\
        10                        & 0.380                                                & 13.250                                    & \underline{0.073}                          & \B3.959           & \B0.072           & \underline{4.094}  \\
        \bottomrule
    \end{tabular}
    \vspace{-0.25cm}
    \caption{Quantitative comparison for KITTI. The significant gap between MRO and KLT-NEC confirms the benefit of using KLT tracks. The difference between KLT-NEC and KLT-\ac{pnec} shows the effectiveness of our \ac{pnec} compared to the \ac{nec}.
        *In seq.~01 the KLT implementation of \cite{Basalt_Usenko2020} fails and produces many wrong tracks with incorrect covariances due to self-similar structure.}
    \label{tab:kitti}
    \vspace{-0.6cm}
\end{table}
The results show the following: With a single exception (seq. 01), using KLT tracks instead of ORB features is beneficial for relative rotation estimation with the \ac{nec}. \ac{pnec} outperforms \ac{nec} on 8 out of 11 sequences in both metrics, often significantly. Excluding seq. 01, the \ac{pnec} on average improves the $\text{RPE}_1$ for frame-to-frame rotational error by $10\%$ and the $\text{RPE}_n$ for long term drift by $19\%$.

\subsection{Runtime}
\label{sec:runtime}

\autoref{tab:runtime} shows the average frame processing time on the KITTI dataset for MRO, KLT-NEC and KLT-PNEC. The experiments were performed on a laptop with a 2.4 GHz Quad-Core Intel Core i5 processor and 8 GB of memory. For MRO we use the same configuration as for their demo.
The results show the runtime advantage of KLT tracks that do not need feature matching like ORB features. While the proposed optimization scheme for the \ac{pnec} is slightly slower than the simpler \ac{nec} optimization algorithm, the odometry with KLT-\ac{pnec} runs in real-time on KITTI.

\begin{table}[t]
    \small
    \centering
    \sisetup{detect-weight,mode=text}
    \renewrobustcmd{\bfseries}{\fontseries{b}\selectfont}
    \renewrobustcmd{\boldmath}{}
    \newrobustcmd{\B}{\bfseries}
    \addtolength{\tabcolsep}{-3.5pt}

    \begin{tabular} {p{3.0cm} r|r|r}
        \toprule
                         & {\scshape MRO~\cite{MRO_Chng2020}} & {\scshape KLT-NEC} & {\scshape KLT-\ac{pnec}} \\
        \midrule
        feature creation & 36                                 & 30                 & 30                       \\
        matching         & 120                                &                    &                          \\
        optimization     & 5                                  & 23                 & 47                       \\
        \midrule
        total time       & 161                                & 53                 & 77                       \\
        \bottomrule
    \end{tabular}
    \caption{Average frame processing time in milliseconds. For MRO, most of the time is needed for matching. KLT-NEC and KLT-\ac{pnec} (Ours) achieve real-time performance on KITTI.}
    \label{tab:runtime}
    \vspace{-0.5cm}
\end{table}

\section{Discussion and Future Work}
While the proposed optimization scheme effectively optimizes the \ac{pnec} energy function, it relies on two consecutive stages, and is thus more involved than the optimization scheme proposed for the \ac{nec}~\cite{EigenNEC_Kneip2013}.
Further and more detailed limitations of the proposed approach are given in \autoref{sec:limitations}.
Nevertheless, we have shown in \autoref{sec:runtime} that the proposed algorithm is real-time capable. As we explain in \autoref{sec:opt_scf}, the optimization over the translation alone is an actively studied problem for which no simple solution is known. Nevertheless, investigating improved optimization schemes for our \ac{pnec} energy function is  a promising direction for future work. Recent works have shown that deep learning can boost the performance of visual odometry algorithms \cite{DBLP:conf/eccv/YangWSC18,DBLP:conf/cvpr/YangSWC20,DBLP:conf/icra/GreeneR20}. However, the focus of our work is on the correct modelling of the uncertainty for relative pose estimation, similar to \cite{OrigNEC_Kneip2012,EigenNEC_Kneip2013}. As such, while in our work we do not consider deep learning, we believe that the integration of our ideas into learning systems may be an interesting direction for follow-up works. For example, the 2D feature position covariance matrices could be predicted by a neural network.

\section{Conclusion}
This paper shows how to utilise 2D feature position uncertainties to obtain more accurate relative pose estimates from a pair of images. To this end, we introduce the \acf{pnec}, and we propose an effective optimization scheme that runs in real-time. In synthetic experiments, the \ac{pnec} gives more accurate rotation estimates than the \ac{nec} and several popular relative rotation estimation algorithms for different noise levels and for the pure-rotation case. The results on KITTI show, that the relative rotation estimation of the \ac{pnec} improves upon the \ac{nec}-based MRO, a state-of-the-art rotation-only VO system, and can be used \eg for global initialization in SfM.
\vspace{-0.75cm}
\subsubsection*{Acknowledgements}
\vspace{-0.2cm}
We express our appreciation to our colleagues and the WARP student team who have supported us.
We thank Laurent Kneip for providing us with the suppl.\ for \cite{EigenNEC_Kneip2013}. This work was supported by the ERC Advanced Grant SIMULACRON and by the Munich Center for Machine Learning.
\clearpage

{\small
    \bibliographystyle{ieee_fullname}
    \bibliography{egbib}

\begin{thebibliography}{10}\itemsep=-1pt

\bibitem{DBLP:journals/ijcv/BakerM04}
Simon Baker and Iain~A. Matthews.
\newblock Lucas-kanade 20 years on: {A} unifying framework.
\newblock {\em IJCV}, 56, 2004.

\bibitem{bay2006surf}
Herbert Bay, Tinne Tuytelaars, and Luc Van~Gool.
\newblock Surf: Speeded up robust features.
\newblock In {\em ECCV}, 2006.

\bibitem{binbuhaer2019optimizing}
Aohud~Abdulrahman Binbuhaer.
\newblock {\em On Optimizing the Sum of Rayleigh Quotients on the Unit Sphere}.
\newblock PhD thesis, The University of Texas at Arlington, 2019.

\bibitem{DBLP:journals/jei/BishopN07}
Christopher~M. Bishop.
\newblock {\em Pattern recognition and machine learning, 5th Edition}.
\newblock Springer, 2007.

\bibitem{Briales2018globalNEC}
Jesus Briales, Laurent Kneip, and Javier Gonzalez-Jimenez.
\newblock A certifiably globally optimal solution to the non-minimal relative
  pose problem.
\newblock In {\em CVPR}, 2018.

\bibitem{pro_cov_Brooks2001}
M.J. Brooks, W. Chojnacki, D. Gawley, and A. van~den Hengel.
\newblock What value covariance information in estimating vision parameters?
\newblock In {\em ICCV}, 2001.

\bibitem{cadena2016past}
Cesar Cadena, Luca Carlone, Henry Carrillo, Yasir Latif, Davide Scaramuzza,
  Jos{\'e} Neira, Ian Reid, and John~J Leonard.
\newblock Past, present, and future of simultaneous localization and mapping:
  Toward the robust-perception age.
\newblock {\em IEEE Transactions on robotics}, 32, 2016.

\bibitem{cayley1846algebraic}
Arthur Cayley.
\newblock About the algebraic structure of the orthogonal group and the other
  classical groups in a field of characteristic zero or a prime characteristic.
\newblock {\em Reine Angewandte Mathematik}, 32, 1846.

\bibitem{chandra1936generalised}
Mahalanobis~Prasanta Chandra et~al.
\newblock On the generalised distance in statistics.
\newblock In {\em Proceedings of the National Institute of Sciences of India},
  1936.

\bibitem{MRO_Chng2020}
Chee-Kheng Chng, {\'A}lvaro Parra, Tat-Jun Chin, and Yasir Latif.
\newblock Monocular rotational odometry with incremental rotation averaging and
  loop closure.
\newblock {\em Digital Image Computing: Techniques and Applications (DICTA)},
  2020.

\bibitem{choncha2021vo_init}
Alejo Concha, Michael Burri, Jesús Briales, Christian Forster, and Luc Oth.
\newblock Instant visual odometry initialization for mobile ar.
\newblock {\em IEEE Transactions on Visualization and Computer Graphics}, 27,
  2021.

\bibitem{cvivsic2018soft}
Igor Cvi{\v{s}}i{\'c}, Josip {\'C}esi{\'c}, Ivan Markovi{\'c}, and Ivan
  Petrovi{\'c}.
\newblock Soft-slam: Computationally efficient stereo visual simultaneous
  localization and mapping for autonomous unmanned aerial vehicles.
\newblock {\em Journal of field robotics}, 2018.

\bibitem{Cvisic2021}
Igor Cvišić, Ivan Marković, and Ivan Petrović.
\newblock {Recalibrating the KITTI Dataset Camera Setup for Improved Odometry
  Accuracy}.
\newblock In {\em European Conference on Mobile Robots (ECMR)}, 2021.

\bibitem{dorini2011unscented}
Leyza~Baldo Dorini and Siome~Klein Goldenstein.
\newblock Unscented feature tracking.
\newblock {\em Computer Vision and Image Understanding}, 115, 2011.

\bibitem{DSO_Engel2016}
J. Engel, V. Koltun, and D. Cremers.
\newblock Direct sparse odometry.
\newblock {\em IEEE Transactions on Pattern Analysis and Machine Intelligence},
  2018.

\bibitem{engel14eccv}
J. Engel, T. Schöps, and D. Cremers.
\newblock {LSD-SLAM}: Large-scale direct monocular {SLAM}.
\newblock In {\em ECCV}, 2014.

\bibitem{fathian2018quest}
Kaveh Fathian, J~Pablo Ramirez-Paredes, Emily~A Doucette, J~Willard Curtis, and
  Nicholas~R Gans.
\newblock Quest: A quaternion-based approach for camera motion estimation from
  minimal feature points.
\newblock {\em {{IEEE} Robotics and Automation Letters ({RAL})}}, 3, 2018.

\bibitem{Faugeras1989mul_solutions}
O.D. Faugeras and S. Maybank.
\newblock Motion from point matches: multiplicity of solutions.
\newblock In {\em Workshop on Visual Motion}, 1989.

\bibitem{fischler1981random}
Martin~A Fischler and Robert~C Bolles.
\newblock Random sample consensus: a paradigm for model fitting with
  applications to image analysis and automated cartography.
\newblock {\em Communications of the ACM}, 24, 1981.

\bibitem{forstner1987fast}
Wolfgang F{\"o}rstner and Eberhard G{\"u}lch.
\newblock A fast operator for detection and precise location of distinct
  points, corners and centres of circular features.
\newblock In {\em ISPRS intercommission conference on fast processing of
  photogrammetric data}, 1987.

\bibitem{Geiger2012CVPR}
Andreas Geiger, Philip Lenz, and Raquel Urtasun.
\newblock {Are we ready for Autonomous Driving? The KITTI Vision Benchmark
  Suite}.
\newblock In {\em CVPR}, 2012.

\bibitem{gonzalez2010measurement}
{\'A}lvaro Gonz{\'a}lez.
\newblock Measurement of areas on a sphere using fibonacci and
  latitude--longitude lattices.
\newblock {\em Mathematical Geosciences}, 42, 2010.

\bibitem{DBLP:conf/icra/GreeneR20}
W.~Nicholas Greene and Nicholas Roy.
\newblock Metrically-scaled monocular {SLAM} using learned scale factors.
\newblock In {\em {{IEEE} International Conference on Robotics and Automation
  ({ICRA})}}, 2020.

\bibitem{hartley2013rotation}
Richard Hartley, Jochen Trumpf, Yuchao Dai, and Hongdong Li.
\newblock Rotation averaging.
\newblock {\em International journal of computer vision}, 103(3):267--305,
  2013.

\bibitem{hartley1997defense}
Richard~I Hartley.
\newblock In defense of the eight-point algorithm.
\newblock {\em IEEE Transactions on pattern analysis and machine intelligence},
  19, 1997.

\bibitem{Hartley2004}
R.~I. Hartley and A. Zisserman.
\newblock {\em Multiple View Geometry in Computer Vision}.
\newblock Cambridge University Press, second edition, 2004.

\bibitem{hartree1928wave}
Douglas~R Hartree.
\newblock The wave mechanics of an atom with a non-coulomb central field. part
  i. theory and methods.
\newblock In {\em Mathematical Proceedings of the Cambridge Philosophical
  Society}, 1928.

\bibitem{DBLP:journals/inffus/HertzbergWFS13}
Christoph Hertzberg, Ren{\'{e}} Wagner, Udo Frese, and Lutz Schr{\"{o}}der.
\newblock Integrating generic sensor fusion algorithms with sound state
  representations through encapsulation of manifolds.
\newblock {\em Inf. Fusion}, 14, 2013.

\bibitem{kanatani2004geometric}
Kenichi Kanatani.
\newblock For geometric inference from images, what kind of statistical model
  is necessary?
\newblock {\em Systems and Computers in Japan}, 35, 2004.

\bibitem{kanatani2008statistical}
Kenichi Kanatani.
\newblock Statistical optimization for geometric fitting: Theoretical accuracy
  bound and high order error analysis.
\newblock {\em IJCV}, 80, 2008.

\bibitem{con_cov_Kanazawa2001}
Y. Kanazawa and K. Kanatani.
\newblock Do we really have to consider covariance matrices for image features?
\newblock In {\em ICCV}, 2001.

\bibitem{Klein2007PTAM}
Georg Klein and David Murray.
\newblock Parallel tracking and mapping for small ar workspaces.
\newblock In {\em IEEE and ACM International Symposium on Mixed and Augmented
  Reality}, 2007.

\bibitem{Kneip2014opengv}
Laurent Kneip and Paul Furgale.
\newblock Opengv: A unified and generalized approach to real-time calibrated
  geometric vision.
\newblock In {\em {{IEEE} International Conference on Robotics and Automation
  ({ICRA})}}, 2014.

\bibitem{EigenNEC_Kneip2013}
Laurent Kneip and Simon Lynen.
\newblock Direct optimization of frame-to-frame rotation.
\newblock In {\em ICCV}, 2013.

\bibitem{OrigNEC_Kneip2012}
Laurent Kneip, Roland Siegwart, and Marc Pollefeys.
\newblock Finding the exact rotation between two images independently of the
  translation.
\newblock In {\em ECCV}, 2012.

\bibitem{kruppa1913ermittlung}
Erwin Kruppa.
\newblock {\em Zur Ermittlung eines Objektes aus zwei Perspektiven mit innerer
  Orientierung}.
\newblock H{\"o}lder, 1913.

\bibitem{kukelova2008polynomial}
Zuzana Kukelova, Martin Bujnak, and Tomas Pajdla.
\newblock Polynomial eigenvalue solutions to the 5-pt and 6-pt relative pose
  problems.
\newblock In {\em BMVC}, 2008.

\bibitem{lawson1961contribution}
Charles~Lawrence Lawson.
\newblock {\em Contribution to the theory of linear least maximum
  approximation}.
\newblock PhD thesis, University of California, 1961.

\bibitem{lee2020geometric}
S.~H. Lee and Javier Civera.
\newblock Geometric interpretations of the normalized epipolar error.
\newblock {\em ArXiv}, abs/2008.01254, 2020.

\bibitem{ROBA_Lee2020}
Seong~Hun Lee and Javier Civera.
\newblock Rotation-only bundle adjustment.
\newblock In {\em CVPR}, 2021.

\bibitem{levenberg1944method}
Kenneth Levenberg.
\newblock A method for the solution of certain non-linear problems in least
  squares.
\newblock {\em Quarterly of applied mathematics}, 2, 1944.

\bibitem{li2006five}
Hongdong Li and Richard Hartley.
\newblock Five-point motion estimation made easy.
\newblock In {\em {IEEE} International Conference on Pattern Recognition
  ({ICPR})}, 2006.

\bibitem{lim2010estimating}
John Lim, Nick Barnes, and Hongdong Li.
\newblock Estimating relative camera motion from the antipodal-epipolar
  constraint.
\newblock {\em IEEE TPAMI}, 32, 2010.

\bibitem{longuet1987readings}
HC Longuet-Higgins.
\newblock Readings in computer vision: issues, problems, principles, and
  paradigms.
\newblock {\em A computer algorithm for reconstructing a scene from two
  projections}, 1987.

\bibitem{lowe2004distinctive}
David~G Lowe.
\newblock Distinctive image features from scale-invariant keypoints.
\newblock {\em IJCV}, 60, 2004.

\bibitem{KLT_Lukas1981}
Bruce~D. Lucas and Takeo Kanade.
\newblock An iterative image registration technique with an application to
  stereo vision.
\newblock In {\em International Joint Conference on Artificial Intelligence
  ({IJCAI})}, 1981.

\bibitem{marquardt1963algorithm}
Donald~W Marquardt.
\newblock An algorithm for least-squares estimation of nonlinear parameters.
\newblock {\em Journal of the society for Industrial and Applied Mathematics},
  11, 1963.

\bibitem{meidow2009reasoning}
Jochen Meidow, Christian Beder, and Wolfgang F{\"o}rstner.
\newblock Reasoning with uncertain points, straight lines, and straight line
  segments in 2d.
\newblock {\em ISPRS Journal of Photogrammetry and Remote Sensing}, 64, 2009.

\bibitem{moulon2013global}
Pierre Moulon, Pascal Monasse, and Renaud Marlet.
\newblock Global fusion of relative motions for robust, accurate and scalable
  structure from motion.
\newblock In {\em Proceedings of the IEEE International Conference on Computer
  Vision}, pages 3248--3255, 2013.

\bibitem{ORB_SLAM_Mur-Artal2015}
R. {Mur-Artal}, J.~M.~M. {Montiel}, and J.~D. {Tardós}.
\newblock Orb-slam: A versatile and accurate monocular slam system.
\newblock {\em IEEE Transactions on Robotics}, 31, 2015.

\bibitem{ORB_SLAM2_Mur-Artal2017}
R. {Mur-Artal} and J.~D. {Tardós}.
\newblock Orb-slam2: An open-source slam system for monocular, stereo, and
  rgb-d cameras.
\newblock {\em IEEE Transactions on Robotics}, 33, 2017.

\bibitem{EM_Nister2003}
D. {Nister}.
\newblock An efficient solution to the five-point relative pose problem.
\newblock In {\em CVPR}, 2003.

\bibitem{DBLP:books/sp/NocedalW99}
Jorge Nocedal and Stephen~J. Wright.
\newblock {\em Numerical Optimization}.
\newblock Springer, 1999.

\bibitem{Schindler2003mahala_plane}
K. Schindler and H. Bischof.
\newblock On robust regression in photogrammetric point clouds.
\newblock In {\em DAGM-Symposium}, 2003.

\bibitem{sheorey2014uncertainty}
Sameer Sheorey, Shalini Keshavamurthy, Huili Yu, Hieu Nguyen, and Clark~N
  Taylor.
\newblock Uncertainty estimation for klt tracking.
\newblock In {\em Asian Conference on Computer Vision}, 2014.

\bibitem{Steele2005Foerster}
R.M. Steele and C. Jaynes.
\newblock Feature uncertainty arising from covariant image noise.
\newblock In {\em CVPR}, 2005.

\bibitem{stewenius2006recent}
Henrik Stewenius, Christopher Engels, and David Nist{\'e}r.
\newblock Recent developments on direct relative orientation.
\newblock {\em ISPRS Journal of Photogrammetry and Remote Sensing}, 60, 2006.

\bibitem{stew5pt}
Henrik Stewénius, David Nistér, Magnus Oskarsson, and Kalle Åström.
\newblock Solutions to minimal generalized relative pose problems.
\newblock {\em Workshop on omnidirectional vision}, 2005.

\bibitem{szeliski2010computer}
Richard Szeliski.
\newblock {\em Computer vision: algorithms and applications}.
\newblock Springer Science \& Business Media, 2010.

\bibitem{KLT_Tomasi1991}
Carlo Tomasi and Takeo Kanade.
\newblock Detection and tracking of point features.
\newblock {\em IJCV}, 9, 1991.

\bibitem{triggs1999bundle}
Bill Triggs, Philip~F McLauchlan, Richard~I Hartley, and Andrew~W Fitzgibbon.
\newblock Bundle adjustment—a modern synthesis.
\newblock In {\em International workshop on vision algorithms}, 1999.

\bibitem{uhlmann1995dynamic}
Jeffrey~K Uhlmann.
\newblock {\em Dynamic map building and localization: New theoretical
  foundations}.
\newblock PhD thesis, University of Oxford Oxford, 1995.

\bibitem{Basalt_Usenko2020}
Vladyslav Usenko, Nikolaus Demmel, David Schubert, J{\"{o}}rg St{\"{u}}ckler,
  and Daniel Cremers.
\newblock Visual-inertial mapping with non-linear factor recovery.
\newblock {\em {{IEEE} Robotics and Automation Letters ({RAL})}}, 5, 2020.

\bibitem{DBLP:conf/cvpr/YangSWC20}
Nan Yang, Lukas von Stumberg, Rui Wang, and Daniel Cremers.
\newblock {D3VO:} deep depth, deep pose and deep uncertainty for monocular
  visual odometry.
\newblock In {\em CVPR}, 2020.

\bibitem{DBLP:conf/eccv/YangWSC18}
Nan Yang, Rui Wang, J{\"{o}}rg St{\"{u}}ckler, and Daniel Cremers.
\newblock Deep virtual stereo odometry: Leveraging deep depth prediction for
  monocular direct sparse odometry.
\newblock In {\em ECCV}, 2018.

\bibitem{SIFTCOV_Zeisl2009}
Bernhard Zeisl, Pierre Georgel, Florian Schweiger, Eckehard Steinbach, and
  Nassir Navab.
\newblock Estimation of location uncertainty for scale invariant feature
  points.
\newblock In {\em BMVC}, 2009.

\bibitem{zhang2017uncertainty}
Hongmou Zhang, Denis Grie{\ss}bach, J{\"u}rgen Wohlfeil, and Anko B{\"o}rner.
\newblock Uncertainty model for template feature matching.
\newblock In {\em Pacific-Rim Symposium on Image and Video Technology}, pages
  406--420. Springer, 2017.

\bibitem{GRQZhang1}
Lei-Hong Zhang.
\newblock On optimizing the sum of the rayleigh quotient and the generalized
  rayleigh quotient on the unit sphere.
\newblock {\em Computational Optimization and Applications}, 54, 2013.

\bibitem{GRQZhang2}
Lei-Hong Zhang and Rui Chang.
\newblock A nonlinear eigenvalue problem associated with the
  sum-of-rayleigh-quotients maximization.
\newblock {\em Transactions on Applied Mathematics}, 2, 2021.

\end{thebibliography}
}
\newpage
\appendix
\twocolumn[\centering \section*{The Probabilistic Normal Epipolar Constraint for Frame-To-Frame Rotation Optimization under Uncertain Feature Positions\\[1.3ex]Supplementary Material}]

\section{Overview}
In this supplementary material, we present additional insight into the \ac{pnec} energy function, give more details about the practical implementations of the \ac{pnec} rotation estimation, and show further experimental results. We address limitations of the \ac{pnec} in \autoref{sec:limitations}. \autoref{sec:klt} shows how we use the KLT tracking to extract position covariance matrices. The unscented transform and its use for uncertainty propagation are presented in \autoref{sec:unsc_transform}. \autoref{sec:geometry} gives geometric insights into the \ac{pnec}. We derive the directional limit for the singularities in \autoref{sec:limit}. The \ac{scf} optimization is explained in \autoref{sec:scf}. \autoref{sec:Hyperparameters} gives an overview of the hyperparameters we use in our experiments. The simulated experiments for additional noise types not presented in the main paper are shown in \autoref{sec:simulated}. \autoref{sec:energy} presents further results of the simulated experiments with regard to the energy function. Experiments on the influence of anisotropy and wrongly estimated covariance matrices are given in \autoref{sec:anisotropy} and \autoref{sec:offset}. We show further analysis of the results on the KITTI dataset together with an ablation study in \autoref{sec:kitti}.

\section{Limitations} \label{sec:limitations}
This section will give further details and explanations, which were not mentioned in the main paper due to the constrained space, regarding the limitations of the proposed method.

As written in the main paper, the \ac{pnec} optimization scheme consists of two stages that result in a more involved optimization than for the original \ac{nec}. Due to the iterative optimization and the joint refinement, the \ac{pnec} will not be as fast as the \ac{nec}, which is shown by the runtime experiment in the main paper. However, the integration of both methods into a RANSAC scheme mitigates the difference between them. Overall, our \ac{pnec} and the \ac{nec} both run in real-time on the KITTI dataset.

A further limitation of the \ac{pnec} is its dependence on additional positional uncertainty. In contrast, the \ac{nec} only requires feature positions. While the positional uncertainty allows our \ac{pnec} to achieve more accurate rotation estimation, the information needs to be sufficiently accurate to provide a benefit. The influence of insufficient information can be seen in the results on seq.~01 of the KITTI dataset, where the KLT tracker produces wrong feature positions and uncertainty. The the poor performance of the \ac{nec} cannot be overcome by the \ac{pnec}. The \ac{pnec} puts emphasis on accurately tracked feature correspondences during its optimization. However, given poor feature correspondences it can emphasize wrong features. Therefore, outlier removal is a necessary step for the \ac{pnec}.

In our formulation of the \ac{pnec} we assumed a Gaussian error model for the feature position. This assumption is not accurate for most feature position errors. A more sophisticated error model could lead to even more accurate rotation estimates. However, it would also lead to a more intricate energy function. We consider this an interesting direction for further investigations. The performance of the \ac{pnec} on the KITTI dataset shows that the Gaussian assumption can deliver good performance for real-world data.

\section{Extracting Feature Position Uncertainties from KLT Tracks}\label{sec:klt}
\begin{figure}[t]
    \centering
    \includegraphics[width=0.49\textwidth]{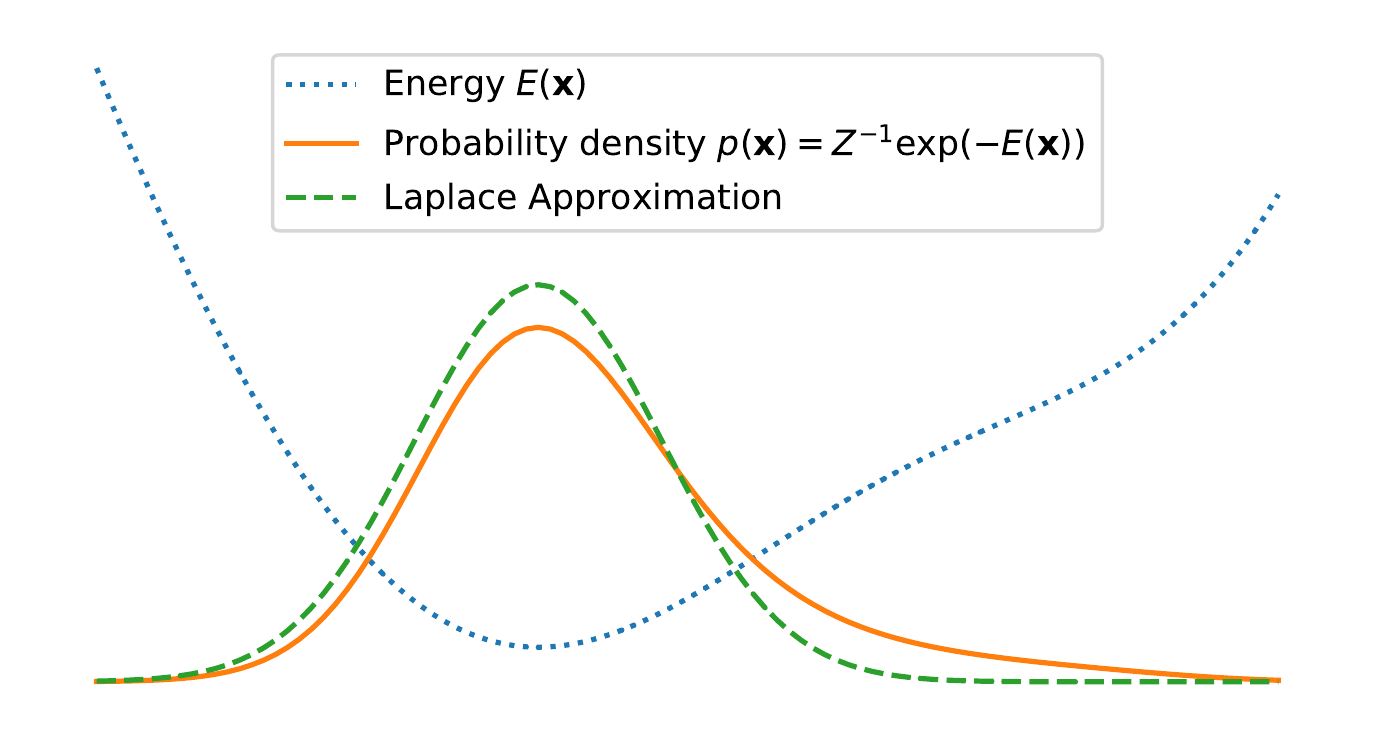}
    \caption{The plot shows the energy function $E(\fx)$ (\textit{dotted blue}), the normalized distribution $p(\fx)$ (\textit{orange}), together with the Laplace approximation centered on the mode $\fx^\ast$ of $p(\fx)$ (\textit{dashed green}).}
    \label{fig:laplace}
\end{figure}
In the following, we explain how to obtain position uncertainties from the energy function. To this end, we first introduce the relationship between an energy function and the Boltzmann distribution, then the Laplace approximation, and finally apply both to \ac{klt} tracks.

Given an energy function ${E(\fx): \R^d \to \R}$ we can derive an associated probability distribution
\begin{equation}
    p(\fx) = Z^{-1} \exp(-E(\fx))
\end{equation}
that is often referred to as the \emph{Boltzmann distribution}~\cite[Eqn. (8.41)]{DBLP:journals/jei/BishopN07}. The constant $Z > 0$ ensures that the distribution is normalized. If $\fx^\ast$ is a local minimizer of $E(\fx)$ and thus at a local maximum of $p(\fx)$ we can use the \emph{Laplace approximation} to derive a Gaussian distribution that locally approximates $p(\fx)$ around $\fx^\ast$ \cite[Sec. 4.4]{DBLP:journals/jei/BishopN07}. For an illustration please see \autoref{fig:laplace}. Specifically, the Gaussian distribution has mean $\f{\mu} = \fx^\ast$ and inverse covariance
\begin{equation}
    \f{\Sigma}^{-1} = \left. \frac{d^2}{d\fx^2} E(\fx) \right\vert_{\fx = \fx^\ast} \,,
\end{equation}
where $\frac{d^2}{d\fx^2}$ denotes the Hessian matrix.

For \ac{klt} tracking we use the sum of squared normalized differences as the energy function
\begin{equation} \label{eqn:klt-energy}
    E_{KLT}(\fT) = \sum_{\fp \in P} \left( \frac{I_h(\fp)}{\bar{I}_h} - \frac{I_t(\fT(\fp))}{\bar{I}_t} \right)^2 \,,
\end{equation}
where
\begin{equation*}
    \bar{I}_h = \frac{1}{|P|} \sum_{\fp_i \in P} I_h(\fp_i), \quad
    \bar{I}_t = \frac{1}{|P|} \sum_{\fp_i \in P} I_t(\fT(\fp_i))
\end{equation*}
are the mean intensity in the host frame $I_h$ and target frame $I_t$, respectively. Furthermore, $\fT$ is the transformation between host and target frame that is being optimized, $P$ is the pattern of pixels over which the sum is computed, and $|P|$ denotes the number of pixels in $P$.

The \ac{klt} tracking implementation from \cite{Basalt_Usenko2020} that we use for our experiments on KITTI uses the \emph{inverse compositional formulation}~\cite{DBLP:journals/ijcv/BakerM04} that allows for more efficient tracking. Due to this formulation, the tracking optimizes a proxy for \autoref{eqn:klt-energy}
and thus the approximation of the Hessian of the energy function is computed once in the host frame. The full explanation of the inverse compositional formulation is outside the scope of this paper and for further details we kindly refer the reader to the excellent paper by Baker and Matthews~\cite{DBLP:journals/ijcv/BakerM04}, which is the first in a multi-paper series.

Because the approximate Hessian is computed in the host frame only we omit the subscript and denote the host frame by $I$ in the following. We first compute the Jacobian for each pixel $\fp_i$ in the pattern $P$ and then accumulate all Jacobians into the Gauss-Newton approximation of the Hessian. We denote the image gradient by $\nabla I(\boldsymbol{p})$ and the Jacobian w.r.t. the pixel position by $\f{J}_{\fp_i}$, which gives
\begin{equation}
    \boldsymbol{J}_{\xi, i} = \begin{pmatrix}
        1 & 0 & -p_{i,y} \\
        0 & 1 & p_{i,x}
    \end{pmatrix}
\end{equation}
\begin{equation}
    \f{J}_{\fp_i} = |P| \frac{\nabla I(\boldsymbol{p}_i)^\top \boldsymbol{J}_{\xi, i} \sum\limits_{\boldsymbol{p}_j \in P} I(\boldsymbol{p}_j) - I(\boldsymbol{p}_i)  \sum\limits_{\boldsymbol{p}_j \in P} \nabla I(\boldsymbol{p}_j)^\top \boldsymbol{J}_{\xi, j}  }{\left(\sum\limits_{\boldsymbol{p}_j \in P}I(\boldsymbol{p}_j)\right)^2}
\end{equation}
\begin{equation}
    \boldsymbol{J}_{SE(2)} = \begin{pmatrix}
        \boldsymbol{J}_1 \\
        \boldsymbol{J}_2 \\
        \vdots           \\
        \boldsymbol{J}_n
    \end{pmatrix}
\end{equation}
\begin{equation}
    \boldsymbol{H}_{\text{SE}(2)} = \boldsymbol{J}_{SE(2)}^\top  \boldsymbol{J}_{SE(2)}
\end{equation}
\begin{equation}
    \f{\Sigma}_{\text{SE}(2)} = \boldsymbol{H}_{\text{SE}(2)}^{-1}
\end{equation}

The patches are tracked w.r.t.\ an $\text{SE}(2)$ transforms and thus the covariance $\f{\Sigma}_{\text{SE}(2)}$ is for the full $\text{SE}(2)$ transform, i.e. the translation $u, v$ in pixel coordinates as well as the 2D rotation by an angle $\theta$. As we only consider the position, we take the marginal over the first two coordinates by selecting the upper left $2 \times 2$ sub-matrix $\f{\Sigma}_{\text{2D},h}$ in the host frame. We then transform this matrix to the target frame using the estimated rotation
\begin{equation}
    \f{\Sigma}_{\text{2D},t} = \fR_\theta \f{\Sigma}_{\text{2D},h} \fR_\theta^\top \,,
\end{equation}
where $\fR_\theta$ is the 2D rotation matrix corresponding to a rotation by an angle $\theta$. Finally, the matrix $\f{\Sigma}_{\text{2D},t}$ is the matrix we use for the \ac{pnec} and denote $\f{\Sigma}_{\text{2D}}$ in the main paper.

\section{Unscented Transform}\label{sec:unsc_transform}
\begin{figure}[t]
    \centering
    \begin{subfigure}[b]{0.22\textwidth}
        \centering
        {\includegraphics[trim={0.5cm 1.2cm 5.7cm 2.5cm},clip,width=\textwidth]{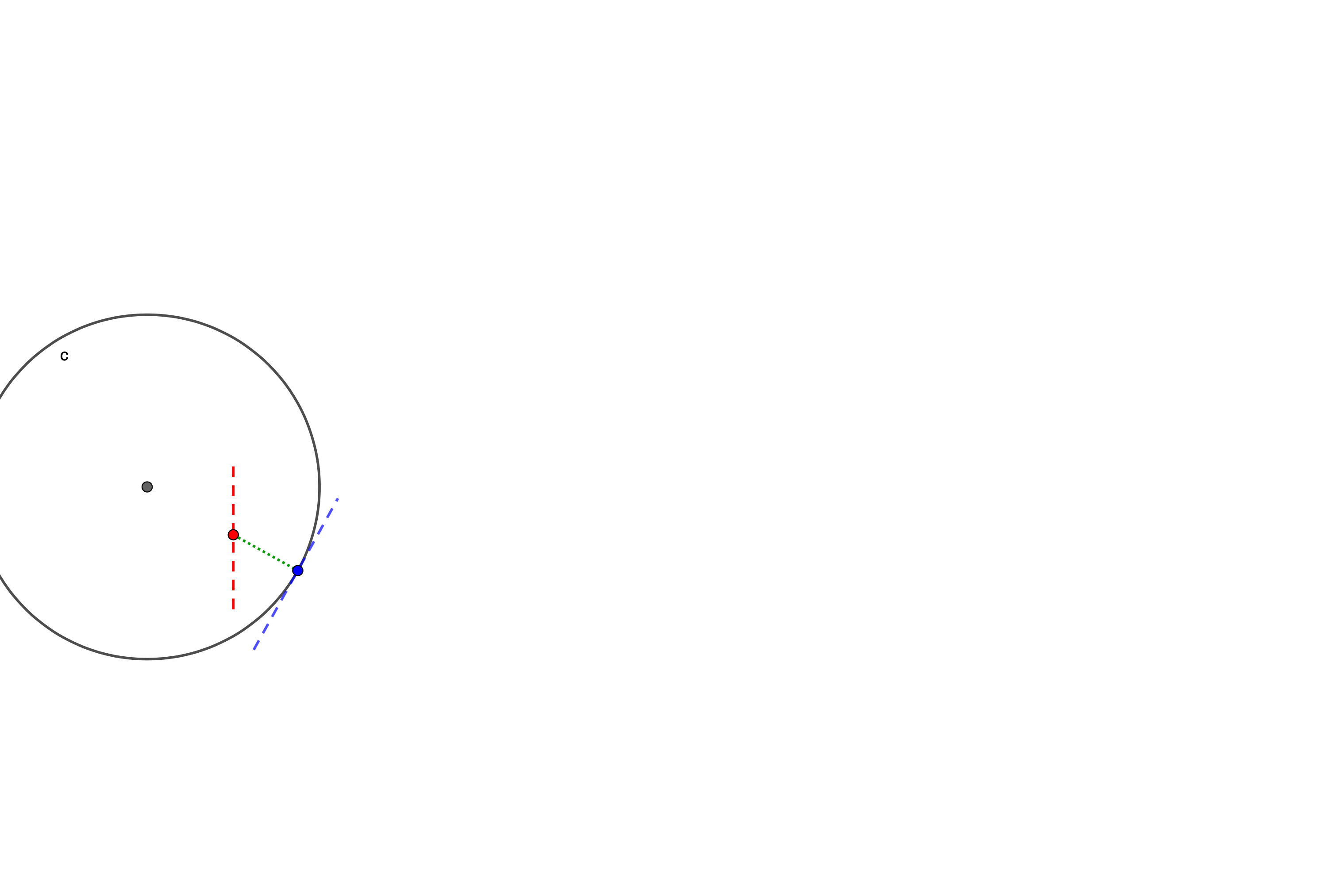}}
        \caption{Linear approximation}
        \label{fig:linear}
    \end{subfigure}
    \hfill
    \begin{subfigure}[b]{0.22\textwidth}
        \centering
        {\includegraphics[trim={0.5cm 1.2cm 5.7cm 2.5cm},clip,width=\textwidth]{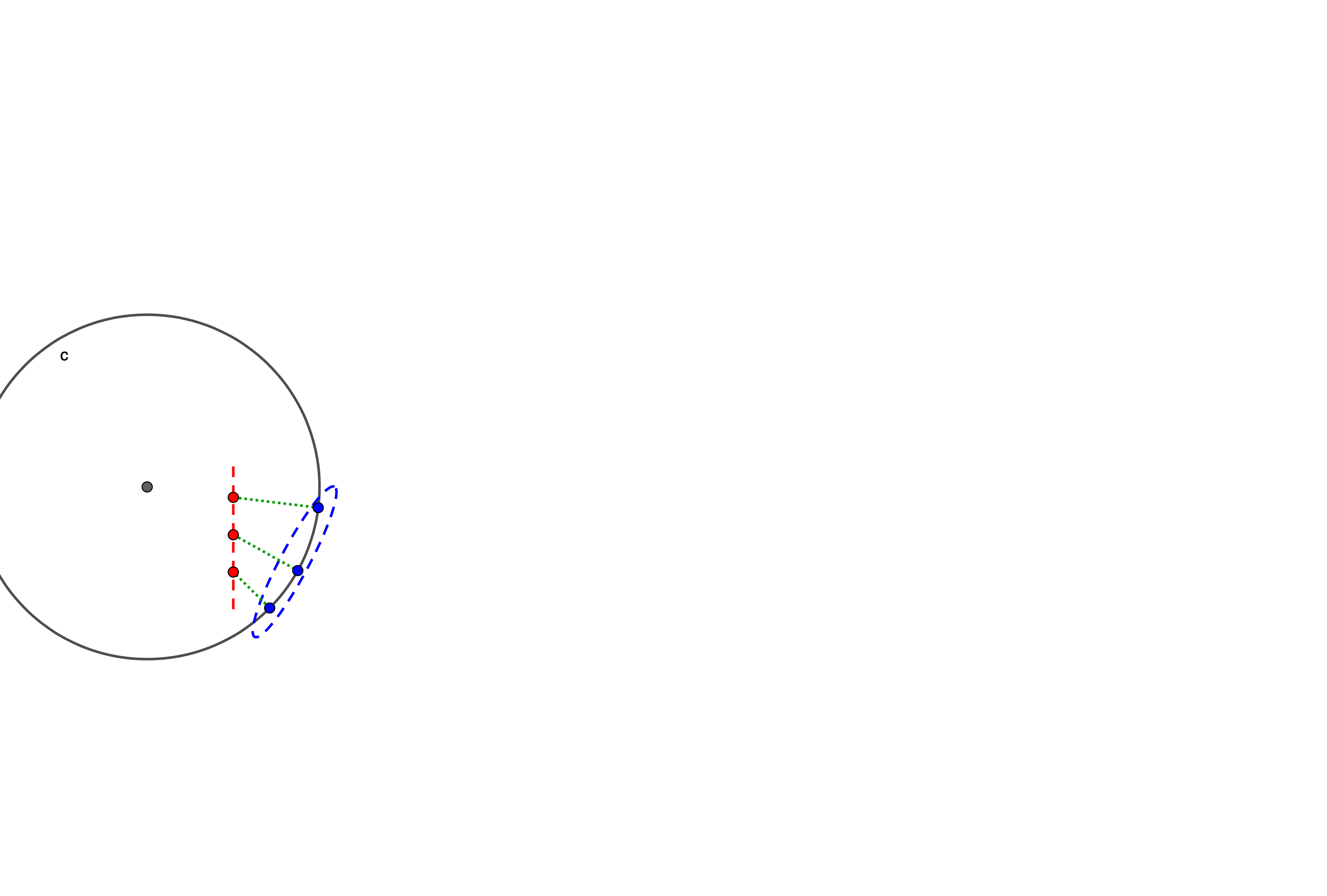}}
        \caption{Unscented transform}
        \label{fig:unsc}
    \end{subfigure}
    \caption{Illustration of the difference between linear approximation (a) and unscented transform (b) of the projection of the covariance onto the unit-sphere in 2D. The linear approximation of the projection results in a covariance tangential to the unit-sphere---the covariance matrix does not have full rank. The unscented transform captures the non-linearity of the projection---the covariance matrix has full rank.}
    \label{fig:transform}
\end{figure}
In the following, we give an overview over the unscented transform \cite{uhlmann1995dynamic} and present how it it used in the \ac{pnec}. For an illustration of the unscented transform see \autoref{fig:transform}. The unscented transform gives an approximation for the mean and covariance if a non-linear transformation is applied to a Gaussian distribution. The unscented transform computes the mean and covariance from selected points to which the non-linear transformation is applied.
Given a mean $\boldsymbol{\mu} \in \mathbb{R}^n$ and covariance $\boldsymbol{\Sigma} \in \mathbb{R}^{n \times n}$ the unscented transform selects $2n + 1$ points
\begin{equation}
    \begin{aligned}
        \boldsymbol{\xi}_0        & = \boldsymbol{\mu} \,,                                    &                   \\
        w_0                       & = \frac{\kappa}{n + \kappa} \,,                           &                   \\
        \boldsymbol{\xi}_{i, i+n} & = \boldsymbol{\mu} \pm \sqrt{n + \kappa} \boldsymbol{C}_i & i = 1 \dots n \,, \\
        w_{i, i+n}                & = \frac{1}{2 (n + \kappa)}                                & i = 1 \dots n \,,
    \end{aligned}
\end{equation}
with corresponding weights around the mean, where $\boldsymbol{C}_i$ is the i-th column of the matrix $\boldsymbol{C}$ such that $\boldsymbol{\Sigma} = \boldsymbol{C} \boldsymbol{C}^\top$, and $\kappa$ controls the spread of the points, which we set to the default value $\kappa=1$ . A popular way to compute $\boldsymbol{C}$ is using the Cholesky-decomposition of $\boldsymbol{\Sigma}$. The non-linear function $f: \mathbb{R}^n \mapsto \mathbb{R}^m$ is applied to the points
\begin{equation}
    \boldsymbol{\zeta}_i = f(\boldsymbol{\xi}_i)
\end{equation}
giving $2n+1$ points from which the new mean and covariance are computed
\begin{equation}
    \begin{aligned}
        \boldsymbol{\mu}_f    & = \sum_{i=0}^{2n} w_i \boldsymbol{\zeta}_i \,,                                                                         \\
        \boldsymbol{\Sigma}_f & = \sum_{i=0}^{2n} w_i (\boldsymbol{\zeta}_i - \boldsymbol{\mu}_f) (\boldsymbol{\zeta}_i - \boldsymbol{\mu}_f)^\top \,.
    \end{aligned}
\end{equation}
For the \ac{pnec} we project the 2D covariance $\boldsymbol{\Sigma}_{2\text{D}}$ feature in the image plane onto the unit-sphere in 3D.

\textbf{Pinhole Cameras}.
The unscented transform for pinhole cameras is straight forward. We have the feature uncertainty matrix in the image plane $\boldsymbol{\Sigma}_{2\text{D}}$ and can easily compute its Cholesky-decomposition. We sample 5 points around the feature position in the image and project them through the non-linear function $f(\boldsymbol{\xi})$. The function $f(\boldsymbol{\xi})=h(g(\boldsymbol{\xi}))$ consist of the unprojection function
\begin{equation}
    g(\boldsymbol{\xi}) = \boldsymbol{K}_{inv} \begin{pmatrix} \xi_1 \\ \xi_2 \\ 1 \end{pmatrix}
\end{equation}
of the pinhole camera with the inverse camera matrix $\boldsymbol{K}_{inv}$ and the projection onto the unit sphere
\begin{equation}
    h(\boldsymbol{x}) = \frac{\boldsymbol{x}}{\| \boldsymbol{x} \|} \,.
\end{equation}

\textbf{Omnidirectional Cameras}.
The unscented transform for omnidirectional cameras is a bit more involved. For omnidirectional cameras image points are not on a plane but a sphere. The covariance matrix $\boldsymbol{\Sigma}_{spherical}$ corresponding to a point are therefore tangential to this sphere. They are living in a 2D subspace in a 3D space, the $3 \times 3$ covariance matrices are not positive definite and therefore the Cholesky-decomposition is not defined for them. Instead, we use the Cholesky-decomposition of a $2 \times 2$ sub-matrix $ \boldsymbol{C} \boldsymbol{C}^\top = \boldsymbol{\Sigma}_{2\text{D}}$ of the form
\begin{equation}
    \boldsymbol{R}^\top \begin{pmatrix}
        \boldsymbol{\Sigma}_{2\text{D}} & \boldsymbol{0} \\
        \boldsymbol{0}^\top             & 0
    \end{pmatrix}\boldsymbol{R} =  \boldsymbol{\Sigma}_{spherical}\,.
\end{equation}
A matrix $\boldsymbol{R}$ that gives us such a form is the rotation matrix
\begin{equation}
    \boldsymbol{R} = \frac{1}{\| \boldsymbol{\mu} \| }\begin{pmatrix} \| \boldsymbol{\mu} \|  - \frac{\mu_1^2}{\| \boldsymbol{\mu} \|  +\mu_3} & - \frac{\mu_1 \mu_2}{1+\mu_3}                                           & - \mu_1 \\
        - \frac{\mu_1 \mu_2}{1+\mu_3}                                            & \| \boldsymbol{\mu} \|  - \frac{\mu_2^2}{\| \boldsymbol{\mu} \| +\mu_3} & - \mu_2 \\
        \mu_1                                                                    & \mu_2                                                                   & \mu_3\end{pmatrix}\,,
\end{equation}
that aligns the feature point with the z-axis and the covariance matrix with the xy-plane, where $\mu_i$ denotes the ith element of $\boldsymbol{\mu}$.

The 5 points for the unscented transform are selected using
\begin{equation}
    \begin{aligned}
        \boldsymbol{\xi}_0        & = \boldsymbol{\mu} \,,                                                                  &                   \\
        \boldsymbol{\xi}_{i, i+n} & = \boldsymbol{\mu} \pm \sqrt{n + \kappa} \boldsymbol{R}^\top \begin{pmatrix} \boldsymbol{C}_i \\ 0 \end{pmatrix} & i = 1 \dots n \,.
    \end{aligned}
\end{equation}
Unlike the pinhole camera we do not need a unprojection function and the non-linear function is given by
\begin{equation}
    f(\boldsymbol{\xi}) = \frac{\boldsymbol{\xi}}{\| \boldsymbol{\xi} \|}\,.
\end{equation}

\section{Geometric Interpretations of the PNEC}\label{sec:geometry}
As we explain in the main paper, the \ac{pnec} incoorperates uncertainty into the \ac{nec} and leads to a Mahalanobis-distance-based energy function. In this section, we take an in-depth look into the derivation of the energy function as a Mahalanobis distance and give a geometric reasoning why the regularization removes the singularity of the \ac{pnec}.

In the main paper, we look at the univariate distribution of an error residual with its variance $\sigma_i^2$ that leads to the \ac{pnec} energy function. Another way to derive the same energy function is by looking at the distribution of an {\em epipolar plane normal vector} $\boldsymbol{n}_i$. It has a Gaussian distribution with the covariance $\boldsymbol{\Sigma}_{n, i}$. Using the formulation of the Mahalanobis distance to a plane by Schindler and Bischof \cite{Schindler2003mahala_plane} we can derive the distance of this normal vector to the {\em epipolar normal plane}. The idea of Schindler and Bischof is to apply a whitening transform to the error distribution and the plane and compute the Euclidean distance of the new center to the transformed plane.

For the \ac{pnec} we have the {\em epipolar normal plane} described in homogeneous coordinates by
\begin{equation}
    \boldsymbol{p} = \begin{pmatrix}
        \boldsymbol{t} \\ 0
    \end{pmatrix}
\end{equation}
and the covariance with its singular value decomposition
\begin{equation}
    \boldsymbol{\Sigma}_{n, i} = \boldsymbol{R}_i^\top \boldsymbol{V}_i \boldsymbol{R}_i, \quad \boldsymbol{V}_i = diag(a^2, b^2, c^2)
\end{equation}
from which we can derive the whitening transform. For the plane it is given by
\begin{equation}
    \boldsymbol{q} = \begin{pmatrix}
        \boldsymbol{V}_i^{1/2}\boldsymbol{R}_i & \boldsymbol{0} \\
        -\boldsymbol{n}_i^\top                 & 1
    \end{pmatrix} \boldsymbol{p} = \begin{pmatrix}
        \boldsymbol{V}_i^{1/2}\boldsymbol{R}_i \boldsymbol{t} \\ -\boldsymbol{n}_i^\top \boldsymbol{t}
    \end{pmatrix}.
\end{equation}
The original Mahalanobis distance is now given by the distance of the origin to the transformed plane
\begin{equation}
    d_M = \frac{q_4}{\sqrt{q_1^2 + q_2^2 + q_3^2}}
\end{equation}
giving us
\begin{equation}
    d_M^2 = \frac{|\boldsymbol{n}_i^\top \boldsymbol{t} |^2}{\boldsymbol{t}^\top  \boldsymbol{R}_i^\top \boldsymbol{V}_i^{1/2} \boldsymbol{V}_i^{1/2} \boldsymbol{R}_i \boldsymbol{t}} = \frac{|\boldsymbol{n}_i^\top \boldsymbol{t} |^2}{\boldsymbol{t}^\top   \boldsymbol{\Sigma}_{n, i} \boldsymbol{t}}
\end{equation}
for the \ac{pnec}, which is the residual of the energy function.

Using the geometric interpretation of the \ac{pnec} as a Mahalanobis distance of the normal vector to the {\em epipolar normal plane} we can now explain the singularity of the \ac{pnec} geometrically. Since $\boldsymbol{n}_i = \boldsymbol{f}_i \times \boldsymbol{R} \boldsymbol{f}_i^\prime$ is derived as a cross product its distribution is only a 2D plane embedded in 3D space. The 2D plane is described by its normal vector $\boldsymbol{f}_i$ (note that $\boldsymbol{f}_i$ is not random while $\boldsymbol{f}_i^\prime$ is random). Therefore, the Mahalanobis distance is only defined for points in this 2D subspace. In all configuration except for $\boldsymbol{f}_i = \boldsymbol{t}$ the 2D plane and the {\em epipolar normal plane} intersect in a single line giving a meaningful Mahalanobis distance. The proposed regularization is equivalent to removing the 2D subspace constraint on the Mahalanobis distance by giving the covariance matrix full rank due to
\begin{equation}
    \begin{aligned}
        \boldsymbol{t}^\top \left( \boldsymbol{\Sigma}_{n, i} + c \boldsymbol{I}_3 \right)  \boldsymbol{t} & =   \boldsymbol{t}^\top \boldsymbol{\Sigma}_{n, i}  \boldsymbol{t} + c \boldsymbol{t}^\top \boldsymbol{I}_3 \boldsymbol{t} \\
                                                                                                           & = \boldsymbol{t}^\top \boldsymbol{\Sigma}_{n, i}  \boldsymbol{t} + c \,.
    \end{aligned}
\end{equation}

\section{Directional Limit at the Singularity}\label{sec:limit}
In this section, we further investigate the singularity of the \ac{pnec}. Since no limit for the singularity exists, we present its directional limit. To this end, we use spherical coordinates to approach the singularity on the unit-sphere. For convenience we restate the \ac{pnec} weighted residual
\begin{equation}
    e_{P, i}^2(\boldsymbol{R}, \boldsymbol{t}) = \frac{e_i^2}{\sigma_i^2} = \frac{| \boldsymbol{t}^\top (\boldsymbol{f}_i \times \boldsymbol{R} \boldsymbol{f}^\prime_i) |^2}{\boldsymbol{t}^\top \hat{\boldsymbol{f}_i} \boldsymbol{R} \boldsymbol{\Sigma}_i \boldsymbol{R}^\top \hat{\boldsymbol{f}_i}{}^\top \boldsymbol{t}} \,.
\end{equation}
Since the limit
\begin{equation}\label{eqn:limit-full}
    \lim_{\boldsymbol{f}_i \to \boldsymbol{t}}  e_{P, i}^2(\boldsymbol{R}, \boldsymbol{t})
\end{equation}
does not exist, we look at the directional limit of the singularity. Without loss of generality we choose
\begin{equation}
    \boldsymbol{t} = \begin{pmatrix} 0 \\ 0 \\ 1 \end{pmatrix}.
\end{equation}
We can now approach the translation on the unit sphere by choosing an arbitrary vector
\begin{equation}
    \boldsymbol{f}_i(\theta, \phi) = \begin{pmatrix} \sin{\theta}\sin{\phi} \\  - \sin{\theta}\cos{\phi} \\ \cos{\theta} \end{pmatrix}
\end{equation}
on the unit sphere in spherical coordinates with radius $1$. Letting $\theta \to 0$ implies $\boldsymbol{f}_i \to \boldsymbol{t}$. We can rewrite the residual as
\begin{equation}
    e_{P, i}^2(\boldsymbol{R}, \boldsymbol{t}) = \frac{|(\boldsymbol{t} \times \boldsymbol{f}_i)^\top \boldsymbol{R} \boldsymbol{f}^\prime_i|^2}{(\boldsymbol{t} \times \boldsymbol{f}_i)^\top \boldsymbol{R} \boldsymbol{\Sigma}_i \boldsymbol{R}{}^\top (\boldsymbol{t} \times \boldsymbol{f}_i)}
\end{equation}
and the directional limit is given by
\begin{equation}
    \lim_{\theta \to 0} \frac{|(\boldsymbol{t} \times \boldsymbol{f}_i(\theta))^\top \boldsymbol{R} \boldsymbol{f}^\prime_i|^2}{(\boldsymbol{t} \times \boldsymbol{f}_i(\theta))^\top \boldsymbol{R} \boldsymbol{\Sigma}_i \boldsymbol{R}{}^\top (\boldsymbol{t} \times \boldsymbol{f}_i(\theta))} \,.
\end{equation}
The cross product is given by
\begin{equation}
    \boldsymbol{t} \times \boldsymbol{f}_i(\theta) = - \sin{\theta} \begin{pmatrix} \cos{\phi} \\ \sin{\phi} \\ 0 \end{pmatrix} = - \sin{\theta} \boldsymbol{k}
\end{equation}
with $\boldsymbol{k}$ being the unit length vector orthogonal to $\boldsymbol{f}_i \text{ and } \boldsymbol{t}$.
\begin{equation}
    \begin{aligned}
         & \lim_{\theta \to 0} \frac{|(\boldsymbol{t} \times \boldsymbol{f}_i(\theta))^\top \boldsymbol{R} \boldsymbol{f}^\prime_i|^2}{(\boldsymbol{t} \times \boldsymbol{f}_i(\theta))^\top \boldsymbol{R} \boldsymbol{\Sigma}_i \boldsymbol{R}{}^\top (\boldsymbol{t} \times \boldsymbol{f}_i(\theta))} \\
         & = \lim_{\theta \to 0} \frac{\sin^2\theta}{\sin^2\theta} \frac{|\boldsymbol{k}^\top \boldsymbol{R} \boldsymbol{f}^\prime_i|^2}{\boldsymbol{k}^\top \boldsymbol{R} \boldsymbol{\Sigma}_i \boldsymbol{R}{}^\top \boldsymbol{k}}                                                                   \\
         & = \frac{|\boldsymbol{k}^\top \boldsymbol{R} \boldsymbol{f}^\prime_i|^2}{\boldsymbol{k}^\top \boldsymbol{R} \boldsymbol{\Sigma}_i \boldsymbol{R}{}^\top \boldsymbol{k}}
    \end{aligned}
\end{equation}
From the above equation we can clearly see that the directional limit for $\theta \to 0$ exists and depends on the direction $\f{k}$. Consequentially, the limit in \autoref{eqn:limit-full} does not exist.

\section{SCF}\label{sec:scf}
\begin{figure}[t]
    \centering
    \begin{subfigure}[b]{0.23\textwidth}
        \centering
        {\includegraphics[trim={1cm 3cm 1cm 2cm},clip,width=\textwidth]{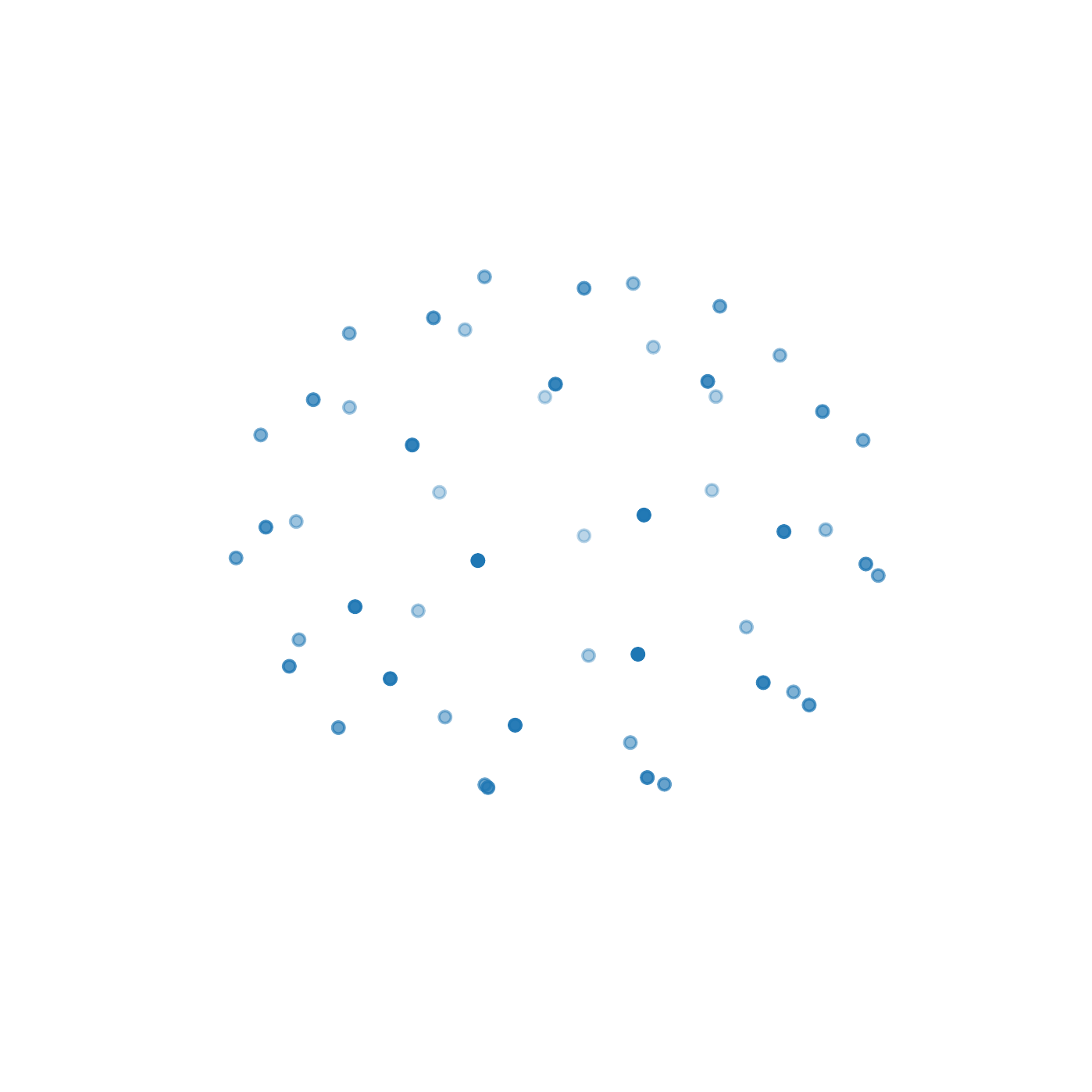}}
        \caption{50 points}
        \label{fig:fib_50}
    \end{subfigure}
    \hfill
    \begin{subfigure}[b]{0.23\textwidth}
        \centering
        {\includegraphics[trim={1cm 3cm 1cm 2cm},clip,width=\textwidth]{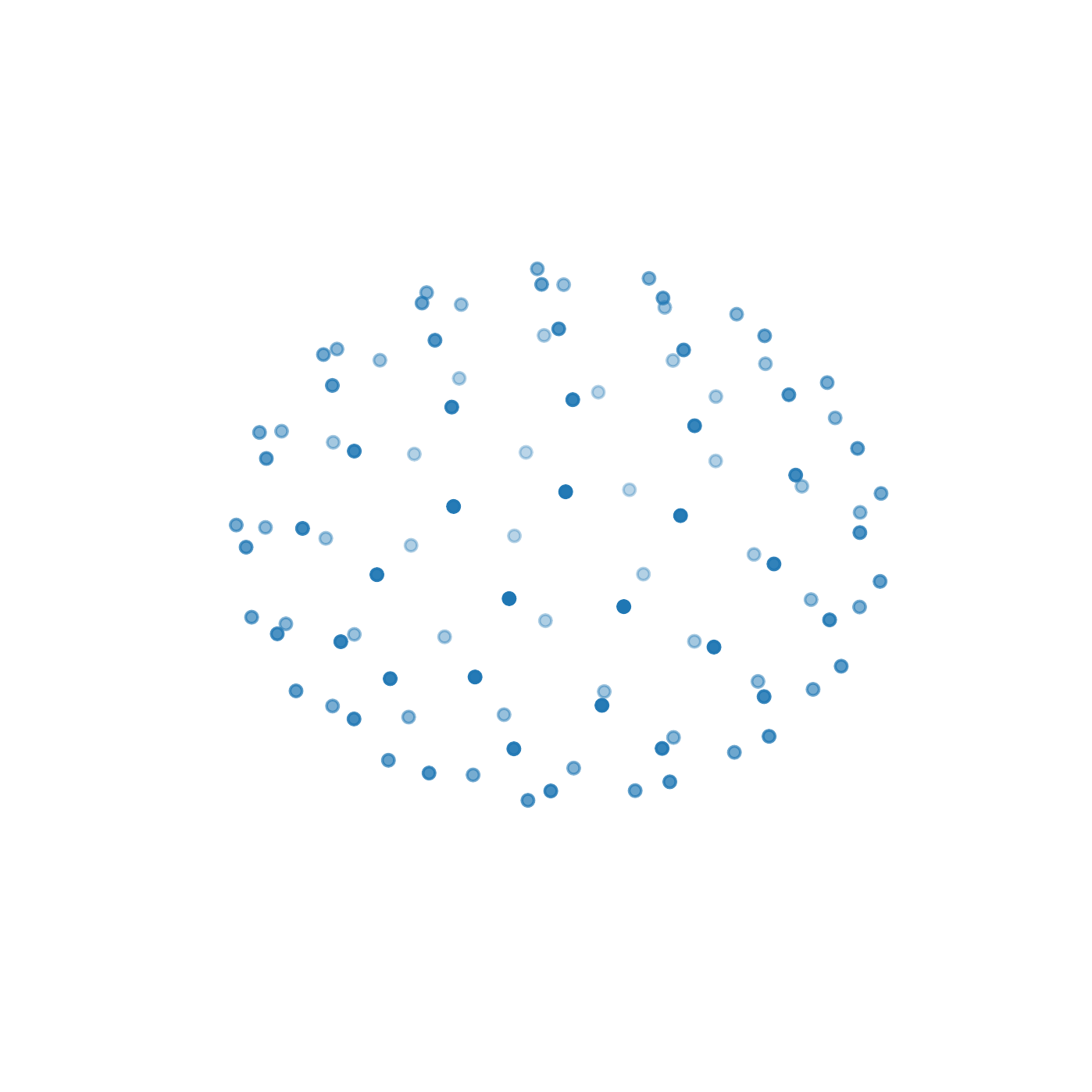}}
        \caption{100 points}
        \label{fig:fib_100}
    \end{subfigure}
    \hfill
    \begin{subfigure}[b]{0.23\textwidth}
        \centering
        {\includegraphics[trim={1cm 3cm 1cm 2cm},clip,width=\textwidth]{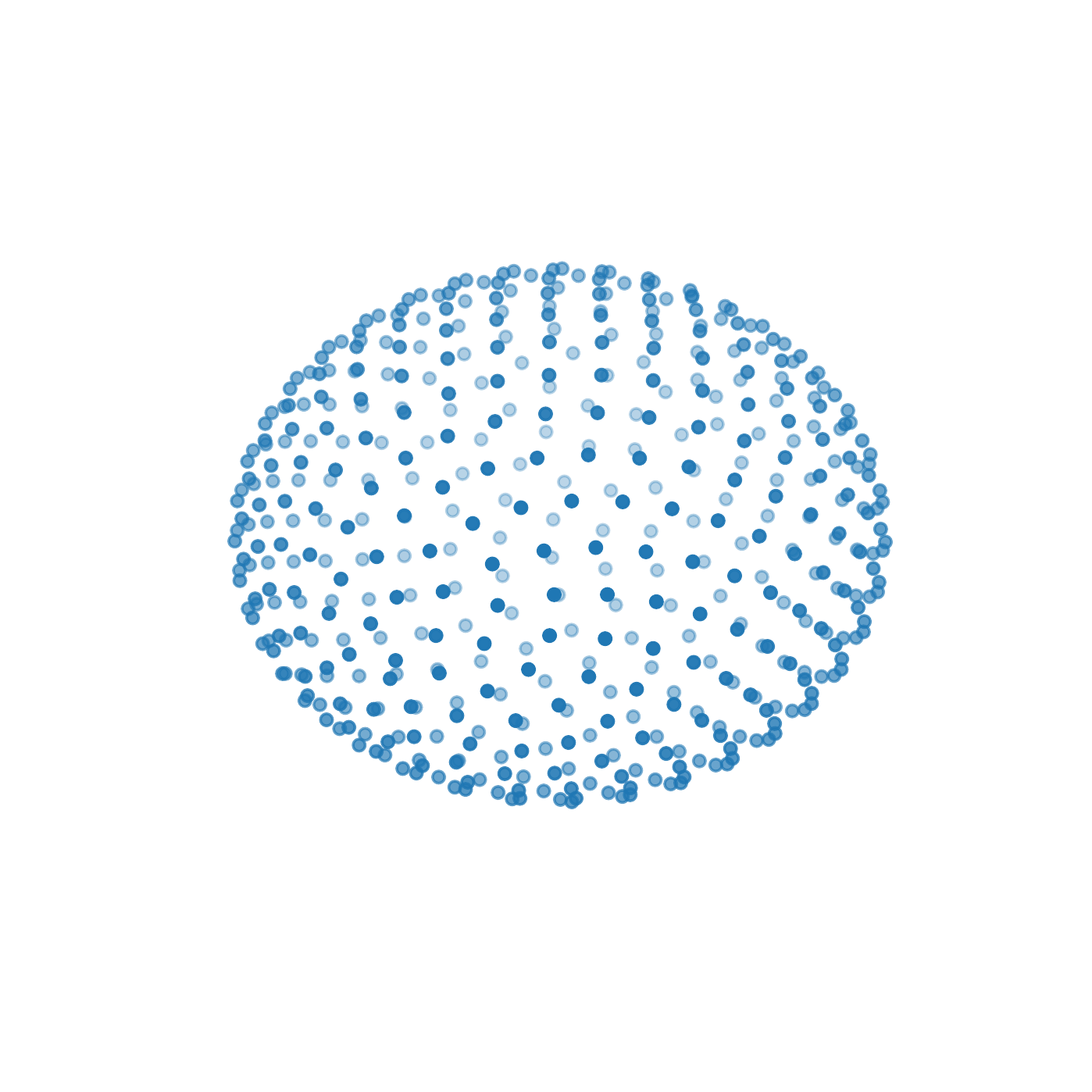}}
        \caption{500 points}
        \label{fig:fib_500}
    \end{subfigure}
    \hfill
    \begin{subfigure}[b]{0.23\textwidth}
        \centering
        {\includegraphics[trim={1cm 3cm 1cm 2cm},clip,width=\textwidth]{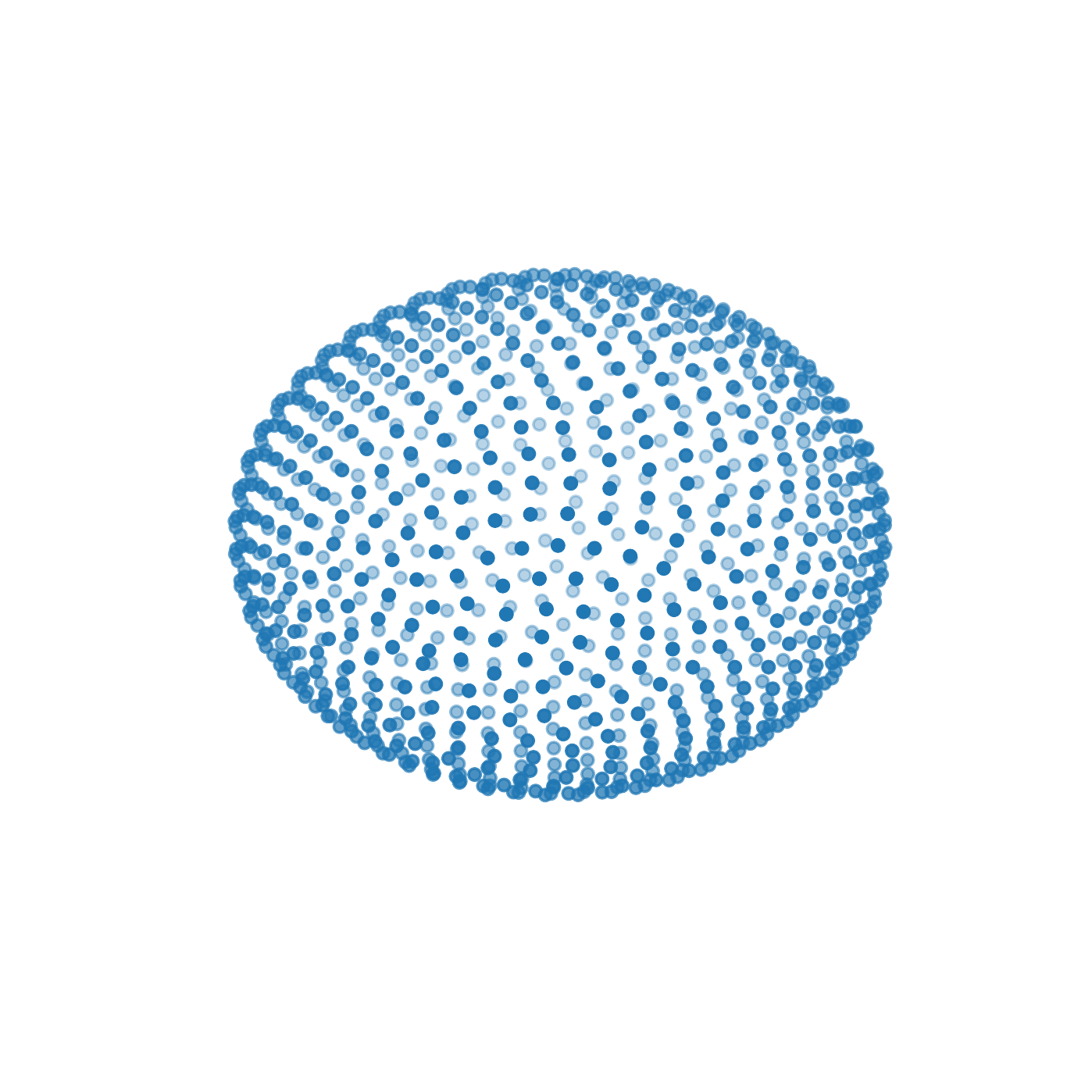}}
        \caption{1000 points}
        \label{fig:fib_1000}
    \end{subfigure}
    \caption{Fibonacci lattice point generation for different number of points. See \autoref{algo:fibonacci} for more details.}
    \label{fig:fib}
\end{figure}
We show the \ac{scf} iteration and the proposed globalization strategy in \autoref{algo:scf}. For convenience we restate the \ac{pnec} energy function
\begin{equation} \label{eqn:pnec-energy}
    E_P(\boldsymbol{R}, \boldsymbol{t}) = \sum_i \frac{e_i^2}{\sigma_i^2} = \sum_i \frac{| \boldsymbol{t}^\top (\boldsymbol{f}_i \times \boldsymbol{R} \boldsymbol{f}^\prime_i) |^2}{\boldsymbol{t}^\top \hat{\boldsymbol{f}_i} \boldsymbol{R} \boldsymbol{\Sigma}_i \boldsymbol{R}^\top \hat{\boldsymbol{f}_i}{}^\top \boldsymbol{t}}
\end{equation}
that we optimize over the translation $\ft$ using the \ac{scf} iteration. The Fibonacci-lattice-based point generation on the sphere in $\R^3$ is given in \autoref{algo:fibonacci}.

The main step of the \ac{scf} iteration is the construction of the $\f{E}$-matrix~\cite[Eqn. (2.3)]{binbuhaer2019optimizing}, which is a $3 \times 3$ symmetric matrix in our case
\begin{equation} \label{eqn:scf-E}
    \begin{aligned}
        \f{E}(\fR, \ft) & = \sum_i w_i \cdot \left( \ft^\top\f{B}_i \ft \cdot \f{A}_i - \ft^\top\f{A}_i \ft \cdot \f{B}_i \right) \,,                             \\
        \f{A}_i         & = \hff_i \fR \ff_i^\prime (\hff_i \fR \ff_i^\prime)^\top \,,                                                                            \\
        \f{B}_i         & = \hat{\boldsymbol{f}_i} \boldsymbol{R} \boldsymbol{\Sigma}_i \boldsymbol{R}^\top \hat{\boldsymbol{f}_i}{}^\top + c \boldsymbol{I}_3\,, \\
        w_i             & = (\ft^\top \f{B}_i \ft)^{-2} \cdot \prod_j \ft^\top \f{B}_j \ft \,.
    \end{aligned}
\end{equation}
In the \ac{scf} iteration the eigenvector corresponding to the largest eigenvalue is computed and thus, if necessary for numerical reasons, the weights can be scaled by a constant. Specifically, we use weights of the form ${w_i = (\ft^\top \f{B}_i \ft)^{-2}}$ because the product $\prod_j \ft^\top \f{B}_j \ft$ is common to all weights and leads to numerical issues. Due to the regularization, the matrices $\f{B}_i$ are positive definite, which is required for the \ac{scf} iteration. For experiments on synthetic data that follow \cite{EigenNEC_Kneip2013} and KITTI experiments the same regularization as for the overall \ac{pnec} optimization can be used for $c$. This ensures, that the matrices $\f{B}_i$ are positive definite.

\begin{algorithm}[tb]
    \setstretch{1.25}
    \footnotesize
    \SetAlgoLined
    \KwData{Fixed rotation $\tilde{\fR}$}
    \KwResult{Optimized translation $\ft^\ast$}

    \nl Sample the Fibonacci Lattice with $K$ points (cf.~\autoref{algo:fibonacci})\\
    $\{ \bar{\ft}_k \}_k \gets \FibonacciLattice(K)$\\[1.1ex]

    \nl Select the starting point with minimal Energy (cf.~\autoref{eqn:pnec-energy})\\
    $\ft_0 \gets \arg\min_{k} E_P(\tilde{\fR}, \bar{\ft}_k)$\\[1.1ex]

    \For{$s\gets1$ \KwTo $S$}{
        \nl Construct the $\f{E}$-matrix (cf.~\autoref{eqn:scf-E})\\
        $\f{E}_s \gets \f{E}(\tilde{\fR}, \bar{\ft}_{s-1})$\\[1.1ex]

        \nl Eigendecompose $\f{E}_s \in \R^{3 \times 3}$ using $\f{E}_s = \f{E}_s^\top$\\
        $\lambda_1, \lambda_2, \lambda_3, \f{v}_1, \f{v}_2, \f{v}_3 \gets \operatorname{eig}(\f{E}_s) \; s.t. \; \lambda_1 \leq \lambda_2 \leq \lambda_3$\\[1.1ex]

        \nl Set $\ft_s$ as the eigenvector with maximal eigenvalue\\
        $\ft_s \gets \f{v}_3$
    }

    \caption{\ac{scf} Optimization w/ Globalization}
    \label{algo:scf}
\end{algorithm}

\begin{algorithm}[tb]
    \setstretch{1.25}
    \footnotesize
    \SetAlgoLined
    \KwData{Number of points $K$}
    \KwResult{Points on the sphere $\{ \bar{\ft}_k = (x_k, y_k, z_k)\}_k$}

    \nl Compute the golden ratio angle\\
    $\phi \gets \pi \cdot (3 - \sqrt{5})$\\[1.1ex]

    \For{$k\gets1$ \KwTo $K$}{

        \nl Compute the $k$\textsuperscript{th} $y$-coordinate $y_k \in [-1, 1]$\\
        $y_k \gets 1 - 2\cdot\frac{k-1}{K-1}$\\[1.1ex]

        \nl Compute the radius in the $x$-$z$-plane\\
        $r_{xz} \gets \sqrt{1 - y_k^2}$\\[1.1ex]

        \nl Compute the remaining coordinates $x_k$, $z_k$ for $\bar{\ft}_k$\\
        $x_k \gets r_{xz} \cdot \cos((k-1)\phi)$\\
        $z_k \gets r_{xz} \cdot \sin((k-1)\phi)$\\
    }
    \caption{Fibonacci Lattice Point Generation}
    \label{algo:fibonacci}
\end{algorithm}

\section{Hyperparameters}\label{sec:Hyperparameters}
In the following we give an overview over the parameters we use for the simulated experiments and on the KITTI dataset \cite{Geiger2012CVPR}. We show both the parameters used in the \ac{pnec} optimization and the parameters used to generate the KLT tracks.

For \ac{pnec} we use: \textbf{Alg. 1 iterations} is the number of iterations we use in \autoref{algo:optimization_scheme} of the main paper before we start the least squares refinement; \textbf{SCF iterations} is the number of iteration we run \autoref{algo:scf} presented in this supplementary material in each optimization over $\boldsymbol{t}$; \textbf{Fibonacci lattice points} is the number of points we generate using the Fibonacci lattice; \textbf{regularization} is the regularization constant we proposed to avoid the singularities of the \ac{pnec}.

For the KLT parameters the parameters are: \textbf{pattern size} the pattern layout used by the KLT tracker\footnote{see \href{https://github.com/VladyslavUsenko/basalt-mirror/blob/master/include/basalt/optical_flow/patterns.h}{\texttt{include/basalt/optical\_flow/patterns.h}} in the implementation of \cite{Basalt_Usenko2020}}; \textbf{grid size} is the length of each square for which a track is extracted; \textbf{pyramid-levels} is the number of pyramid levels over which tracks are tracked, where the scale factor between each pyramid level is 2; \textbf{optical flow iterations} is the number of iterations tracking is done on each pyramid level; \textbf{optical flow max recovered distance} is the maximum distance a track can have from its original position after forward-backward tracking (otherwise it is discarded).

For RANSAC\footnote{see \href{https://laurentkneip.github.io/opengv/page_how_to_use.html\#sec_threshold}{OpenGV} for details of the RANSAC scheme}
the parameters are: \textbf{iterations} the maximum number of iterations for the RANSAC scheme; \textbf{threshold} the threshold for a point to be classified as an inlier.
\begin{table}[t]
    \small
    \centering
    \sisetup{detect-weight,mode=text}
    \renewrobustcmd{\bfseries}{\fontseries{b}\selectfont}
    \renewrobustcmd{\boldmath}{}
    \newrobustcmd{\B}{\bfseries}
    \addtolength{\tabcolsep}{-3.5pt}

    \begin{tabular} {p{3.6cm} c|c}
        \toprule
        Hyperparameter                      & Simulated  & KITTI      \\
        \midrule
        Alg. 1 iterations                   & 10         & 10         \\
        SCF itertations                     & 10         & 10         \\
        Fibonacci lattice points            & 500        & 500        \\
        regularization                      & $10^{-10}$ & $10^{-13}$ \\
        \midrule
        KLT parameters
                                            &            &            \\
        \midrule
        pattern size                        &            & 52         \\
        grid size                           &            & 30         \\
        pyramid-levels                      &            & 4          \\
        optical flow iterations             &            & 40         \\
        optical flow max recovered distance &            & 0.04       \\
        \midrule
        RANSAC parameters                   &            &            \\
        \midrule
        iterations                          &            & 5000       \\
        threshold                           &            & $10^{-6}$  \\
        \bottomrule
    \end{tabular}
    \caption{Parameters used for our experiments.}
    \label{tab:parameters}
\end{table}

\subsection{Hyperparameter Study}
\autoref{tab:hyperparameter-study} shows a the results for different hyperparameter settings on the synthetic data with anisotropic and inhomogeneous noise for different noise levels. We compare the \ac{pnec} and the \ac{nec} from the main paper and the following: sampling the Fibonacci lattice only on the first iteration of the optimization; only doing 3 iterations of the SCF algorithm after the first iteration of the optimization; sampling 5000 points with the Fibonacci lattice; increasing the regularization to $c=10^{-5}$ and $c=10^{-0}$. The changes to the Fibonacci lattice and the SCF algorihm lead to the same results for the \ac{pnec}. Increasing the regularization leads to worse results especially in the rotation estimation. A higher regularization leads to a more equal weighting of the residuals, the \ac{pnec} approaches the orignal \ac{nec}.

\begin{table*}[t]
    \small
    \centering
    \sisetup{detect-weight,mode=text}
    \renewrobustcmd{\bfseries}{\fontseries{b}\selectfont}
    \renewrobustcmd{\boldmath}{}
    \newrobustcmd{\B}{\bfseries}
    \addtolength{\tabcolsep}{-3.5pt}
    \begin{tabular} {p{3.1cm} r r  r r  r r|r  r  r|r r  r r  r r|r  r  r  r}
        \toprule
                                                            & \multicolumn{9}{c}{\scshape Omnidirectional} & \multicolumn{9}{c}{\scshape Pinhole} &                                                                                                                                                                                                                                                                                                                                                                                                                                \\[1.1ex]
                                                            & \multicolumn{6}{c}{\scshape w/ t}            & \multicolumn{3}{c}{\scshape w/o t}   & \multicolumn{6}{c}{\scshape w/ t} & \multicolumn{3}{c}{\scshape w/o t} &                                                                                                                                                                                                                                                                                                                                                       \\
        Noise level [px]                                    &
        \multicolumn{2}{c}{\scshape 0.5}                    & \multicolumn{2}{c}{\scshape 1.0}             & \multicolumn{2}{c}{\scshape 1.5}     & {\scshape 0.5}                    & {\scshape 1.0}                     & {\scshape 1.5}     & \multicolumn{2}{c}{\scshape 0.5} & \multicolumn{2}{c}{\scshape 1.0} & \multicolumn{2}{c}{\scshape 1.5} &
        {\scshape 0.5}                                      & {\scshape 1.0}                               & {\scshape 1.5}                                                                                                                                                                                                                                                                                                                                                                                                                                                        \\
        Metric [degree]                                     & {$e_{\text{rot}}$}                           & {$e_{\text{t}}$}                     & {$e_{\text{rot}}$}                & {$e_{\text{t}}$}                   & {$e_{\text{rot}}$} & {$e_{\text{t}}$}                 & {$e_{\text{rot}}$}               & {$e_{\text{rot}}$}               & {$e_{\text{rot}}$} & {$e_{\text{rot}}$} & {$e_{\text{t}}$} & {$e_{\text{rot}}$} & {$e_{\text{t}}$} & {$e_{\text{rot}}$} & {$e_{\text{t}}$} & {$e_{\text{rot}}$} & {$e_{\text{rot}}$} & {$e_{\text{rot}}$} &             \\
        \midrule
        PNEC                                                & 0.08                                         & 1.29                                 & 0.12                              & 1.60                               & 0.14               & 1.66                             & 0.09                             & 0.13                             & 0.15               & 0.20               & 2.06             & 0.28               & 2.38             & 0.34               & 2.54             & 0.15               & 0.21               & 0.25               &             \\
        NEC                                                 & 0.11                                         & 1.90                                 & 0.15                              & 2.10                               & 0.17               & 2.11                             & 0.11                             & 0.15                             & 0.18               & 0.25               & 2.41             & 0.34               & 2.78             & 0.41               & 2.91             & 0.19               & 0.25               & 0.29               & \cu{+24 \%} \\
        Fib. lattice only for 1\textsuperscript{st}         & 0.08                                         & 1.30                                 & 0.12                              & 1.60                               & 0.14               & 1.66                             & 0.09                             & 0.13                             & 0.15               & 0.20               & 2.05             & 0.28               & 2.37             & 0.34               & 2.56             & 0.15               & 0.21               & 0.25               & \cd{0 \%}   \\
        Only 3 SCF iter. after 1\textsuperscript{st}        & 0.08                                         & 1.32                                 & 0.12                              & 1.60                               & 0.14               & 1.68                             & 0.09                             & 0.13                             & 0.15               & 0.20               & 2.05             & 0.28               & 2.37             & 0.34               & 2.55             & 0.15               & 0.21               & 0.25               & \cd{0 \%}   \\
        Fib. lattice $5000$ pts                             & 0.08                                         & 1.29                                 & 0.12                              & 1.60                               & 0.14               & 1.66                             & 0.09                             & 0.13                             & 0.15               & 0.20               & 2.06             & 0.28               & 2.38             & 0.34               & 2.54             & 0.15               & 0.21               & 0.25               & \cd{0 \%}   \\
        Reg. $\hphantom{\uparrow}\uparrow$ \; ($c=10^{-5}$) & 0.10                                         & 1.23                                 & 0.14                              & 1.54                               & 0.16               & 1.66                             & 0.10                             & 0.14                             & 0.17               & 0.24               & 2.32             & 0.33               & 2.91             & 0.40               & 3.11             & 0.12               & 0.16               & 0.19               & \cu{+6 \%}  \\
        Reg. ${\uparrow}\uparrow$ \; ($c=10^{0}$)           & 0.10                                         & 1.23                                 & 0.14                              & 1.54                               & 0.16               & 1.66                             & 0.10                             & 0.14                             & 0.17               & 0.24               & 2.32             & 0.33               & 2.91             & 0.40               & 3.11             & 0.12               & 0.16               & 0.19               & \cu{+6 \%}  \\
        \bottomrule
    \end{tabular}
    \caption{Hyperparameter study on the synthetic data. We compare: the \ac{pnec} and the \ac{nec} from the main paper; sampling the Fibonacci lattice only on the first iteration of the optimization; only doing 3 iterations of the SCF algorithm after the first iteration of the optimization; sampling 5000 points with the Fibonacci lattice; increasing the regularization to $c=10^{-5}$ and $c=10^{-0}$. The changes to the Fibonacci lattice and the SCF algorihm lead to the same results for the \ac{pnec}. Increasing the regularization leads to worse results especially in the rotation estimation. A higher regularization leads to a more equal weighting of the residuals, the \ac{pnec} approaches the orignal \ac{nec}.}
    \label{tab:hyperparameter-study}
\end{table*}

\section{Experiments with Simulated Data}\label{sec:simulated}
\begin{figure}
    \centering
    \begin{subfigure}[b]{0.23\textwidth}
        \centering
        {\includegraphics[trim={24cm 14cm 24cm 14cm},clip,width=\textwidth]{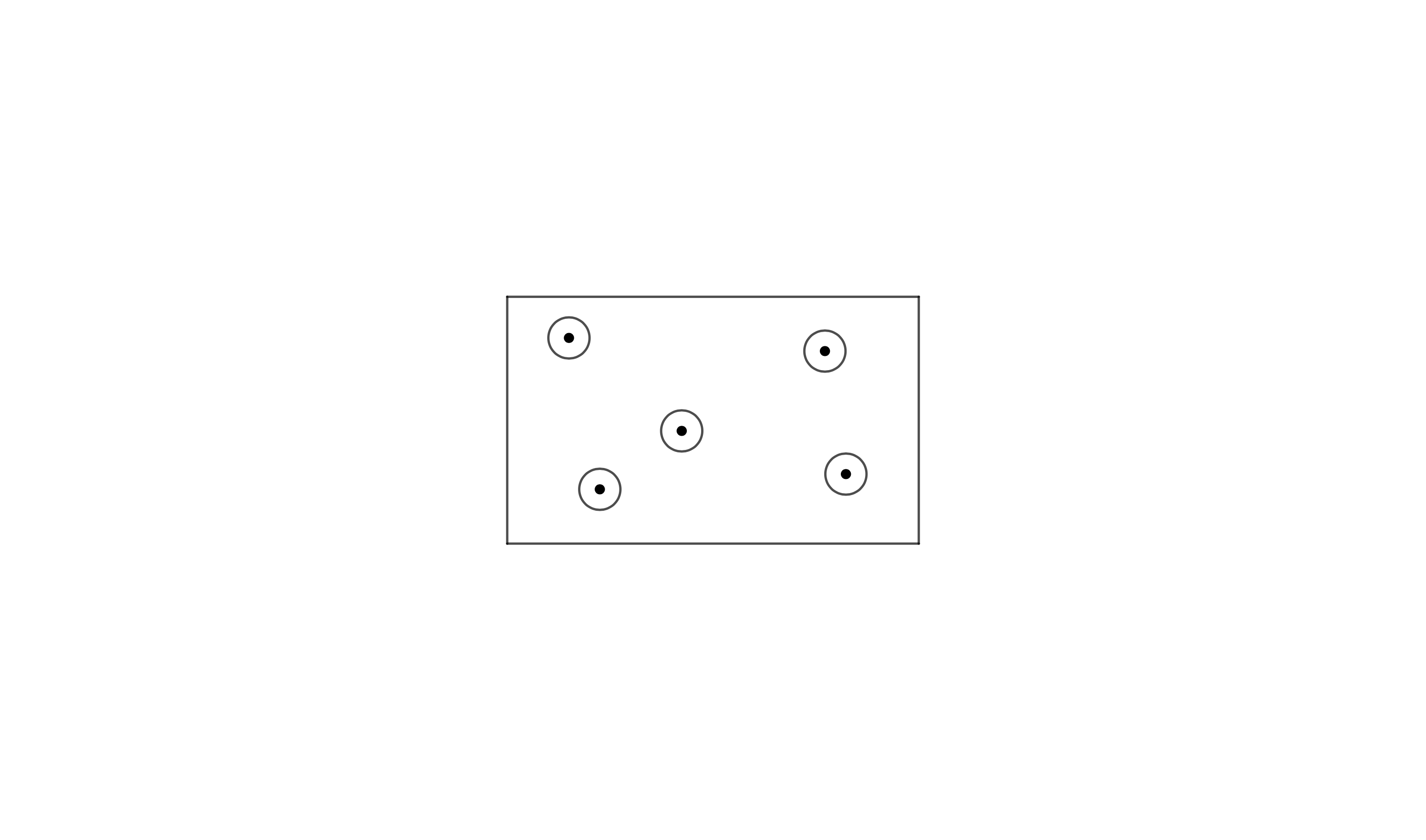}}
        \caption{isotropic homogeneous}
        \label{fig:noise_0}
    \end{subfigure}
    \hfill
    \begin{subfigure}[b]{0.23\textwidth}
        \centering
        {\includegraphics[trim={24cm 14cm 24cm 14cm},clip,width=\textwidth]{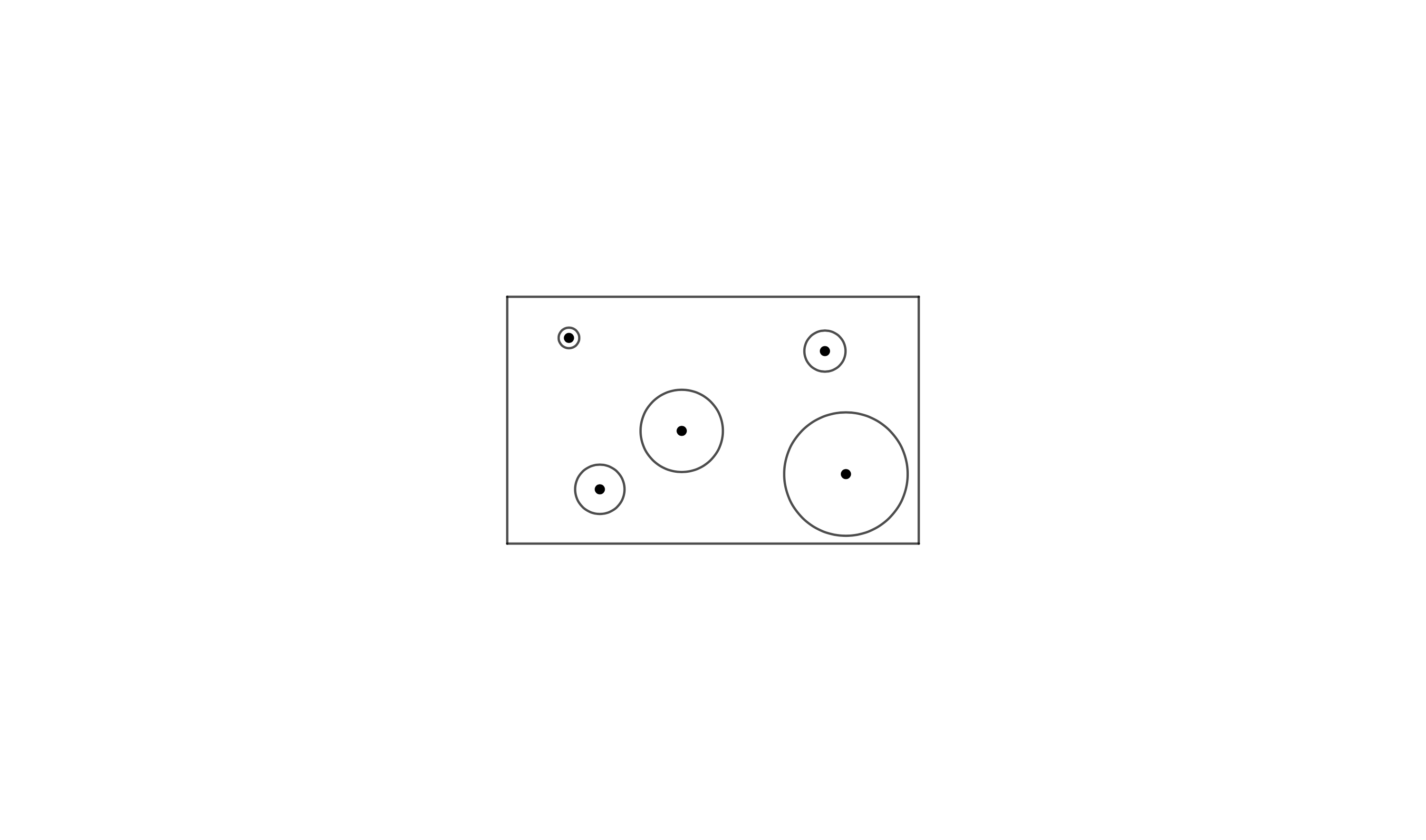}}
        \caption{isotropic inhomogeneous}
        \label{fig:noise_1}
    \end{subfigure}
    \hfill
    \begin{subfigure}[b]{0.23\textwidth}
        \centering
        {\includegraphics[trim={24cm 14cm 24cm 14cm},clip,width=\textwidth]{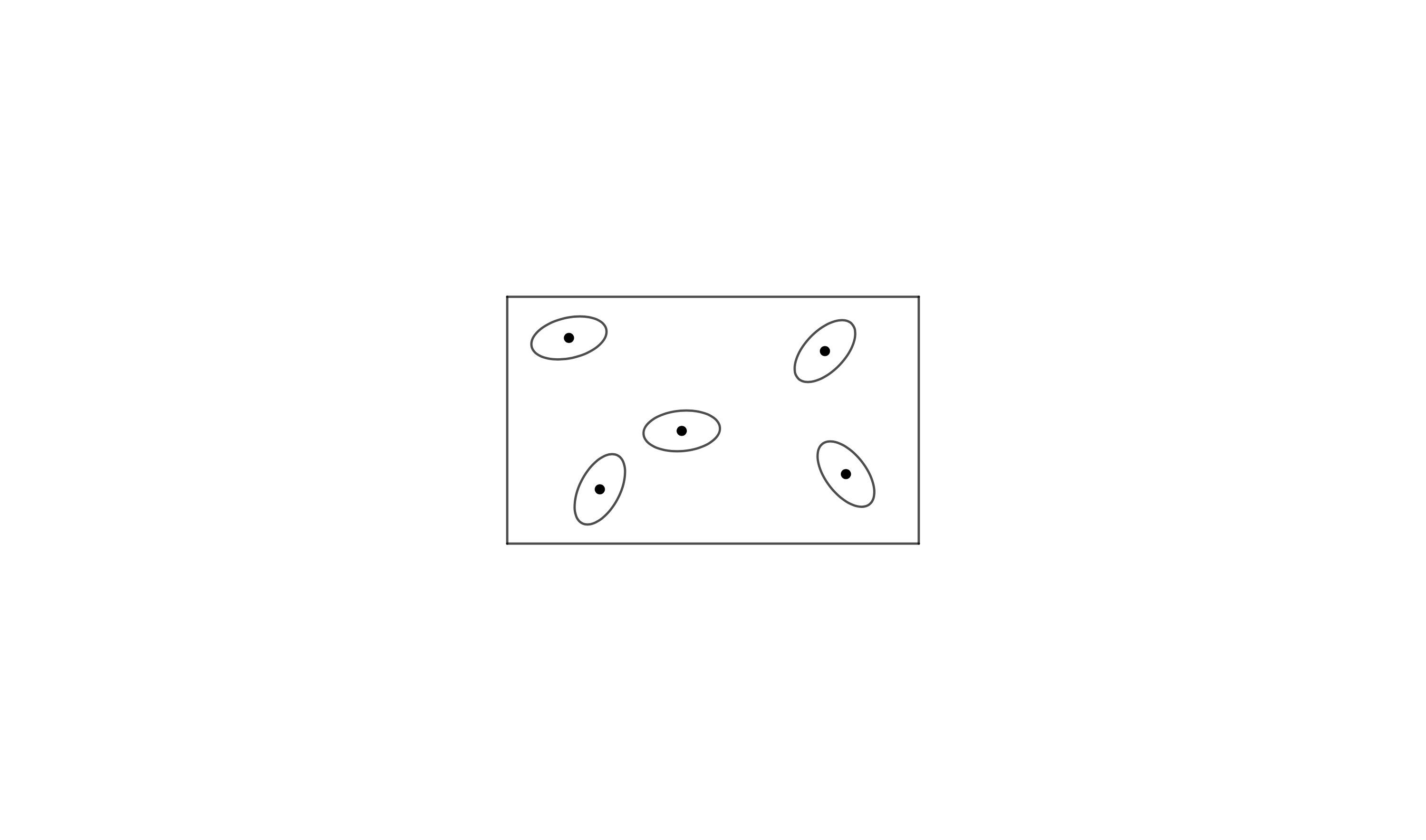}}
        \caption{anisotropic homogeneous}
        \label{fig:noise_2}
    \end{subfigure}
    \hfill
    \begin{subfigure}[b]{0.23\textwidth}
        \centering
        {\includegraphics[trim={24cm 14cm 24cm 14cm},clip,width=\textwidth]{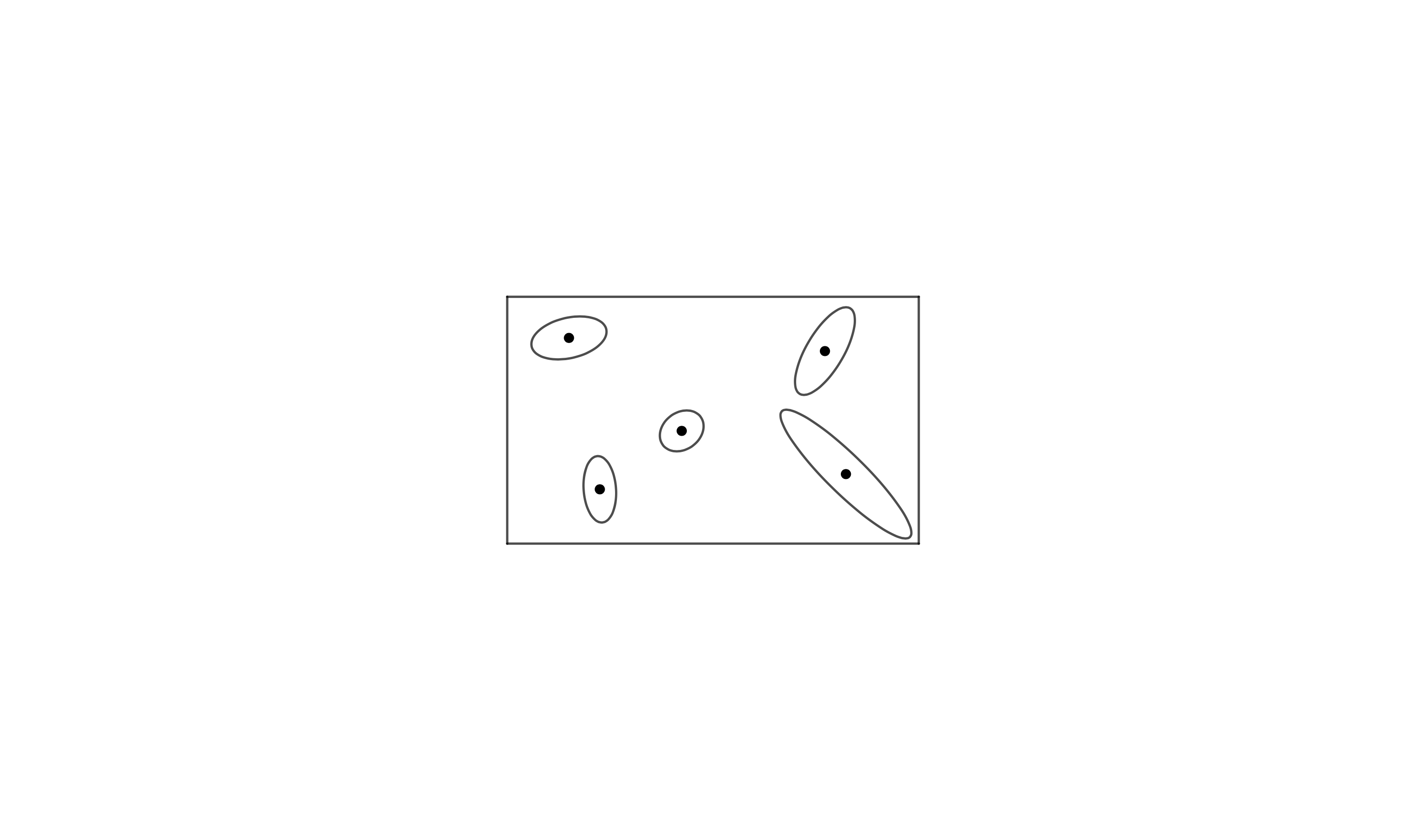}}
        \caption{anisotropic inhomogeneous}
        \label{fig:noise_3}
    \end{subfigure}
    \caption{Illustration of different noise types based on Brooks \etal~\cite{pro_cov_Brooks2001}.}
    \label{fig:noise}
\end{figure}
We present the results for the simulated experiments for other noise types. \autoref{fig:noise} shows an illustration of the noise type classification by Brooks \etal~\cite{pro_cov_Brooks2001} on which we base our experiments. As for the {\em anisotropic inhomogonoeous} noise we present the average results for {\em isotropic homogonoeous}, {\em isotropic inhomogonoeous}, and {\em anisotropic homogonoeous} noise over $10\,000$ random instantiations.

The covariance matrices for the simulated experiments are generated using the following parameterization
\begin{equation}
    \boldsymbol{\Sigma}_{2\text{D}} = s  \boldsymbol{R}_{\alpha} \begin{pmatrix} \beta & 0 \\ 0 & 1 - \beta \end{pmatrix} \boldsymbol{R}_{\alpha}^\top
\end{equation}
with a scaling factor $s$, an anisotropy term $\beta$, and a $2\text{D}$ rotation matrix
\begin{equation}
    \boldsymbol{R}_{\alpha} = \begin{pmatrix} \cos{\alpha} & -\sin{\alpha} \\ \sin{\alpha} & \cos{\alpha} \end{pmatrix}.
\end{equation}
The parameters for {\em isotropic homogeneous} noise are $s = 1$, $\beta = 0.5$, and $\alpha = 0$. For {\em isotropic inhomogeneous} noise they are $\beta = 0.5$, $\alpha = 0$, and $s$ is sampled uniformly between $0.5$ and $1.5$ for each covariance. For {\em anisotropic homogeneous} noise $s = 1$, $\beta$ is sampled uniformly between $0.5$ and $1$ once for each experiment, and $\alpha$ is sampled uniformly between $0$ and $\pi$ for each covariance. For {\em anisotropic inhomogeneous} noise all parameters are uniformly sampled for each covariance, $s$ between $0.5$ and $1.5$, $\beta$  between $0.5$ and $1$, and $\alpha$ between $0$ and $\pi$.

\subsection{Omnidirectional Cameras}
\begin{figure*}[t]
    \centering
    \begin{subfigure}[b]{0.32\textwidth}
        \centering
        {\includegraphics[width=\textwidth]{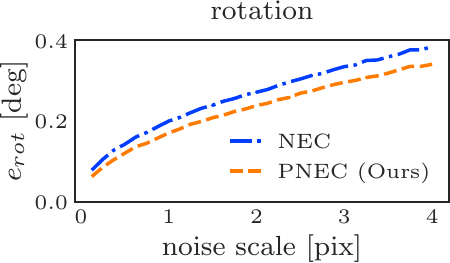}}
        \caption{Experiment w/ translation: $e_{rot}$}
        \label{fig:omni_0_rot}
    \end{subfigure}
    \hfill
    \begin{subfigure}[b]{0.32\textwidth}
        \centering
        {\includegraphics[width=\textwidth]{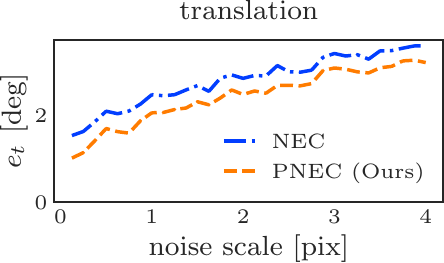}}
        \caption{Experiment w/ translation: $e_{t}$}
        \label{fig:omni_0_t}
    \end{subfigure}
    \hfill
    \begin{subfigure}[b]{0.32\textwidth}
        \centering
        {\includegraphics[width=\textwidth]{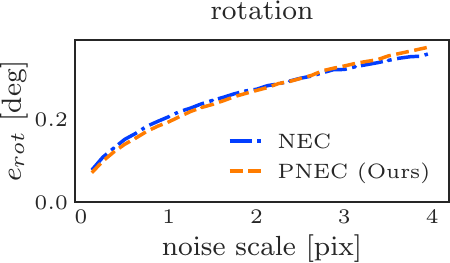}}
        \caption{Experiment w/o translation: $e_{rot}$}
        \label{fig:omni_0_no_t}
    \end{subfigure}
    \caption{\textbf{isotropic homogeneous}: Experiments for omnidirectional cameras. The \ac{pnec} outperforms the \ac{nec} for experiments with translation, without translation both methods perform similar. The results show that next to the shape of the covariances the geometry also influences the variance of the residuals, resulting in better rotation estimation of the \ac{pnec}.}
    \label{fig:omni_0}
\end{figure*}

\begin{figure*}[t]
    \centering
    \begin{subfigure}[b]{0.32\textwidth}
        \centering
        {\includegraphics[width=\textwidth]{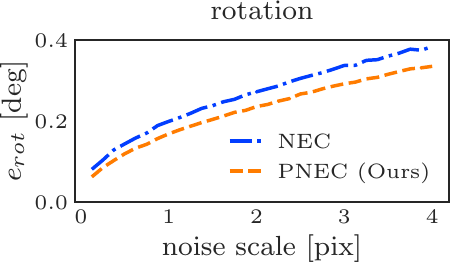}}
        \caption{Experiment w/ translation: $e_{rot}$}
        \label{fig:omni_1_rot}
    \end{subfigure}
    \hfill
    \begin{subfigure}[b]{0.32\textwidth}
        \centering
        {\includegraphics[width=\textwidth]{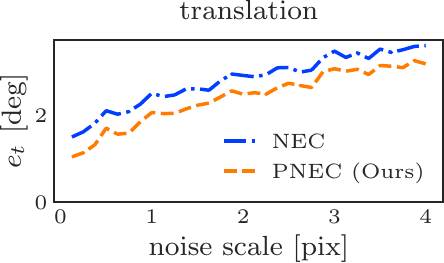}}
        \caption{Experiment w/ translation: $e_{t}$}
        \label{fig:omni_1_t}
    \end{subfigure}
    \hfill
    \begin{subfigure}[b]{0.32\textwidth}
        \centering
        {\includegraphics[width=\textwidth]{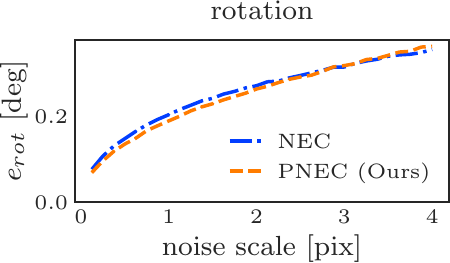}}
        \caption{Experiment w/o translation: $e_{rot}$}
        \label{fig:omni_1_no_t}
    \end{subfigure}
    \caption{\textbf{isotropic inhomogeneous}: Experiments for omnidirectional cameras. The \ac{pnec} outperforms the \ac{nec} for experiments with translation, without translation the \ac{pnec} only performs slightly better. As expected the inhomogeneity of the covariance matrices is beneficial for the \ac{pnec}.}
    \label{fig:omni_1}
\end{figure*}

\begin{figure*}[t]
    \centering
    \begin{subfigure}[b]{0.32\textwidth}
        \centering
        {\includegraphics[width=\textwidth]{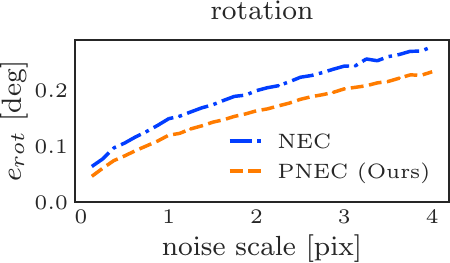}}
        \caption{Experiment w/ translation: $e_{rot}$}
        \label{fig:omni_2_rot}
    \end{subfigure}
    \hfill
    \begin{subfigure}[b]{0.32\textwidth}
        \centering
        {\includegraphics[width=\textwidth]{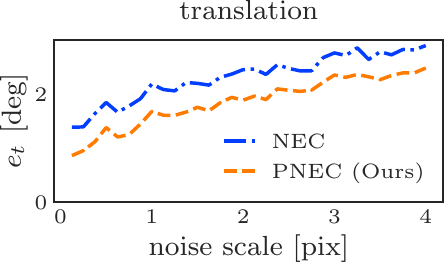}}
        \caption{Experiment w/ translation: $e_{t}$}
        \label{fig:omni_2_t}
    \end{subfigure}
    \hfill
    \begin{subfigure}[b]{0.32\textwidth}
        \centering
        {\includegraphics[width=\textwidth]{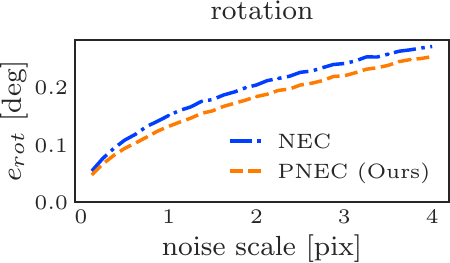}}
        \caption{Experiment w/o translation: $e_{rot}$}
        \label{fig:omni_2_no_t}
    \end{subfigure}
    \caption{\textbf{anisotropic homogeneous}: Experiments for omnidirectional cameras. The \ac{pnec} outperforms the \ac{nec} for experiments with translation and without translation. This experiment shows the importance the directional information of the anisotropic covariances provides the \ac{pnec}. The performance difference is significantly bigger than for isotropic noise.}
    \label{fig:omni_2}
\end{figure*}

\autoref{fig:omni_0}, \autoref{fig:omni_1}, and \autoref{fig:omni_2} show the results for omnidirectional cameras. The \ac{pnec} consistently achieves better results than the \ac{nec} on all noise types. Notable is that even for \textbf{isotropic homogeneous} noise the \ac{pnec} ouperforms the \ac{nec}. This experiment shows that the geometry of the problem has an influence on the energy function. Although all covariances are equal in the image plane the same does not hold for the variance of each residual. The experiments for \textbf{isotropic inhomogeneous} and \textbf{anisotropic homogeneous} noise show the benefit the differnent sizes and shape of covariances has, respectively. Both widen the gap between the \ac{pnec} and \ac{nec}. %

\subsection{Pinhole Cameras}
\begin{figure*}[t]
    \centering
    \begin{subfigure}[b]{0.32\textwidth}
        \centering
        {\includegraphics[width=\textwidth]{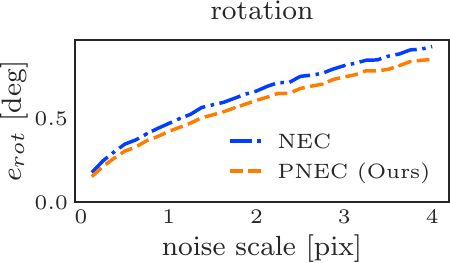}}
        \caption{Experiment w/ translation: $e_{rot}$}
        \label{fig:pin_0_rot}
    \end{subfigure}
    \hfill
    \begin{subfigure}[b]{0.31\textwidth}
        \centering
        {\includegraphics[width=\textwidth]{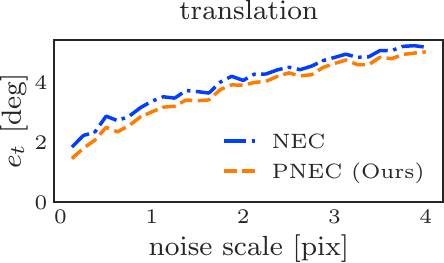}}
        \caption{Experiment w/ translation: $e_{t}$}
        \label{fig:pin_0_t}
    \end{subfigure}
    \hfill
    \begin{subfigure}[b]{0.32\textwidth}
        \centering
        {\includegraphics[width=\textwidth]{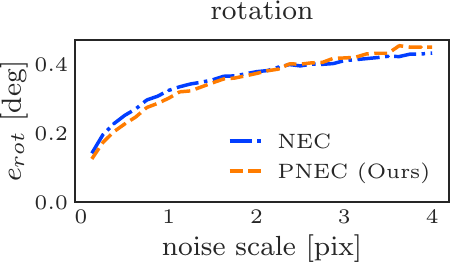}}
        \caption{Experiment w/o translation: $e_{rot}$}
        \label{fig:pin_0_no_t}
    \end{subfigure}
    \caption{\textbf{isotropic homogeneous}: Experiments for pinhole cameras. Similar to omnidirectional cameras the \ac{pnec} outperforms the \ac{nec} even for isotropic homogeneous noise. However, for purely rotational experiments the \ac{pnec} performs slightly worse for experiments with a high noise level.}
    \label{fig:pin_0}
\end{figure*}

\begin{figure*}[t]
    \centering
    \begin{subfigure}[b]{0.32\textwidth}
        \centering
        {\includegraphics[width=\textwidth]{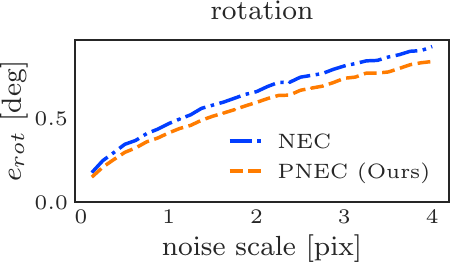}}
        \caption{Experiment w/ translation: $e_{rot}$}
        \label{fig:pin_1_rot}
    \end{subfigure}
    \hfill
    \begin{subfigure}[b]{0.31\textwidth}
        \centering
        {\includegraphics[width=\textwidth]{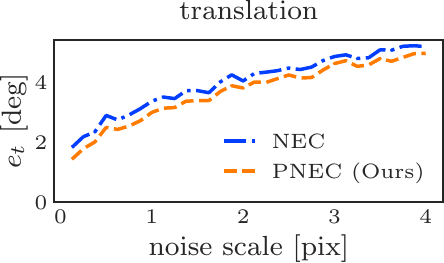}}
        \caption{Experiment w/ translation: $e_{t}$}
        \label{fig:pin_1_t}
    \end{subfigure}
    \hfill
    \begin{subfigure}[b]{0.32\textwidth}
        \centering
        {\includegraphics[width=\textwidth]{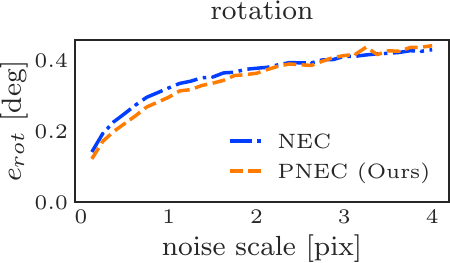}}
        \caption{Experiment w/o translation: $e_{rot}$}
        \label{fig:pin_1_no_t}
    \end{subfigure}
    \caption{\textbf{isotropic inhomogeneous}: Experiments for pinhole cameras. Similar to omnidirectional cameras the inhomogeneity helps the \ac{pnec}.  However, for purely rotational experiments both methods performs similar for experiments with a high noise level.}
    \label{fig:pin_1}
\end{figure*}

\begin{figure*}[t]
    \centering
    \begin{subfigure}[b]{0.32\textwidth}
        \centering
        {\includegraphics[width=\textwidth]{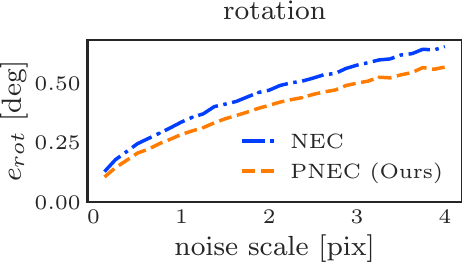}}
        \caption{Experiment w/ translation: $e_{rot}$}
        \label{fig:pin_2_rot}
    \end{subfigure}
    \hfill
    \begin{subfigure}[b]{0.31\textwidth}
        \centering
        {\includegraphics[width=\textwidth]{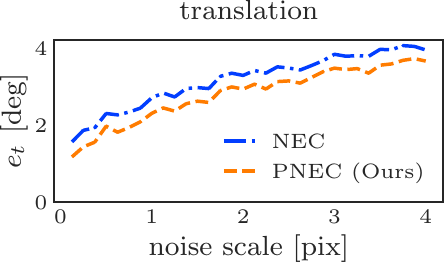}}
        \caption{Experiment w/ translation: $e_{t}$}
        \label{fig:pin_2_t}
    \end{subfigure}
    \hfill
    \begin{subfigure}[b]{0.32\textwidth}
        \centering
        {\includegraphics[width=\textwidth]{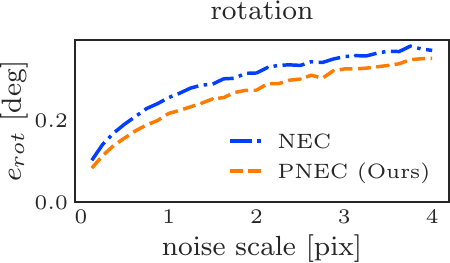}}
        \caption{Experiment w/o translation: $e_{rot}$}
        \label{fig:pin_2_no_t}
    \end{subfigure}
    \caption{\textbf{anisotropic homogeneous}: Experiments for pinhole cameras. Similar to omnidirectional cameras the directional information of the anisotropic covariances helps the \ac{pnec} significantly, especially in the zero translation case.}
    \label{fig:pin_2}
\end{figure*}

\autoref{fig:pin_0}, \autoref{fig:pin_1}, and \autoref{fig:pin_2} show the results for pinhole cameras. They show similar results to the omnidirectional cameras.

\section{Energy Function Results}\label{sec:energy}
\begin{figure*}
    \centering
    \begin{subfigure}[b]{0.245\textwidth}
        \centering
        {\includegraphics[width=\textwidth]{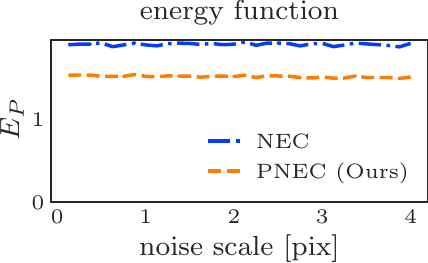}}
        \caption{}
        \label{fig:energy_omni_t_0}
    \end{subfigure}
    \hfill
    \begin{subfigure}[b]{0.245\textwidth}
        \centering
        {\includegraphics[width=\textwidth]{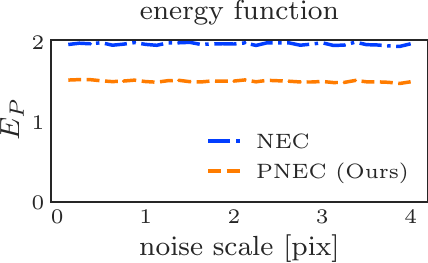}}
        \caption{}
        \label{fig:energy_omni_t_1}
    \end{subfigure}
    \hfill
    \begin{subfigure}[b]{0.245\textwidth}
        \centering
        {\includegraphics[width=\textwidth]{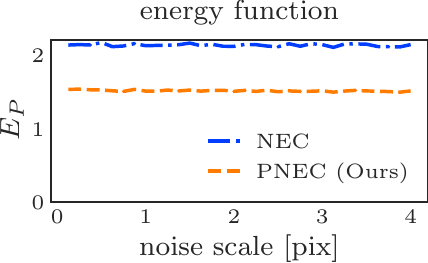}}
        \caption{}
        \label{fig:energy_omni_t_2}
    \end{subfigure}
    \hfill
    \begin{subfigure}[b]{0.245\textwidth}
        \centering
        {\includegraphics[width=\textwidth]{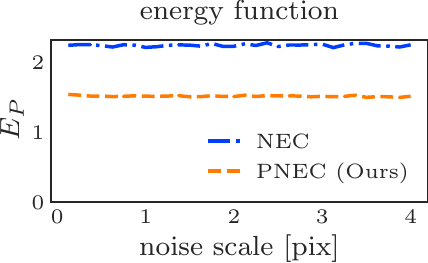}}
        \caption{}
        \label{fig:energy_omni_t_3}
    \end{subfigure}

    \begin{subfigure}[b]{0.245\textwidth}
        \centering
        {\includegraphics[width=\textwidth]{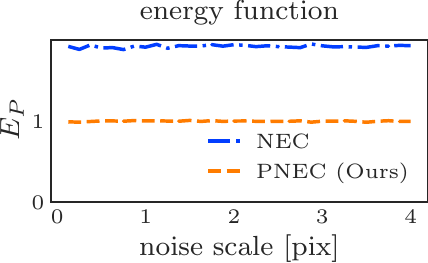}}
        \caption{}
        \label{fig:energy_omni_no_t_0}
    \end{subfigure}
    \hfill
    \begin{subfigure}[b]{0.245\textwidth}
        \centering
        {\includegraphics[width=\textwidth]{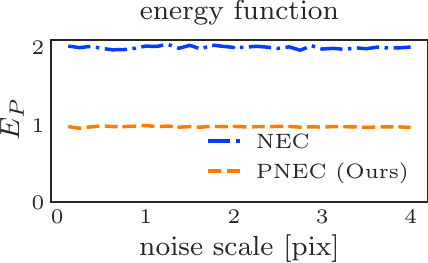}}
        \caption{}
        \label{fig:energy_omni_no_t_1}
    \end{subfigure}
    \hfill
    \begin{subfigure}[b]{0.245\textwidth}
        \centering
        {\includegraphics[width=\textwidth]{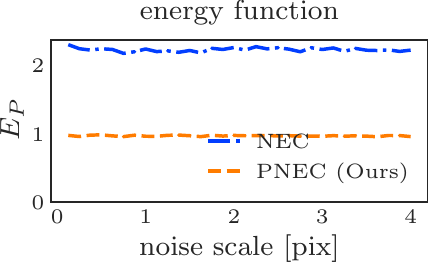}}
        \caption{}
        \label{fig:energy_omni_no_t_2}
    \end{subfigure}
    \hfill
    \begin{subfigure}[b]{0.245\textwidth}
        \centering
        {\includegraphics[width=\textwidth]{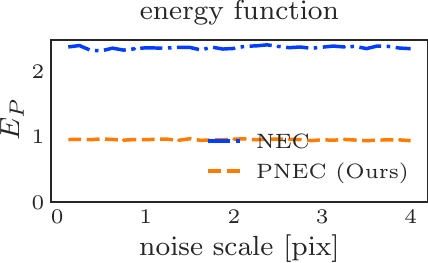}}
        \caption{}
        \label{fig:energy_omni_no_t_3}
    \end{subfigure}

    \begin{subfigure}[b]{0.245\textwidth}
        \centering
        {\includegraphics[width=\textwidth]{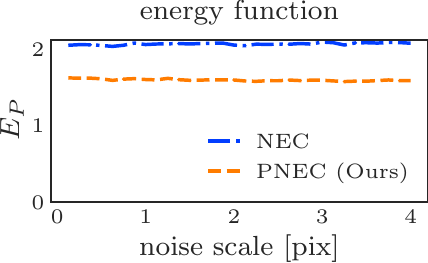}}
        \caption{}
        \label{fig:energy_pin_t_0}
    \end{subfigure}
    \hfill
    \begin{subfigure}[b]{0.245\textwidth}
        \centering
        {\includegraphics[width=\textwidth]{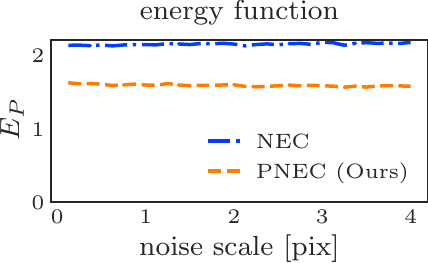}}
        \caption{}
        \label{fig:energy_pin_t_1}
    \end{subfigure}
    \hfill
    \begin{subfigure}[b]{0.245\textwidth}
        \centering
        {\includegraphics[width=\textwidth]{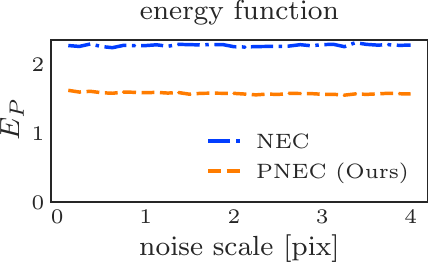}}
        \caption{}
        \label{fig:energy_pin_t_2}
    \end{subfigure}
    \hfill
    \begin{subfigure}[b]{0.245\textwidth}
        \centering
        {\includegraphics[width=\textwidth]{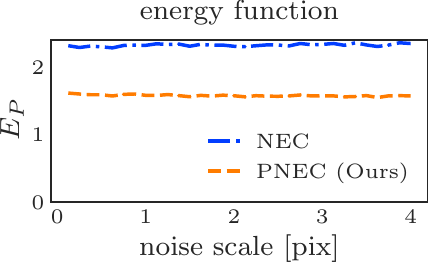}}
        \caption{}
        \label{fig:energy_pin_t_3}
    \end{subfigure}

    \begin{subfigure}[b]{0.245\textwidth}
        \centering
        {\includegraphics[width=\textwidth]{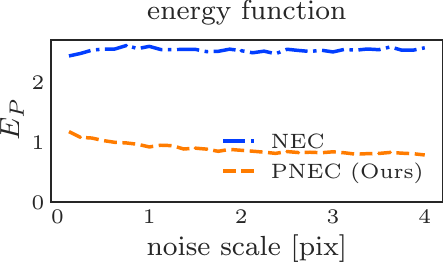}}
        \caption{}
        \label{fig:energy_pin_no_t_0}
    \end{subfigure}
    \hfill
    \begin{subfigure}[b]{0.245\textwidth}
        \centering
        {\includegraphics[width=\textwidth]{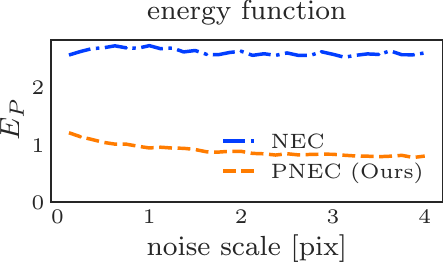}}
        \caption{}
        \label{fig:energy_pin_no_t_1}
    \end{subfigure}
    \hfill
    \begin{subfigure}[b]{0.245\textwidth}
        \centering
        {\includegraphics[width=\textwidth]{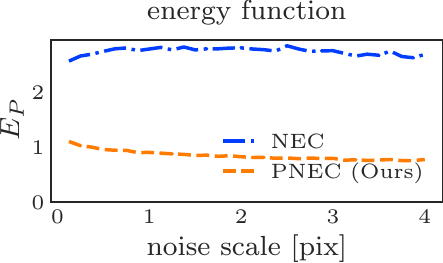}}
        \caption{}
        \label{fig:energy_pin_no_t_2}
    \end{subfigure}
    \hfill
    \begin{subfigure}[b]{0.245\textwidth}
        \centering
        {\includegraphics[width=\textwidth]{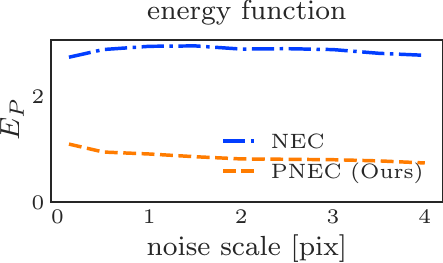}}
        \caption{}
        \label{fig:energy_pin_no_t_3}
    \end{subfigure}

    \caption{The energy function values for all simulated experiments. We present the median instead of the average value for the energy function, due to its volatility. The first column shows the experiments for {\em isotropic homogeneous} noise, the second for {\em isotropic inhomogeneous} noise, the third for {\em anisotropic homogeneous} noise, and the fourth for {\em anisotropic inhomogeneous} noise. The first two rows show the experiments for omnidirectional cameras with and without translation, respectively. The last two rows show the experiments for pinhole cameras with and without translation, respectively. The results show the effectiveness of our proposed optimization scheme to minimize the energy function.}
    \label{fig:energy}
    \vspace{5cm}
\end{figure*}

\autoref{fig:energy} shows the median energy values for all the simulated experiments presented in the main paper and this supplementary material. We show the median values instead of the average due to the volatility of the energy function.

\section{Anisotropy Experiments} \label{sec:anisotropy}
\begin{figure*}[t]
    \centering
    \begin{subfigure}[b]{0.32\textwidth}
        \centering
        {\includegraphics[width=\textwidth]{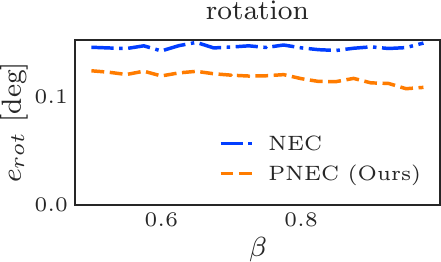}}
        \caption{Experiment w/ translation: $e_{rot}$}
        \label{fig:omni_aniso_rot}
    \end{subfigure}
    \hfill
    \begin{subfigure}[b]{0.31\textwidth}
        \centering
        {\includegraphics[width=\textwidth]{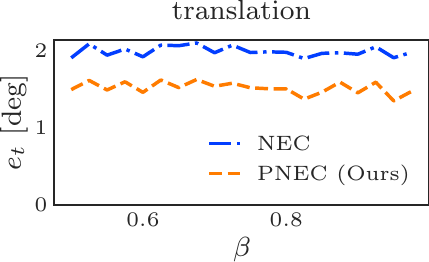}}
        \caption{Experiment w/ translation: $e_{t}$}
        \label{fig:omni_aniso_t}
    \end{subfigure}
    \hfill
    \begin{subfigure}[b]{0.32\textwidth}
        \centering
        {\includegraphics[width=\textwidth]{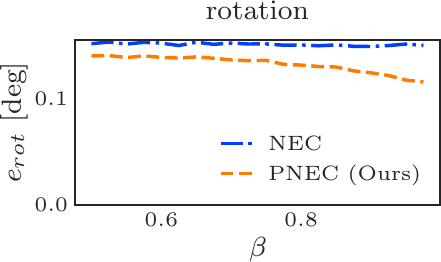}}
        \caption{Experiment w/o translation: $e_{rot}$}
        \label{fig:omni_aniso_no_t}
    \end{subfigure}
    \caption{Influence of the degree of anisotropy for omnidirectional cameras. Results averaged of 10\,000 random problems with varying degree of anisotropy. While a higher anisotropy is beneficial for our \ac{pnec} it outperforms the \ac{nec} for isotropic noise ($\beta = 0.5$) as well.}
    \label{fig:omni_aniso}
\end{figure*}

\begin{figure*}[t]
    \centering
    \begin{subfigure}[b]{0.32\textwidth}
        \centering
        {\includegraphics[width=\textwidth]{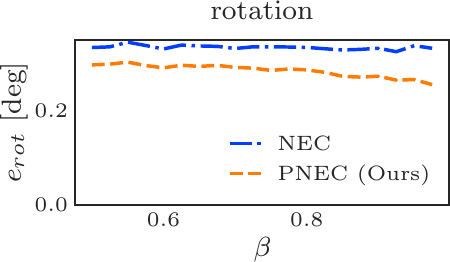}}
        \caption{Experiment w/ translation: $e_{rot}$}
        \label{fig:pin_aniso_rot}
    \end{subfigure}
    \hfill
    \begin{subfigure}[b]{0.31\textwidth}
        \centering
        {\includegraphics[width=\textwidth]{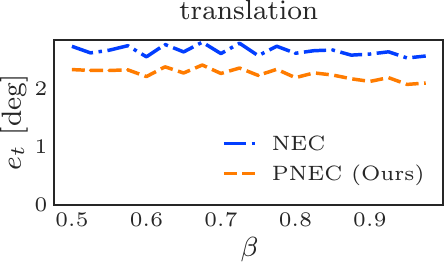}}
        \caption{Experiment w/ translation: $e_{t}$}
        \label{fig:pin_aniso_t}
    \end{subfigure}
    \hfill
    \begin{subfigure}[b]{0.32\textwidth}
        \centering
        {\includegraphics[width=\textwidth]{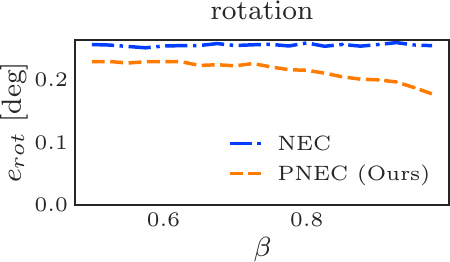}}
        \caption{Experiment w/o translation: $e_{rot}$}
        \label{fig:pin_aniso_no_t}
    \end{subfigure}
    \caption{Influence of the degree of anisotropy for pinhole cameras. Results are similar as for omnidirectional cameras.}
    \label{fig:pin_aniso}
\end{figure*}

As the results of the previous experiment shows, the performance of the \ac{pnec} is dependent on the noise type. The following experiments show the effect the anisotropy of the noise has on the \ac{pnec} optimization. The setup for this experiment is the same as for the previous experiments, apart from the noise generation. While the noise level stays constant  ($1.0\, \text{[pix]}$), inhomogenous noise is sampled over different levels of anisotropy $\beta$ ($\beta = 0.5 $: isotropic, $\beta = 1.0$: anisotropic). \autoref{fig:omni_aniso} and \autoref{fig:pin_aniso} show the results for omnidirectional and pinhole cameras, respectively. We repeated the experiments for pure rotation. An increasing level of anisotropy is beneficial for our \ac{pnec}, increasing the gap between the \ac{nec} and the \ac{pnec} noticeably. Additionally, the results show that our \ac{pnec} outperforms the \ac{nec} even for isotropic noise.

\section{Offset Experiments} \label{sec:offset}
\begin{figure*}[t]
    \centering
    \begin{subfigure}[b]{0.32\textwidth}
        \centering
        {\includegraphics[width=\textwidth]{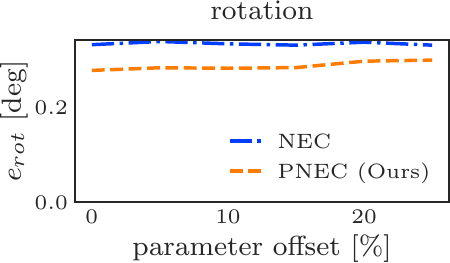}}
        \caption{Experiment w/ translation: $e_{rot}$}
        \label{fig:pin_offset_rot}
    \end{subfigure}
    \hfill
    \begin{subfigure}[b]{0.31\textwidth}
        \centering
        {\includegraphics[width=\textwidth]{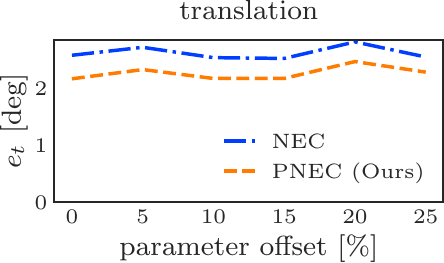}}
        \caption{Experiment w/ translation: $e_{t}$}
        \label{fig:pin_offset_t}
    \end{subfigure}
    \hfill
    \begin{subfigure}[b]{0.32\textwidth}
        \centering
        {\includegraphics[width=\textwidth]{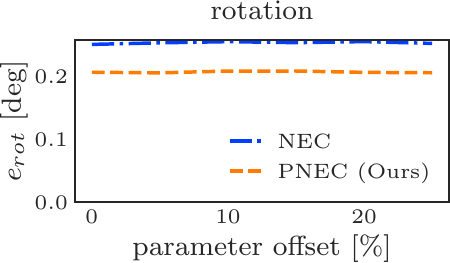}}
        \caption{Experiment w/o translation: $e_{rot}$}
        \label{fig:pin_offset_no_t}
    \end{subfigure}
    \caption{Influence of wrongly estimated noise parameters. Our \ac{pnec} is given covariance matrices with wrong noise parameters (see \autoref{sec:simulated}) for the optimization. Results are averaged over 10\,000 random problems over different levels of maximum parameter offset. Our \ac{pnec} outperforms the \ac{nec} even if the noise parameters have an offset of up to $25\%$.}
    \label{fig:pin_offset}
\end{figure*}

The previous synthetic experiments showed the robustness of our \ac{pnec} against different noise intensities. For these experiments, we assumed perfect knowledge about the noise distribution. \autoref{fig:pin_offset} shows the influence of wrongly estimated noise parameters on the performance of our \ac{pnec}. For these experiments we generated random pinhole camera problems but gave the \ac{pnec} covariance matrices with a random offset on the noise parameters. The offset is uniformly sampled with a deviation up to a certain percentage of the range of the noise parameters $\alpha, \beta, s$, respectively. The results show that the \ac{pnec} performs better, the more accurate the covariance matrices are. However, our \ac{pnec} outperforms the \ac{nec} even with noise parameters that are off by up to $25\%$.
\section{KITTI experiments} \label{sec:kitti}
In the following we present the results on the KITTI dataset in more detail. In \autoref{subsec:trans-variance} we compare the \ac{nec} and \ac{pnec} in their translation estimation as well as the consistency of their results. \autoref{subsec:ablation} gives an ablation study on the KITTI dataset to evaluate the performance of both stages of our optimization scheme.

\subsection{Translation and Variance} \label{subsec:trans-variance}
\begin{figure*}[t]
    \centering
    \begin{subfigure}[b]{0.48\textwidth}
        \centering
        {\includegraphics[width=\textwidth]{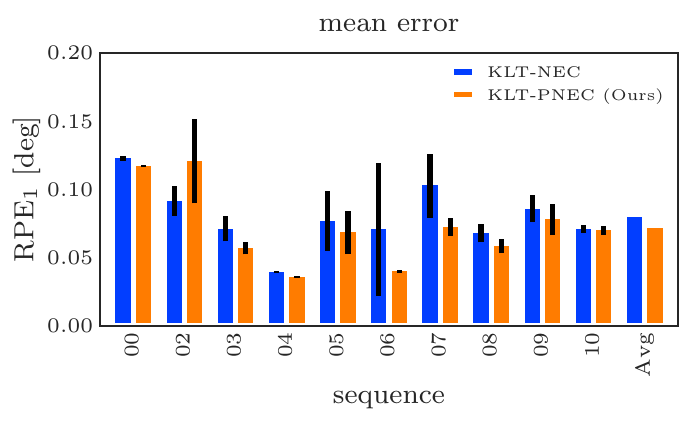}}
        \caption{$\text{RPE}_1$}
        \label{fig:kitti_variance_rpe1}
    \end{subfigure}
    \hfill
    \begin{subfigure}[b]{0.48\textwidth}
        \centering
        {\includegraphics[width=\textwidth]{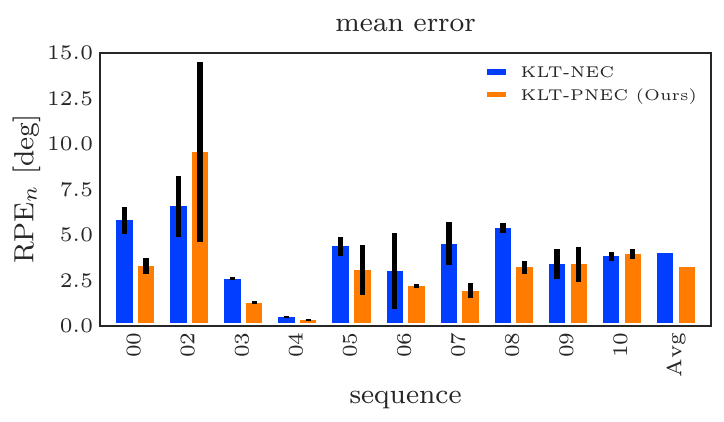}}
        \caption{$\text{RPE}_n$}
        \label{fig:kitti_variance_rpen}
    \end{subfigure}
    \begin{subfigure}[b]{0.48\textwidth}
        \centering
        {\includegraphics[width=\textwidth]{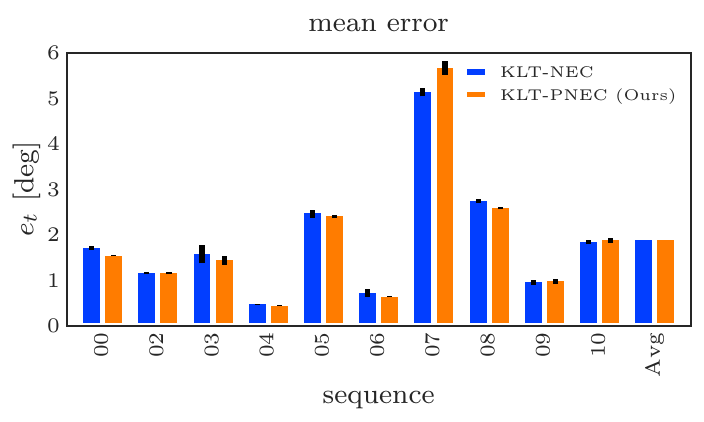}}
        \caption{$e_{\text{t}}$}
        \label{fig:kitti_variance_t}
    \end{subfigure}
    \caption{Mean error with standard deviation over 5 runs of the KITTI dataset. Comparison between the \ac{nec} and our \ac{pnec} shows that not only does our \ac{pnec} better results on average for both the $\text{RPE}_1$ (see \autoref{fig:kitti_variance_rpe1}) and the $\text{RPE}_n$ (see \autoref{fig:kitti_variance_rpen}) but the results are often more consistent. The translational error for both methods are very similar on average.}
    \label{fig:kitti_variance}
\end{figure*}

\autoref{fig:kitti_variance} shows the mean error and standard deviation of the \ac{klt}-\ac{nec} and the \ac{klt}-\ac{pnec} on the KITTI dataset for the $\text{RPE}_1$, the $\text{RPE}_1$ and $e_{\text{t}}$ metrics. We chose to omit seq.~01 since neither tracks nor covariances provided by the \ac{klt} tracker are correct. The results show that our \ac{pnec} not only achieves better result on average in the rotation estimation but its results are more consistent for most sequences. The translation estimation for both methods is very similar.

\subsection{Ablation study} \label{subsec:ablation}

\begin{figure*}[t]
    \centering
    \begin{subfigure}[b]{0.48\textwidth}
        \centering
        \includegraphics[width=\textwidth]{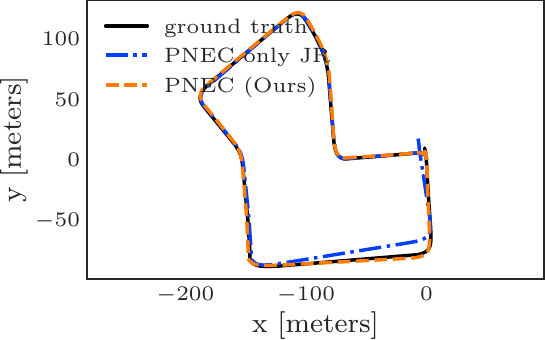}
        \label{fig:kitti_trajectory1}
    \end{subfigure}
    \hfill
    \begin{subfigure}[b]{0.48\textwidth}
        \centering
        \includegraphics[width=\textwidth]{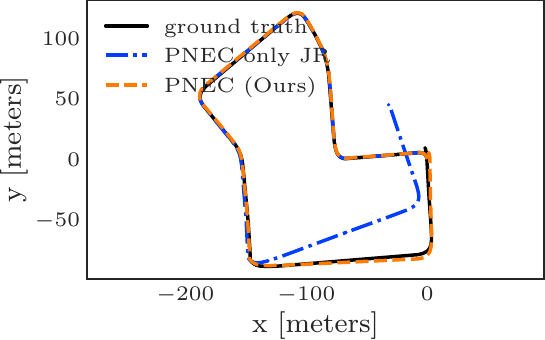}
        \label{fig:kitti_trajectory2}
    \end{subfigure}
    \hfill
    \begin{subfigure}[b]{0.48\textwidth}
        \centering
        \includegraphics[width=\textwidth]{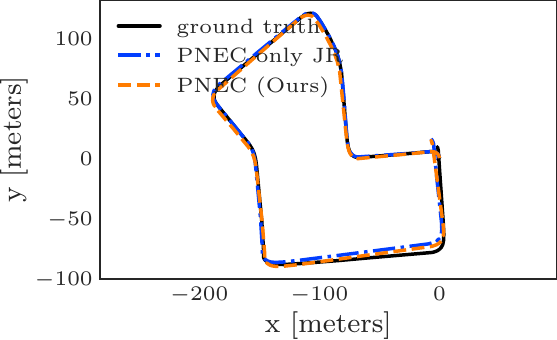}
        \label{fig:kitti_trajectory3}
    \end{subfigure}
    \hfill
    \begin{subfigure}[b]{0.48\textwidth}
        \centering
        \includegraphics[width=\textwidth]{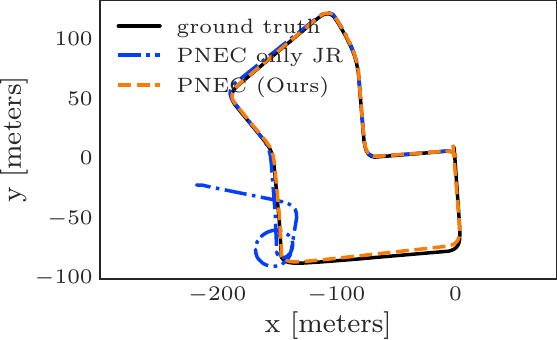}
        \label{fig:kitti_trajectory4}
    \end{subfigure}
    \hfill
    \begin{subfigure}[b]{0.48\textwidth}
        \centering
        \includegraphics[width=\textwidth]{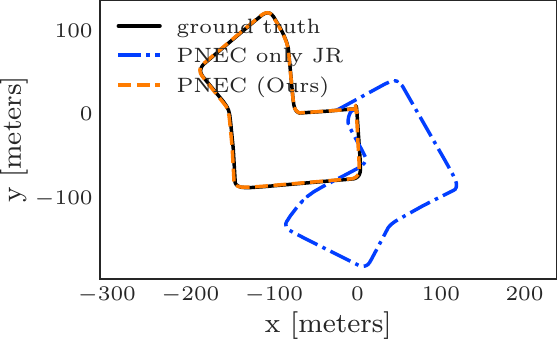}
        \label{fig:kitti_trajectory5}
    \end{subfigure}
    \caption{Qualitative evaluation for KITTI seq.~07. The trajectories are generated with the estimated rotations of \ac{pnec} only JR and our \ac{pnec}, respectively, and are combined with the ground truth translations for visualization purposes. The trajectories show the volatility of the \ac{pnec} only JR. While the rotations estimated with only the joint refinement are good over most of the sequence, it is prone to local minima due to the least-square formulation. A handful of bad rotation estimations, mainly located in corners, lead to an overall poor performance on the whole sequence. The first stage of the proposed optimization scheme overcomes this issue by providing a good initialization for the joint refinement.}
    \label{fig:kitti_trajectories}
\end{figure*}

\autoref{tab:kitti-supp} shows the results for the ablation study on the KITTI dataset.  We compare: \ac{klt}-\ac{nec}; \ac{nec} followed by the 2nd stage of our optimization scheme (\ac{klt}-\ac{nec} + \ac{pnec}-JR); only the 1st stage (\ac{klt}-\ac{pnec} w/o JR); only the 2nd stage (\ac{klt}-\ac{pnec} only JR); both stages of the optimization (\ac{klt}-\ac{pnec}). Additionally we included the MRO results as a baseline. For KITTI all methods are initialized with a constant velocity motion model. The full optimization with both stages gives the best results on most sequences and performs the best on average. Both stages on their own achieve considerably worse results. \autoref{fig:kitti_trajectories} shows qualitative trajectories for \ac{klt}-\ac{pnec} only JR and \ac{klt}-\ac{pnec} of five runs on seq.~07 of the KITTI dataset. While our full \ac{pnec} achieves consistent results over all runs, \ac{klt}-\ac{pnec} only JR results are considerably more volatile. The joint refinement is prone to local mimima due its least-squares formulation. This leads to a few poor rotation estimations that worsen the performance on the sequence drastically.

\begin{table*}[t]
    \small
    \centering
    \sisetup{detect-weight,mode=text}
    \renewrobustcmd{\bfseries}{\fontseries{b}\selectfont}
    \renewrobustcmd{\boldmath}{}
    \newrobustcmd{\B}{\bfseries}
    \addtolength{\tabcolsep}{-3.5pt}

    \begin{tabular} {p{1.6cm} r r|r r|r r|r r|r r|r r}
        \toprule
                                  & \multicolumn{2}{c}{\scshape MRO~\cite{MRO_Chng2020}} & \multicolumn{2}{c}{\scshape KLT-\ac{nec}} & \multicolumn{2}{c}{\scshape KLT-\ac{nec}} & \multicolumn{2}{c}{\scshape KLT-\ac{pnec}} & \multicolumn{2}{c}{\scshape KLT-\ac{pnec}} & \multicolumn{2}{c}{\scshape KLT-\ac{pnec}}                                                                                                                 \\
                                  & \multicolumn{2}{c}{}                                 & \multicolumn{2}{c}{}                      & \multicolumn{2}{c}{+ \ac{pnec}-JR}        & \multicolumn{2}{c}{w/o JR}                 & \multicolumn{2}{c}{only JR}                & \multicolumn{2}{c}{(Ours)}                                                                                                                                 \\
        Seq.                      &
        {\scshape $\text{RPE}_1$} & {\scshape $\text{RPE}_n$}                            &
        {\scshape $\text{RPE}_1$} & {\scshape $\text{RPE}_n$}                            &
        {\scshape $\text{RPE}_1$} & {\scshape $\text{RPE}_n$}                            &
        {\scshape $\text{RPE}_1$} & {\scshape $\text{RPE}_n$}                            &
        {\scshape $\text{RPE}_1$} & {\scshape $\text{RPE}_n$}                            &
        {\scshape $\text{RPE}_1$} & {\scshape $\text{RPE}_n$}                                                                                                                                                                                                                                                                                                                                                                           \\
        \midrule
        \midrule
        00                        & 0.360                                                & 8.670                                     & \underline{0.125}                         & 5.922                                      & 0.139                                      & \underline{5.392}                          & 0.142     & 9.594             & 0.210             & 14.929            & \B 0.119          & \B 3.429          \\
        02                        & 0.290                                                & 16.030                                    & \underline{0.093}                         & \underline{6.693}                          & 0.111                                      & 8.781                                      & 0.100     & 10.259            & \B 0.077          & \B 5.271          & 0.122             & 9.687             \\
        03                        & 0.280                                                & 5.470                                     & 0.073                                     & 2.728                                      & 0.063                                      & 1.717                                      & 0.065     & 2.022             & \B 0.057          & \underline{1.464} & \underline{0.059} & \B 1.411          \\
        04                        & 0.040                                                & 1.080                                     & 0.041                                     & 0.619                                      & \B  0.038                                  & 0.477                                      & \B  0.038 & 0.959             & \B  0.038         & \B  0.448         & \B  0.038         & \underline{0.463} \\
        05                        & 0.250                                                & 11.360                                    & 0.079                                     & 4.489                                      & 0.082                                      & \underline{3.659}                          & 0.094     & 6.313             & \underline{0.078} & 6.563             & \B  0.070         & \B  3.203         \\
        06                        & 0.180                                                & 4.720                                     & 0.073                                     & 3.162                                      & 0.072                                      & 2.988                                      & 0.054     & \B 1.577          & \underline{0.043} & 2.366             & \B  0.042         & \underline{2.322} \\
        07                        & 0.280                                                & 7.490                                     & 0.105                                     & 4.640                                      & \underline{0.096}                          & \underline{2.092}                          & 0.135     & 7.592             & 0.545             & 25.381            & \B  0.074         & \B  2.065         \\
        08                        & 0.270                                                & 9.210                                     & \underline{0.070}                         & 5.523                                      & 0.084                                      & \underline{5.004}                          & 0.072     & 5.973             & 0.127             & 27.343            & \B  0.060         & \B  3.347         \\
        09                        & 0.280                                                & 9.850                                     & 0.088                                     & 3.533                                      & 0.088                                      & \B  2.699                                  & \B  0.079 & \underline{2.783} & 0.202             & 23.842            & \underline{0.080} & 3.514             \\
        10                        & 0.380                                                & 13.250                                    & \underline{0.073}                         & \B  3.959                                  & 0.075                                      & 4.155                                      & 0.077     & 4.679             & 0.177             & 15.778            & \B     0.072      & \underline{4.094} \\
        \midrule
        mean                      & 0.261                                                & 8.713                                     & \underline{0.082}                         & 4.127                                      & 0.085                                      & \underline{3.696}                          & 0.086     & 5.175             & 0.155             & 12.338            & \B 0.074          & \B 3.354          \\
        \bottomrule
    \end{tabular}
    \caption{Ablation study on the KITTI dataset. We compare: \ac{klt}-\ac{nec}; \ac{nec} followed by the 2nd stage of our optimization scheme (\ac{klt}-\ac{nec} + \ac{pnec}-JR); only the 1st stage (\ac{klt}-\ac{pnec} w/o JR); only the 2nd stage (\ac{klt}-\ac{pnec} only JR); both stages of the optimization (\ac{klt}-\ac{pnec}). Additionally we included the MRO results as a baseline. For KITTI all methods are initialized with a constant velocity motion model, i.e. with the relative rotation generated for the previous frame pair by the same method. The results show that only both stages of our optimization achieve the most consistent and best results on average. Using only one stage performs worse. Especially the 2nd stage struggles on most sequences. While it has the best results on some sequences on others is performs the worst often by a wide margin in the $\text{RPE}_n$ metric. The joint refinement is dependent on a good initialization, which is provided by the 1st stage of our optimization. \ac{klt}-\ac{nec} + \ac{pnec}-JR shows that using the \ac{nec} is not enough to provide the best results.}
    \label{tab:kitti-supp}
\end{table*}

\end{document}